\def\preprint{}

\ifdefined\preprint \def\cleanbuild{}\fi

\ifdefined\cleanbuild
  \AtBeginDocument{\long\def\JScomment#1{}\long\def\SSTcomment#1{}}
\fi

\ifdefined\preprint
  \documentclass[12pt]{article}
  \usepackage[margin=1.2in]{geometry}
  \usepackage{amsmath,amsxtra,amssymb,xspace,url,enumitem,authblk}
  \usepackage[T1]{fontenc}
  \usepackage{newtxtext,newtxmath}
  
  \date{}
  \newcommand{\CUPseries}[1]{}
  \newcommand{\CUPelements}[1]{}
  \newcommand{\copyrightauthor}[1]{}
  \newcommand{\keywords}[1]{\par\noindent\textbf{Keywords:} #1\par}
  \newcommand{\frontmatter}{}
  \newcommand{\mainmatter}{\clearpage\tableofcontents\clearpage}
  \newcommand{\addsource}[1]{%
    \par\vspace*{0.2em}%
    \begingroup\raggedright\fontsize{9pt}{12pt}\selectfont
    \textbf{Source: }#1\par
    \endgroup}
  \newcommand{\email}[1]{\\ \texttt{#1}}
\else
  \documentclass{cup-elements}
  \newcommand{\email}[1]{}
\fi

\let\amscheckmark\checkmark
\let\checkmark\undefined
\usepackage{dingbat}
\let\checkmark\amscheckmark
\usepackage{multirow}
\usepackage{graphicx}
\usepackage{todonotes}

\usepackage[natbibapa]{apacite}
\CUPseries{Elements in Semantics and Pragmatics}
\CUPelements{Computational Explanations of Semantic Universals} %

\title{Computational Explanations of Semantic Universals}

\newcommand{\sv}[1]{\ensuremath{\llbracket #1 \rrbracket}}
\newcommand{\co}{\colon\thinspace}
\DeclareRobustCommand\dash{\unskip\nobreak\thinspace\textemdash\thinspace\ignorespaces}

\usepackage{bbm}
\usepackage{linguex}
\usepackage{example_sentences}

\usepackage{amsthm}
\usepackage{booktabs}
\usepackage{verbatim}

\usepackage{tipa} %

\newcommand{\flavor}{\text{fl}} %
\newcommand{\force}{\text{fo}}

\usepackage{xcolor}

\ifdefined\cleanbuild
  \long\def\JScomment#1{}
  \long\def\SSTcomment#1{}
\else
  \long\def\JScomment#1{\begingroup\sffamily\small\color{blue}[JS: #1]\endgroup}
  \long\def\SSTcomment#1{\begingroup\sffamily\small\color{red!70!black}[SST: #1]\endgroup}
\fi

\ifdefined\cleanbuild
  \AtBeginDocument{\long\def\JScomment#1{}\long\def\SSTcomment#1{}}
\fi

\author{Shane Steinert-Threlkeld}
\affil{Department of Linguistics \\ University of Washington \email{shanest@uw.edu}}

\author{Jakub Szymanik}
\affil{Department of Psychology and Cognitive Science \\ University of Trento \email{jakub.szymanik@gmail.com}}

\ifdefined\preprint \usepackage{hyperref}\fi

\begin{document}

\frontmatter  %
\maketitle

\begin{abstract}
Despite their differences, the languages of the world share many deep features when it comes to how they express meaning.  These semantic universals are likely to reflect shared aspects of human cognitive, communicative, and social structure.  Recent advances in cross-linguistic semantics and computational linguistics allow us to develop explanations of the sources of such properties. This Element presents a body of work doing just that: we explicate a framework for providing computational explanations of semantic universals and then describe several such explanations using principles of ease of learning and efficient communication.  Empirical domains include quantifiers, color terms, modals, and indefinites.  After providing these explanations, we provide a discussion of future directions for advancing the explanatory power of computational methods in semantic typology.
\end{abstract}


\copyrightauthor{Steinert-Threlkeld, Szymanik 2026}

\mainmatter  %

\section{Explanations of Lexical Property Universals}

\label{chapter:introduction}

While linguistics as a field may often seem to be characterized by fiery disagreement, linguists of all stripes agree on at least two fundamental facts: (i) language is a system for pairing meanings and forms and (ii) the languages of the world exhibit \emph{constrained variation}: although languages differ, they also exhibit much shared structure. 

When the languages of the world exhibit shared structure, i.e. have a property in common, that property is called a linguistic universal \citep{greenbergLanguageUniversals1966, hulstQuestionLinguisticUniversals2008, evansMythLanguageUniversals2009}. Examples abound, at every level of linguistic analysis.
At the phonological level, all languages have consonants and vowels.  More robustly, one can say that all languages have at least one unrounded vowel and at least one back vowel \citep{hymanUniversalsPhonology2008}.\footnote{These two are called ``Vocalic Universal \#3'' and ``Vocalic Universal \#4'' on p. 98 in the aforementioned survey.}
At the syntactic level, all languages have verbs and nouns \citep{croftTypologyUniversals1990}.\footnote{\citet{newmeyerUniversalsSyntax2008} calls for hesitation on positing universals about syntactic categories. \citet{hengeveldPartsofspeechSystemsWord2004} discuss examples of languages that seem not to make the distinction.}  Slightly more controversially, generative grammar as an enterprise can be seen as systematically developing syntactic universals.  For example, the basic claim that grammatical rules are structure-dependent is a syntactic universal \citep[i.a.]{chomskyAspectsTheorySyntax1965}. At the semantic level, it has been proposed that all languages which have shape adjectives also have color and size adjectives \citep{dixonWhereHaveAll1977}.

In an influential early survey of semantic universals, \cite{vonfintelUniversalsSemantics2008} claim that ``the state of the art in semantic universals is largely too immature to allow explorations of their sources'' (p. 147).  In the intervening years, this state of the art has moved forward in several directions.  There has been a considerable growth in the amount of cross-linguistic semantic work being done; improvements in fieldwork methodologies have also enabled such work to be done in a way that facilitates direct comparisons of (parts of) semantic systems in many languages.\footnote{See~\citet{bochnakMethodologiesSemanticFieldwork2015, bochnakTechniquesComplexSemantic2020} for surveys of semantic fieldwork methodology and  \citet{vanderklokModalityUnderdescribedLanguages2022} for an example of using such methods for one empirical domain (modality).}   With data like these in hand, several other new methods have been developed which allow one to develop possible explanations as to the origins of particular semantic universals and identifying general principles underlying these explanations \citep{kempSemanticTypologyEfficient2018, steinert-threlkeldEaseLearningExplains2020}.  The present Element offers a snapshot of several strands of this research, focusing on computational explanations of semantic universals.

In this introductory chapter, we will first hone in and make precise what kinds of semantic universals we will discuss and attempt to explain, illustrating with one example: monotonicity of quantifiers.  We will then provide a general setting for offering computational explanations, explicating the various choice points that one encounters along the way, as well as highlighting two prominent hypotheses about semantic universals: that they arise due to ease of learning and due to efficient communication.  Finally, we will summarize this framework by situating existing work within it before going on to provide detailed explanations in the subsequent chapters.

\subsection{Semantic Universals}
\label{sec:semantic-universals}

So far, we have just said that a semantic universal reflects `shared structure' of languages at the level of meaning.  In this section, we will hone in on the exact type of universals that will be our focus and illustrate with one example.

One can first distinguish what we will call \emph{architectural} universals from \emph{expressive} universals.  The former concern the nature of the semantic system, whereas the latter concern the kinds of meanings that can be expressed by natural language expressions.  On the architectural side, examples include arguments that all languages have a distinction between literal content and presuppositions \citep{matthewsonPresuppositionsCrossLinguisticVariation2006, tonhauserProjectionVariabilityParaguayan2020} and that there exist only a limited number of semantic composition rules (e.g.\ function application; \citealp{kleinTypeDrivenTranslation1985, bittnerCrossLinguisticSemantics1994, jacobsonCompositionalSemanticsIntroduction2014}).

On the expressive side, there are examples spanning a wide range of granularities.
At one extreme, one could investigate \emph{the set of all meanings} that can be expressed by sentences in a language.  A natural idea holds that all languages are equally expressive, i.e.\ every natural language can express the same set of meanings \citep{fregeCompoundThoughts1963, katzHypothesisUniquenessNatural1976, benthemLinguisticUniversalsLogical1991}. On this view, all natural languages have the same expressive power: they are equally able to express any thought one may wish to communicate, even though they may differ in the type of sentence that must be used to do so.\footnote{See \S1.2 of \citet{vonfintelUniversalsSemantics2008} for more discussion, including possible counterexamples. This thesis of equal
expressive power should be distinguished from the related Equivalent
Complexity Thesis, that all languages are roughly equally complex overall
\citep{hockett1958course}, which has been debated mainly for morphosyntax
\citep{miestamo2008language, sampson2009language}. The two are sometimes
linked: equal expressive power is one reason given for expecting complexity
in one part of a grammar to be compensated elsewhere.}  At another extreme, one could investigate the simplest meaning-bearing elements of a language: morphemes. On this end, a number of proposals have been made arguing that certain individual meanings are expressed in every language \citep{goddardSemanticLexicalUniversals1994, wierzbickaSemanticsPrimesUniversals1996, goddardLexicoSemanticUniversalsCritical2001}.  For example, the aforecited authors have suggested that negation and the meaning of the verb `think' are universally expressed by single morphemes.  Because these arguments say that an individual meaning is lexicalized (attached to one morpheme), they are often referred to as lexical universals.

\subsubsection{Lexical Property Universals}

Most of the universals for which we will offer explanations in this book lie in between these two extremes; they are what we call \emph{lexical property universals}.  They state that every lexical item (of a given syntactic category, or within a given semantic domain) has a particular property.  Lexical property universals resemble lexical universals in that they focus on the meanings attached to individual items in the lexicon.  But they do not focus on individual meanings, since it is often extremely difficult to say when two morphemes (in two different languages, though the problem arises even within one language) have \emph{the exact same} meaning \citetext{\citealp[p.~145]{bloomfieldLanguage1933}; \citealp[p.~281]{nidaAnalysisMeaningDictionary1958}}. Instead, the universals that we focus on say that all expressions of a particular type share something in common: there is a property that we can more readily verify for individual meanings which is in fact shared by all of the relevant expressions.  We can codify the notion of a lexical property universal into a schema, which will see instantiated throughout this Element:
\begin{itemize}[align=left, leftmargin=*]
	\item[\textsc{Lexical Property Universal Schema:}] All lexical items $e$ of a particular kind $K$ express meanings with property $P$.
\end{itemize}
In order to instantiate this schema, one must specify both the lexical items of interest and the semantic property that these items share in common.  To start, we illustrate this with one example.

\subsubsection{Example: Monotonicity of Quantifiers}
\label{sec:monotonicity-definition}

To give a motivating example, which will be useful in the subsequent discussion and which will be more fully developed in explanations later in this Element, we will focus on a universal about quantifiers, which are the semantic objects expressed by \emph{determiners}.  A determiner is an expression taking a common noun as an argument and generating a Noun Phrase.  Determiners can be monomorphemic---for example, \emph{all, some, no, few}---or they can be morphosyntactically complex---for example, \emph{all but five, at most three, an even number of}.

As a first approximation, and following the influential \citet{barwiseGeneralizedQuantifiersNatural1981}, we assume that determiners denote type $\langle 1, 1 \rangle$ generalized quantifiers.  These can be thought of as being relations between two subsets of a given domain of discourse. For example:
\begin{align*}
	\sv{\text{every}} &= \left\{ \langle M, A, B \rangle \co A \subseteq B \right\}
	\\
	\sv{\text{at most 3}} &= \left\{ \langle M, A, B \rangle \co |A \cap B| \leq 3 \right\}
	\\
	\sv{\text{most}} &= \left\{ \langle M, A, B \rangle \co \lvert A \cap B \rvert > \lvert A \setminus B \rvert \right\}
\end{align*}

Before proceeding, a few small notes on terminology.  As a shorthand, we will say that a determiner has a certain semantic property to mean that the quantifier that the determiner denotes has that property.  Sometimes, for a determiner like \emph{every}, we will write \textsf{every} as a shorthand for $\sv{\text{every}}$. We will use \textsf{Q} and its ilk as variables over quantifiers.  Because quantifiers are viewed as set-theoretic objects, we will write $\mathcal{M} \in \textsf{Q}$ when a structure/model $\mathcal{M}$ belongs to a quantifier.\footnote{See~\citet{petersQuantifiersLanguageLogic2006, Szymanik:2015kq} for a thorough exposition of quantifiers in this tradition.}  In other words, when a sentence \emph{Det N VP} is true when interpreted in a model $\mathcal{M}$, we will write $\langle M, \sv{\text{N}} , \sv{\text{VP}} \rangle \in \textsf{Det}$.

To motivate our first lexical property universal, consider the following sentences.
\begin{examples}
\item(ams-bike) \begin{examples}
	\item(ams-bike-heavy) Many Amsterdammers ride a heavy bike to work.
	\item(ams-bike-only) Many Amsterdammers ride a bike to work.
	\end{examples}
\end{examples}
It is clear that \exref{ams-bike-heavy} entails \exref{ams-bike-only}: the former cannot be true without the latter being true.  Similarly, this entailment does not depend on the choice of the restrictor \dash \emph{Amsterdammers} \dash or scopes \dash \emph{ride a heavy bike to work} and \emph{ride a bike to work} \dash so long as the latter scope is strictly more general than the former. Moreover, competent speakers of English recognize this fact easily. What speakers thereby implicitly know is that \emph{many} is upward monotone:
\begin{examples}
\item \textsf{Q} is \emph{upward monotone} if and only if whenever $\langle M, A, B \rangle \in \textsf{Q}$ and $B \subseteq B'$, then $\langle M, A, B' \rangle \in \textsf{Q}$.
\end{examples}

By contrast, the pattern seems to reverse if we replace \emph{many} with \emph{few}, as seen in the following examples.
\begin{examples}
\item \begin{examples}
	\item(few2) Few Amsterdammers ride a heavy bike to work.
	\item(few1) Few Amsterdammers ride a bike to work.
	\end{examples}
\end{examples}
Here, \exref{few1} entails \exref{few2}.  This is the reverse of the previous case: now, truth is preserved when we move from a more general scope to a more specific scope.  In this case, we say that \emph{few} is downward monotone:
\begin{examples}
\item \textsf{Q} is \emph{downward monotone} if and only if whenever $\langle M, A, B \rangle \in \textsf{Q}$ and $B \supseteq B'$, then $\langle M, A, B' \rangle \in \textsf{Q}$.
\end{examples}

Finally, a determiner is \emph{monotone} just in case it is either upward or downward monotone.  The reader can verify that all of the simple determiners mentioned at the beginning of the section are monotone.  This appears to be no accident of our choice of English or of that particular list of simple determiners.  \citet{barwiseGeneralizedQuantifiersNatural1981} proposed the following semantic universal:
\begin{itemize}[align=left, leftmargin=*]
	\item[\textsc{Monotonicity Universal:}] All monomorphemic determiners are monotone or conjunctions of monotone quantifiers.\footnote{The conjunction clause is tended to cover exact numerals like \textsf{exactly three}, with meaning \textsf{at least three and at most three}.  If bare numerals like ``three'' denote \textsf{at least three} as opposed to \textsf{exactly three}, then the conjunction clause may not be needed.} 
\end{itemize}
This universal rules out quantifiers such as \textsf{an even number of} and \textsf{at least 6 or at most 2}: increasing or decreasing the set $B$ can cause the cardinality of $A \cap B$ to change in a way that flips the truth value of sentences with those determiners, so they are neither monotone nor conjunctions of monotone quantifiers..  The claim then is that no simple determiner in any natural language denotes those quantifiers.

Finally, we note that in our terminology, the \textsc{Monotonicity Universal} is a paradigm lexical property universal, instantiating the schema from the previous subsection: it states that all lexical items of a given kind (here, syntactically defined: determiners) have meanings (quantifiers) with a particular property (monotonicity, in the extended sense just given).

\subsubsection{Degrees of Semantic Universals}
\label{sec:degrees-universals}

Although lexical property universals really are universals---they say that \emph{all} expressions of a given type share a particular property---many of our explanations will leverage \emph{degrees of semantic universals}.  That is to say: for many of the lexical property universals that we discuss in this Element, we will introduce ways of measuring \emph{degrees} of those properties.  In other words, we will consider graded/gradable versions of the properties. Typically, degrees will belong to the closed interval $[ 0, 1 ]$, where having the (binary) property corresponds to having the (graded) property to degree 1.

For example, we can introduce a \emph{degree of monotonicity} for quantifiers, as first defined in  \citet{steinert-threlkeldQuantifiersNaturalLanguage2021, carcassiMonotoneQuantifiersEmerge2021b}.
This measure is designed to capture the fact that some quantifiers are intuitively \emph{more} monotone than other quantifiers. For instance, consider the three quantifiers ``some'', ``between 3 and 5'' and ``an even number of''. While ``some'' is monotone and the other two quantifiers are not, intuitively ``an even number of'' is the least monotone of the three, because it is not equivalent to a Boolean combination of monotone quantifiers. 

While we refer the reader to Appendix~\ref{chap:monotonicity-appendix} for a formal definition, we here provide the core idea.  The definition first defines a `minimal monotone extension' of a given quantifier; roughly, the closest monotone quantifier to a given one.  For example, the minimal monotone extension of \textsf{between 3 and 5} is \textsf{at least 3}.  Then, the degree of monotonicity of a quantifier, denoted $\textsf{mon}(\textsf{Q})$, is defined as the amount of uncertainty about \textsf{Q} that is resolved by knowing its minimal monotone extension.  This definition both has the desirable property that $\textsf{mon}(\textsf{Q}) = 1 \text{ iff } \textsf{Q}$ is upward monotone\footnote{The Appendix shows how to apply the definition to downward monotonicity as well} and captures the intuitive ordering above:
\cite{carcassiMonotoneQuantifiersEmerge2021b} report that ``some'' has degree of monotonicity 1.0, reflecting the fact just mentioned while ``between 3 and 5'' has degree $0.7517$ and ``an even number of'' has degree $0.001$, capturing the intuitive order of monotonicity of these quantifiers.

Having shown as one example that monotonicity can also be measured in degrees, we note several benefits of measuring degrees of semantic universals that will be made more explicit in what follows.  First and foremost, degrees let us say how far a potential exception departs from a universal, which matters given debates about whether exception-less universals exist at all \citep{evansMythLanguageUniversals2009}.
Even if, for example, one were to find a non-monotone monomorphemic determiner, it would likely still have a high degree of monotonicity.  In other words, it enables the expression of finer-grained generalizations than just the lexical property universal schema.  Furthermore, even for full-fledged lexical property universals, which will continue to be our main focus, using degrees of universal properties will allow for finer-grained and more satisfying \emph{explanations}.  As discussed below, many of our explanations will involve different notions of \emph{optimization}.  This concept naturally comes in degrees: languages can be more and less optimal according to the criteria we will defined.  Having degrees of universals will allow us to say not just that the universals themselves are the \emph{outcome} or final state of an optimization process, but rather that being more and less optimal is well-correlated with having a universal property to higher or lower degree.

\subsection{Computational Explanations}

By computational explanations, we mean models (often simulations) of the presence and/or emergence of semantic universals (and lexical property universals in particular).  The models reviewed in this book capture various factors shaping language, either on an individual level, like language learning or on the community level, like language evolution or communication. The explanatory part clarifies possible causal relationships between the modeled phenomenon and the studied universal. In every explanation, we will aim to provide a principle that links outcomes from a model to observed linguistic universals.  Most of these principles will take the form of \emph{optimizations}: we will argue that shared language structure corresponds to those languages which are more optimal in particular senses compared to other non-actual languages.

The computational explanations presented in this book obviously have clarity benefits over verbal theories. All the models' assumptions are transparent, and the underlying code is openly shared. The models are directly testable and evaluated against data. Modeling minimizes vagueness and facilitates replications and the development of further, potentially improved models. 
At the same time, computational models are idealizations abstracting away from many intricacies of the problem. Because this idealization is transparent, it does not obscure the issue but quite the opposite; it makes understanding and explanations possible. Still, one needs to be cautious about the assumptions and evaluate whether the models' explanations are strong enough to withstand the weakening of the idealizing assumptions. 

In the remainder of this section, we will outline and discuss many of the significant modeling assumptions that must be made in order to provide a computational explanation for a semantic universal.  These roughly follow the components of the \textsc{Lexical Property Universal Schema} as well as an optimization principle.

\subsubsection{Parameters}
\label{sec:parameters}

\paragraph{Kind of expression}

The first component for defining a lexical property universal and providing an explanation thereof requires specifying the \emph{kind} of expression of interest (this is the $K$ in the \textsc{Schema}).  In the example of monotonicity, this kind is syntactically defined: it is determiners in the lexicon.

All lexical property universals presuppose that the expressions that fall under the kind/label, e.g., `generalized quantifiers' or `indefinite pronoun', form a uniform category across languages. One minimal way to establish that something is a category is to provide a set of criteria that would separate members from non-members, e.g., a collection of standards that guides us towards classifying the expression `many' as a member or not of the determiner category. 

In what follows, we will provide such minimal criteria for identifying the categories under discussion. We, however, would like to make a provision that such standards are rarely perfect or non-controversial. They often suffer at the same time from being too narrow, excluding expressions that intuitively should fall within the category, and too broad, allowing words that intuitively should fall outside. Usually, this leads to intensive debates in the literature, which often interacts with the literature on universals.  For a canonical case, the expression ``only'' has been argued to be a determiner that lacks a key universal (conservativity); one response, however, has been to argue that it is not in fact a determiner, and so does not belong to the kind of expression in the universal (see Section~\ref{sec:conservativity-learning} for more details and the relevant literature). 

Hence, although we can usually be confident to a certain extent in the reality of posited semantic categories, one must be cautious and take the challenge of justifying the categorial borders seriously. Although the example of determiners defines the kind $K$ in syntactic terms, many universals that we will discuss specify the relevant kind of expression semantically. When dealing with (more or less abstract) content word domains, e.g., numbers, colors, or containers, one can use criteria about the kinds of entities referred to by the expressions to restrict the category. All of these expressions are nouns; while there may be fine-grained syntactic criteria to subcategorize the nominal domain, the intuition behind these kinds of expressions is that they differ semantically.  Whatever semantic criterion one provides for defining these kinds, it must be applicable cross-linguistically in order to be able to identify the relevant expressions. 

While demarcating the category borders by referring to the properties of the objects, there is a relatively low risk of conflating this discussion with the issue of formulating semantic universals. However, the whole task may become much less trivial in the case of functional vocabulary, where the corresponding meaning space is much harder to grasp, criteria more challenging to apply, and often the same properties may be posited as both the category demarcation ($K$) and universal constraint ($P$). Certain universal constraints will be accurate or false depending on how we draw the demarcation line. For instance, the status of conservativity for quantifiers depends on whether we count ``only'' as belonging to the syntactic category of determiners or not (see Section~\ref{sec:conservativity-learning} for more discussion).

We note also that some universals may apply to very broad categories of expression.  For example, the convexity universal for color terms (see Section~\ref{sec:convexity-learning}) has been argued to be an instance of a more general, cross-categorial preference for convex meanings \citep{gardenforsConceptualSpacesGeometry2000,gardenforsGeometryMeaning2014,chemlaConnectingContentLogical2019}.  In cases like that, $K$ may be defined disjunctively, or may in fact just be the set of all expressions.

\paragraph{Meaning space}

The remaining components of the \textsc{Lexical Property Universal Schema} to be specified are (i) the space of all possible meanings that expressions can have and (ii) the subset of all of those satisfying the property $P$.  We will refer to (i) as `the meaning space' for each example.
Thus, computational modeling of a semantic system presupposes that we understand what the meaning space consists of. Whenever we discuss a category of expressions, we will be explicit about its corresponding meaning space and the mapping between particular forms and meanings. However, as it happens in science,  in doing so, we will often abstract away from many subtleties as long as we believe that such simplifications are justified, i.e., they increase the explanatory power of the model without a significant reduction in its descriptive coverage.

However, the issue of defining the meaning space only sometimes boils down to innocent assumptions. The question about logically possible languages/meanings/denotations is central to the whole research endeavor of formulating and explaining linguistic universals. A fruitful way of thinking about universals is as restrictions on the logical space of all possibilities. Whether or not the universal is substantial depends on that logical space of options.\footnote{We thank Fausto Carcassi for discussion of this point.} Intuitively, substantial universals restrict the space significantly. In other words, there exist (many) logically possible meanings that are excluded by the universal. Hence, whether one should count some generalization as a substantial universal depends on what one takes the space of possible meanings to be. In some cases, defining the space of possible meanings is relatively straightforward, e.g., in the case of color terms discussed in Section~\ref{sec:convexity-learning}, we use a cognitively-motivated space of color derived from multiple experiments. In other cases, we may not have access to a well-grounded mental representation corresponding to the phenomenon, but there exists a widespread linguistic consensus on how the meanings should be modeled. For instance, while thinking about universal constraints in the domain of quantification, it is generally agreed that the space of possible meanings can be defined by all functions of type $\langle et, \langle et, t \rangle \rangle$ (i.e.\ as discussed in Section~\ref{sec:monotonicity-definition}). Under such a general definition, constraints like monotonicity become substantial. However, there exist alternative definitions of the generalized quantifier space in the literature that presuppose certain restrictions which make other universals nonsubstantial (e.g. \citealp{knowltonPsycholinguisticEvidenceRestricted2023}; we return to this in Section~\ref{sec:conservativity-learning}). Yet, in some other cases, e.g., indefinite pronouns (Section~\ref{sec:indefinites}), more research into the best way to capture the set of possible meanings may be necessary. 

In general, defining the meaning space requires detailed knowledge of the mental representation of the domain in question.  While we do not necessarily have a complete picture of these representations in every case, the models that we develop in this Element are modular in a certain sense: when there is progress in the scientific discussion about how best to capture a particular meaning space, then the model and the corresponding experiments can and should be re-done. The same holds one level up: our explanations compare attested languages with a space of possible but unattested ones, and how that space is constructed is equally a substantive choice (we return to this in Chapter~\ref{chapter:discussion}).

\paragraph{Optimization Principle}

Having specified the kinds of expressions and the broad meaning space, the final ingredient in our explanations will be an \emph{optimization principle}.  Recall that lexical property universals claim that all lexicalized meanings have a certain property.  The general idea behind our explanations will be to argue that meanings with the universal property are (more) optimal in a well-motivated sense.  To the extent that the sense of optimality can be shown to be at play in language change and emergence, this would explain why we find such universals.  To the extent that the same principle(s) can be used to explain many different semantic universals, we will have more credence in that principle.  In the next two sections, we explain two optimization principles---ease of learning and efficient communication---which will be illustrated with several case studies each in Chapters~\ref{chapter:learning} and \ref{chapter:effcomm} respectively.

\subsection{Ease of Learning}
\label{sec:intro-learning}

One of the most intuitive answers to why languages are the way they are goes through learning. In short, natural languages are of specific forms because they are constrained by what humans can learn or, in our case, what things are easier to learn.  In this section, we spell out learning as an optimization principle, and explicate how it can apply to lexical property universals.  We do so by first explicating a conception of learning which does not go far enough, and then enriching that one.

\subsubsection{Ease of learning as fencing-in}

One can attempt to justify semantic universals by observing that they greatly restrict the search space a language learner must explore when learning the meaning of natural language expressions. For instance, if the possible meanings of determiners are limited to monotone quantifiers, children do not even need to consider non-monotone hypotheses when learning the meaning of a new determiner. Such restriction of the initial space of hypotheses, in principle, should make it easier (or even possible in the first place) for children to learn such meanings from relatively small inputs. Note that this explanation is reminiscent of the poverty of stimulus argument for Universal Grammar \citep{chomskyAspectsTheorySyntax1965, petersQuantifiersLanguageLogic2006, szabolcsiQuantification2010}. 

We call this a fencing-in argument because one can visualize it as follows: the space of all possible meanings to be learned is a region and a universal constitutes a fence, providing a border beyond which the learner cannot wander.
While this argument is valid (learning in a smaller hypothesis space must be easier), we do not find it satisfactory, for two main reasons. 

First, it may happen that a universal constraint is not helpful for the learner. Even though it restricts the learning space, it may only eliminate hypotheses that, for a learner, would be easy to rule out and leave in the hypotheses that are hard to distinguish based on the available evidence. In fact, in a Bayesian simulation study, \citet{piantadosiModelingAcquisitionQuantifier2013} show that it may be very well the case for conservativity universal in the domain of quantification. Their computational experiment showed that the problem of learning the meaning of determiners does not become significantly more manageable when the initial learning space is restricted to only conservative quantifiers.   

Second, the fencing-in argument does not explain \emph{which} universals are attested. Every restriction of the learning space (i.e.\ every property of meanings) would constitute a fence and thus make the learning problem easier. Indeed, many restrictions make the situation easier but do not correspond to any known linguistic universals. We do not think that such a learnability notion is explanatory for the problem at hand, and instead, we propose an alternative learnability argument. 

\subsubsection{Ease of learning as temperature}

\begin{figure}[ht!]
    \centering
    \includegraphics[width=\textwidth]{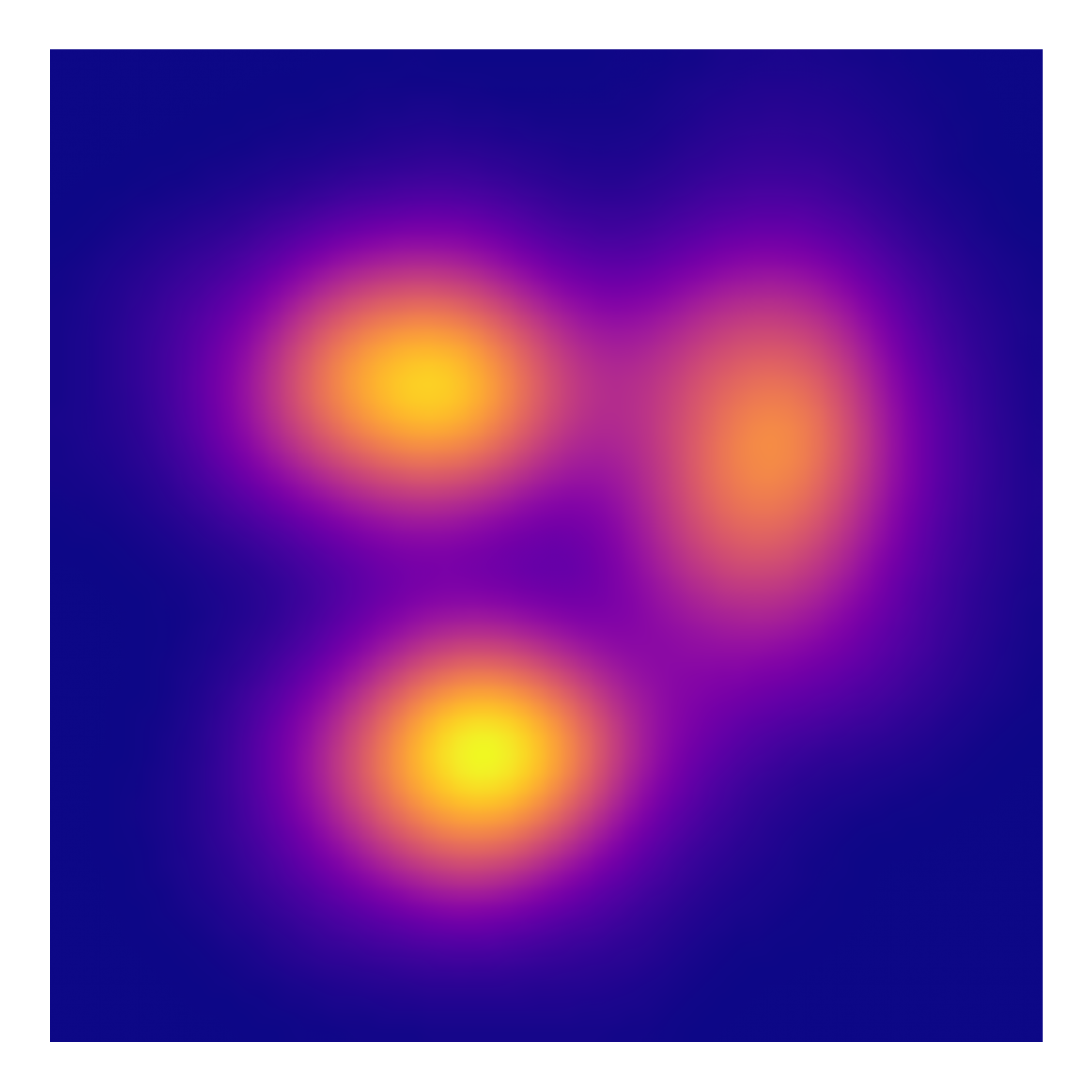}
    \caption{A hypothetical ``heatmap''.  The points in the space correspond to possible meanings (e.g.~of quantifiers) and the color corresponds to ease of learning: yellow is `hot' or easy to learn, and blue is `cold' or hard to learn.  On the \textsc{Ease of Learning Hypothesis}, one expects to find meanings with universal properties in more yellow regions. On the fencing-in picture, a universal would be a sharp boundary drawn across this map; on the temperature picture, it is a region where the map is warm.}
    \label{fig:heatmap}
 \end{figure}

A better way to think about learnability in the context of explaining universal linguistic properties is via learning simplicity. Specific universals aid acquisition because expressions satisfying them are easier to learn. One can imagine, for instance, a space of many possible quantifier meanings. Some of those meanings are `hot', i.e., they are compatible with human learning mechanisms and, therefore, are asking to be lexicalized. Other potential meanings would be much `colder', i.e.\ harder to learn; hence, one shouldn't expect to see them universally lexicalized.  We can make the idea explicit, as visualized in Figure~\ref{fig:heatmap}:

\begin{itemize}[align=left, leftmargin=*]
	\item[\textsc{The Ease of Learning Hypothesis:}] Lexical property universals arise because meanings satisfying the relevant properties are easier to learn than those that do not. 
 \end{itemize}

Most of our explanations in this book will consist in showing that this hypothesis holds true for many lexical property universals: meanings with the relevant property $P$ are in fact easier to learn than those that do not.  For the time being, we will be assuming that languages tend to lexicalize the easier-to-learn meanings.  We will return to this assumption in Section~\ref{sec:iterated-learning}, where we discuss what's known as the linkage problem \citep{kirbyFunctionSelectionInnateness1999} and develop a model for monotone quantifiers in terms of iterated learning \citep{kirbyIteratedLearningEvolution2014}. 

Note that such an approach to ease of learning explains \emph{which} universals should be attested, unlike the fencing-in picture. It is also compatible with the statistical view on linguistic universals. It predicts strong biases toward `hot' meanings but does not rule out that there might be linguistic or extra-linguistic pressures that make other meanings emerge in some exceptional circumstances. Finally, this explanatory approach relies on compatibility between the inherent conceptual structure of possible meaning and the specific learning mechanism, as it's \emph{a priori} conceivable that different learning mechanisms can promote different meanings. Chapter~\ref{chapter:learning} shows that we can explain the presence of semantic universals under some instantiations of learning mechanisms.

\subsection{Efficient Communication}
\label{sec:intro-communication}

Another prominent style of explanation that we will apply in several cases argues that languages are optimized for efficient communication (see \citealp{kempSemanticTypologyEfficient2018} for a survey; many more examples and references will be provided in Chapter~\ref{chapter:effcomm}).  Typical explanations in this tradition argue that languages can be viewed as (nearly) optimally balancing two competing pressures: to be simple and to be useful.

To be simple means to be easy to represent in the mind of the language user.  While we will postpone operationalizing and formally defining this concept until Chapter~\ref{chapter:effcomm}, one can intuitively imagine that a language with a very small lexicon (e.g.\ a single expression) will be very simple, while a language with an extremely large lexicon will be more complex, since the latter requires more memorization than the former.  As a terminological note, complexity is used as the inverse of simplicity, so that maximizing the latter is equivalent to minimizing the former.  On that dimension, then, one principle is: (\emph{ceteris paribus}) languages strive to minimize complexity.

To be useful means to enable speakers of the language to reliably convey their thoughts to one another.  While we once again postpone formal definition, the intuition goes as follows: a speaker has a thought that they wish to convey to a listener (i.e.\ a point in meaning space, as discussed above), using expressions in the language.  If the language has a unique expression for every such point, then such communication will be perfectly successful: the listener will be able to infer the exact point that the speaker had in mind.  As a terminological note, we will use the term \emph{communicative cost} as a measure of the inability of listener to do this recovery. On this dimension, one principle is: (\emph{ceteris paribus}) languages strive to minimize communicative cost.

Unfortunately, these two principles are in tension with each other: it is impossible for a language to fully minimize complexity while simultaneously fully minimizing communicative cost.  For example, a language with a single expression would be, as mentioned above, very simple; it would, however, have a very high communicative cost, since speakers of such a language have to use the same expression to attempt to convey every point and thus cannot make any distinctions.  By contrast, a highly complex language with many expressions for unique points would have a low communicative cost, but rather high complexity.  

Because complexity and communicative cost cannot be jointly minimized, we will want to suggest that languages are ``doing the best they can'' at balancing these two competing pressures.  To make this precise, we can borrow a technical term from the literature on multi-objective optimization \citep{emmerichTutorialMultiobjectiveOptimization2018}, of which the present is an instance: the Pareto frontier is the set of languages for which no other language is \emph{both} less complex and less costly.  In other words, this frontier is the languages which can only become simpler by becoming less useful or can only become more useful by becoming more complex.  With this in hand, we can formulate:

\begin{itemize}[align=left, leftmargin=*]
	\item[\textsc{The Efficient Communication Hypothesis:}] Languages that optimally balance complexity and communicative cost (in the sense of lying on or near the Pareto frontier) have expressions which satisfy semantic universals. 
 \end{itemize}

\begin{figure}
    \centering
    \includegraphics[width=\linewidth]{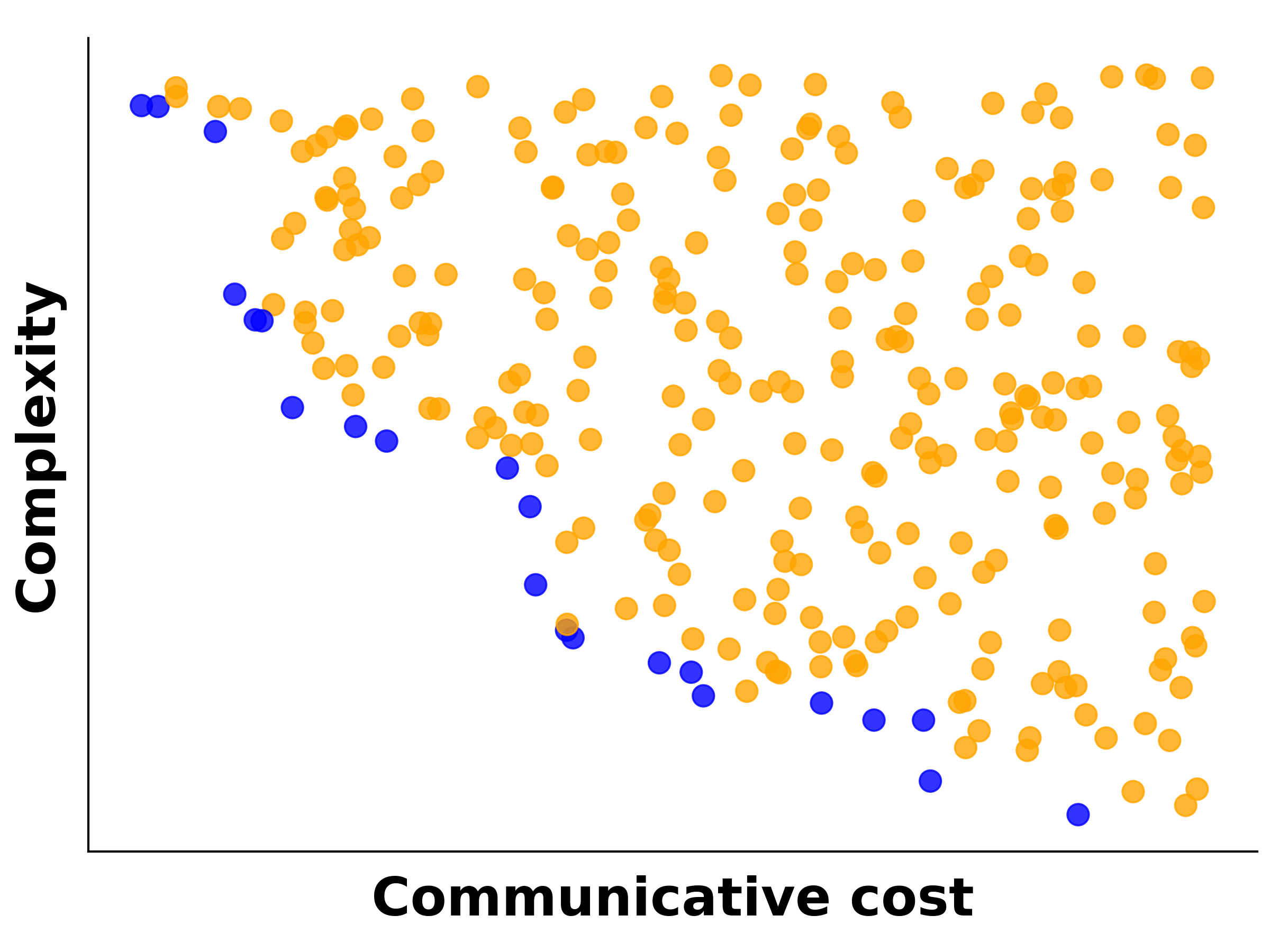}
    \caption{Hypothetical data illustrating the \textsc{Efficient Communication Hypothesis}.  Each dot corresponds to a language, plotted in the space of complexity and communicative cost.  The lack of languages in the lower-left represents the fact that the two properties cannot be jointly optimized.  The blue dots are the Pareto frontier, the set of languages doing as best they can at balancing the two pressures.}
    \label{fig:effcomm-hypothetical}
\end{figure}

 We illustrate this hypothesis with hypothetical data in Figure~\ref{fig:effcomm-hypothetical}.  In this plot, the $x$-axis represents communicative cost, and the $y$-axis complexity.  Each dot corresponds to a language, either an actual natural language or a hypothetical artificial language.  The lack of languages (dots) in the lower-left region of the figure captures the fact that complexity and communicative cost cannot both be jointly minimized.  The blue dots are the Pareto frontier: no language is both simpler and more useful than any of these (there are no dots down \emph{and} left of each of them).  The \textsc{Hypothesis}, then, is that all of these languages will satisfy semantic universals of interest.

\subsection{A Note on Universals and Typology}

We here want to reiterate a point that has been implicit in the discussions of the previous section (e.g.\ in Sections~\ref{sec:degrees-universals}): while the targets of our explanations will always be (alleged) semantic \emph{universals}, we see such universals as the far end of a spectrum of possible variation.  The style of explanations that we offer in this Element---in terms of ease of learning and efficient communication---in effect show such universals as \emph{arising} or \emph{emerging} from more fundamental forces that shape the semantic structure of language.

We take the default position to be that as many possible properties of language structure should be explained in styles like these (though of course there are many other pressures beyond those discussed here, including historical and social factors).  Only in the case where it appears that no such explanation can be offered should a proposed universal be attributed to Universal Grammar or to the architecture of grammar more generally, as we will suggest for conservativity (Section~\ref{sec:conservativity-learning}).

\subsection{Plausibility \& robustness}

The results reported in this element depend on the details of the experimental setups. Above, we have discussed some major issues related to defining the phenomenon in question. Now it's time to point out that our simulations unsurprisingly come with many assumptions, too. We will make those assumptions explicit when we dive into the specific studies but let us foreshadow some issues already here. There are two main questions we need to pay attention to:
Are the models cognitively plausible? 
To what extent are the results robust? 

Let us start with the issue of cognitive plausibility. The models discussed in this element will typically evoke precisely defined mathematical notions or measures whose role is to represent some semantic or mental construct. Examples are abundant: logical formalizations are used to capture the meaning of expressions, machine learning algorithms are used to describe the process of language learning, information-theoretic measures serve to approximate the cognitive difficulty of mental concepts, etc. Representing cognitive constructs with explicit computational formalism is the essence of modeling. By doing that, we gain insights into possible underlying information processing mechanisms responsible for the existence of the specific universals in language. Also, computational formalizations force us to remain transparent and explicit: every aspect of the model should be controlled and justified. As a result, such models lead to natural questions of the cognitive plausibility of various parts of the presented modeling efforts. Discussions like this are one of the main benefits of the modeling efforts as they lead us to a deeper understanding of the problem and constant improvements of the computational explanations.

On the other hand, practically every single model makes assumptions that could be more evidently cognitively plausible. In what follows, we will do our best to highlight those assumptions to inspire the discussion of how the models could be improved. However, two remarks are in order here. First, many models presented in this book aim to demonstrate that something is, in principle, possible. For instance, in some of the chapters (much of Chapter~\ref{chapter:learning}), by using artificial neural networks as learning models, we show that, in principle, some of the universals may result from implicit learning biases. It remains to be established whether such ANNs capture well the learning biases present in animals; however, the presented simulations add evidence to such an explanation. It is, in principle, a reasonable explanation for the observed typological phenomena.  

Such explanations gain strength when they become robust in at least two senses of the word: First, if the same explanation works across different linguistic domains for related phenomena. For instance, the learnability explanation formulated via artificial neural network models applies to various unrelated semantic universals. Second, the explanation is robust if it is indifferent to as many model changes as possible, e.g., the learnability explanation gets more powerful, and hence convincing, every time it gets replicated on a different computational architecture, with the different values of parameters, etc. The more robust it is in this sense, the more likely it is that the critical explanatory mechanism does not depend on the specific architecture but is something at least very common to many learning processes, including learning in humans and animals.

\subsection{Summary}

\begin{table}[th]
    \centering
    \resizebox{\linewidth}{!}{%
    \begin{tabular}{ccccc}
        \toprule
         Expression type & Meaning space & Property & Optimization principle & Section \\
         \midrule
         Determiners & Generalized quantifiers & Monotonicity & Ease of learning & \ref{sec:monotonicity-learning} \\
         Determiners & Generalized quantifiers & Quantity & Ease of learning & \ref{sec:quantity-learning} \\
         Color Terms & CIELab color space & Convexity & Ease of learning & \ref{sec:convexity-learning} \\
         Responsive predicates & Possible-world meanings & Veridical uniformity & Ease of learning & \ref{sec:responsive-predicates} \\
         Determiners & Generalized quantifiers & Conservativity & Ease of learning (mixed)\textsuperscript{*} & \ref{sec:conservativity-learning} \\
         Determiners & Generalized quantifiers & Monotonicity & Efficient communication & \ref{sec:quantifiers-effcomm} \\
         Indefinite pronouns & Semantic feature space & Convexity & Efficient communication & \ref{sec:indefinites} \\
         Modals & Force-flavor pairs & Independence of Force and Flavor (IFF) & Efficient communication & \ref{sec:modals} \\
         \bottomrule
    \end{tabular}%
    }
    \caption{A summary of the computational explanations provided in the next two chapters, organized by the parameters of the \textsc{Lexical Property Universal Schema} and the optimization principle used. The section numbers refer to the chapters in which the explanations are provided. \textsuperscript{*}No advantage in neural networks; an advantage in human learners only when the syntactic frame forces a determiner interpretation. Section~\ref{sec:conservativity-learning} argues that conservativity may not be a lexical property universal at all.}
    \label{tab:explanation-outline}
\end{table}

To summarize: this book focuses on providing computational explanations of what we are calling lexical property universals.  After introducing the concept and a motivating example, we then discussed what components must be specified to provide an explanation, paying special attention to two optimization principles that lie at the core of our explanations: \textsc{The Ease of Learning Hypothesis} and \textsc{The Efficient Communication Hypothesis}.  The next two sections develop explanations for several lexical property universals using each of these hypotheses.  The interested reader can consult Table~\ref{tab:explanation-outline} for a full list of the phenomena explained.  In addition to these core explanations, each chapter summarizes other approaches and experiments with human participants.  After these two chapters, the Element concludes with an extended discussion of what we have learned from the explanations, and where the next frontiers for pushing the project of explaining semantic universals forward lie.

\pagebreak

\section{Ease of Learning}
\label{chapter:learning}

As discussed in Section~\ref{sec:intro-learning}, the first optimization criterion that we will use for developing explanations of semantic universals concerns ease of learning, as codified in \textsc{The Ease of Learning Hypothesis}: Lexical property universals arise because meanings satisfying the relevant properties are easier to learn than those that do not.
This chapter will provide support for this hypothesis by developing several case studies.  After first providing a high-level introduction to how we will model semantic learning in general (\S~\ref{sec:supervised-learning}), we will discuss in detail the monotonicity of quantifiers, showing that the degree of monotonicity is correlated with ease of learning (\S~\ref{sec:monotonicity-learning}). We will then apply the same explanatory framework to convexity in color terms and beyond (\S~\ref{sec:convexity-learning}), quantity for quantifiers (\S~\ref{sec:quantity-learning}), and the veridicality of responsive predicates (\S~\ref{sec:responsive-predicates}). Finally, we will discuss a possible counter-example to \textsc{The Ease of Learning Hypothesis}, the conservativity of quantifiers (\S~\ref{sec:conservativity-learning}), before a more general discussion of the ease of learning hypothesis and directions forward (\S~\ref{sec:learning-discussion}).

\subsection{Supervised Learning in Humans and Machines}
\label{sec:supervised-learning}

To measure the ease of learning of various lexical properties, we will use a supervised learning paradigm. In supervised learning, a learning agent receives input scenarios and produces guesses as an output. For example, when learning the meaning of a new determiner, the learner will receive a sentence with that determiner, e.g., ``Gleeb of the squares are red,'' together with a situation, e.g., a display with geometrical figures of different colors. The learner guesses whether the sentence is true or false in the given situation. The learning is supervised because, after each guess of the learner, the supervisor provides feedback, e.g., saying whether the guess was correct.  This feedback is used to update the learner's internal guess at the meaning of the determiner, making it more likely to guess correctly in the future.

To instantiate this paradigm in any particular example, one has to at least specify the learner(s), the scenarios, and how the feedback is provided.  In many cases, one can understand traditional word learning experiments with human participants (e.g.~\citealp{mcmillanNeuralBasisGeneralized2005,hunterConservativityLearnabilityDeterminers2013,chemlaConnectingContentLogical2019}) as falling under this paradigm.  Although we will discuss particular experiments of this type when they pertain to specific semantic universals in the sections to follow, we here record some limitations with the methodology of measuring ease of learning with artificiial word learning in human participants.

A first issue arises with the scale of such experiments.  In these experiments, simply due to time and memory limitations, participants will usually only be exposed to tens of examples of the novel word meaning and evaluated on a similarly small number.  It is unclear, then, to what extent the target meaning has truly been learned or not in these scenarios.  A more fundamental issue arises, however, when using such experiments to attempt to explain semantic universals: if these properties are in fact universal, then learners who have already acquired a language may have a bias towards meanings with these properties.  For example, when learning a new determiner ``gleeb'', human participants may recognize that it has the same syntactic type as words like ``every'' and ``most'', and therefore (implicitly) assume that its meaning must be similar as well.  Since all of those meanings satisfy relevant universals (e.g.~monotonicity), the participant may assume that ``gleeb'' will also satisfy those universals.  Results that show that a particular universal property is easier to learn may reflect a bias from the participants' existing linguistic knowledge rather than a true difference in ease of learning of the meanings.\footnote{Some experiments attempt to measure these ease of learning differences in earlier stages of learning, i.e.\ in infants or non-human animals \citep{hunterConservativityLearnabilityDeterminers2013,chemlaConstraintsLexiconsHuman2019}.  While these are valuable experiments, they suffer even more seriously from limitations in scale.}

Because of these issues, the main measurement device in this chapter is artificial neural networks. They can be trained and evaluated on far more examples than human participants, which helps ensure that what they learn is indeed the target meaning. Because we train them from scratch, they have no biases arising from earlier learning of similar meanings. Human learning experiments remain essential, however: they tell us whether a difference found in networks also shows up in human cognition, and we report them alongside the computational results wherever they exist. Recent large-scale experiments also go some way toward addressing the worry about prior linguistic knowledge: \cite{RAMOTOWSKA2026106523} found that many adult participants did try to map a novel determiner onto familiar ones, but the learning advantage for meanings satisfying universals held also for participants who learned implicitly or who mapped the word onto the wrong familiar determiner (see \S\ref{sec:convexity-quantifiers}, \S\ref{sec:quantity-learning}, \S\ref{sec:conservativity-learning}).

Seen in light of overcoming some limitations of human artificial word learning experiments, and with the heat map metaphor from \S\ref{sec:intro-learning} in mind, we view the use of artificial neural networks as akin to the use of a thermometer \citep{steinert-threlkeldCognitiveThermometersMachine2026}: they are a measurement device that lets us produce a more fine-grained measure of ease of learning than one can produce in traditional word learning experiments.

\subsection{Monotonicity}\label{sec:monotonicity-learning}

Our first ease of learning case study will focus on the motivating example from the introductory chapter: the monotonicity of quantifiers. Recall from Section~\ref{sec:monotonicity-definition} that a lexical property universal concerning monotonicity states that: all monomorphemic determiners are monotone.  Support for \textsc{The Ease of Learning Hypothesis} in this case would come from showing that quantifiers that are more monotone are easier to learn than those that are less monotone.  In this section, we first review the computational experiments that have been conducted to test this hypothesis, before focusing on a study generalizing the above to degree of monotonicity (\S~\ref{sec:degree-monotonicity-learning}), and another one linking this learning bias to lexicalization explicitly (\S~\ref{sec:iterated-learning}).

In the spirit of the supervised learning approach described above, \citet{steinert-threlkeldLearnabilitySemanticUniversals2019} trained recurrent neural networks (long-short term memory networks---LSTMs---in particular) to learn the meaning of novel quantifiers.  Taking inspiration from sentence-picture verification tasks in psycholinguistics \citep{robertsSentencepictureVerficationTask1994,hunterConservativityLearnabilityDeterminers2013,szymanikComprehensionSimpleQuantifiers2010}, the authors presented the network with a novel determiner sentence (e.g.\ ``gleeb of the dots are red'') and a situation (e.g.\ a sequence of objects with various properties).  The model would predict whether the sentence was true or false in that scenario, and receive feedback in order to update its internal structure to better make future predictions.  We refer the reader to \citet{steinert-threlkeldLearnabilitySemanticUniversals2019} for full details of the experimental setup, including model hyperparameters and input details.  We note here that the model architecture chosen has no \emph{a priori} known bias in favor of monotone quantifiers, i.e.\ it was not designed in such a way as to favor such quantifiers.

\begin{figure}[ht!]
    \centering
    \includegraphics[width=0.95\textwidth]{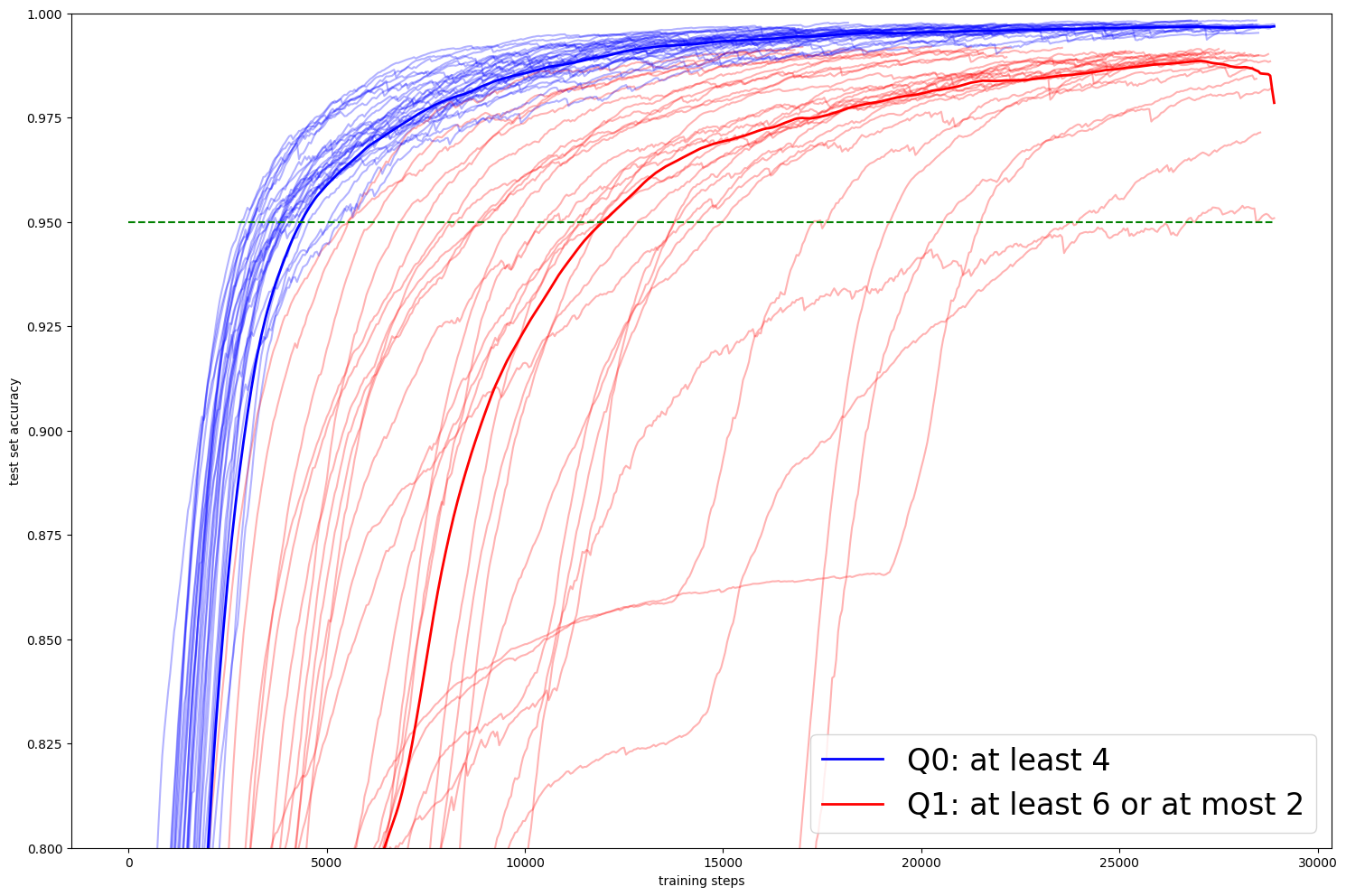}
    \caption{Learning curves for an upward monotonicity experiment: \textsf{at least 4} (blue) compared to \textsf{at least 6 or at most 2} (red).  The $x$-axis corresponds to the number of training steps and the $y$-axis to accuracy on a \emph{test set}: examples of quantifier models that have not been seen during training.}
    \addsource{\cite{steinert-threlkeldLearnabilitySemanticUniversals2019}}
    \label{fig:learning-monotonicity-upward}
\end{figure}

\citet{steinert-threlkeldLearnabilitySemanticUniversals2019} adopted a minimal pair methodology: the model would learn two quantifiers at once which were very similar to each other, but which differed in their monotonicity.  A first experiment compared \textsf{at least 4}---an upward monotone quantifier---with \textsf{at least 6 or at most 2}---a paradigmatically non-monotone quantifier.  Note that the model did not see these expressions, but rather new labels of equal grammatical complexity (e.g.\ ``gleeb'' and ``glerb'').  The authors ran thirty learning trials, and recorded how quickly the model reached and maintain a 95\% accuracy threshold on a test set of examples that it had not seen during training.  The results of this experiment are shown in Figure~\ref{fig:learning-monotonicity-upward}.

As you can see qualitatively in Figure~\ref{fig:learning-monotonicity-upward}, the learning regularly converged faster for the monotone (blue lines) than non-monotone quantifier (red lines). Quantitative analysis supports this conclusion.  This computational experimental result provides initial support for \textsc{The Ease of Learning Hypothesis} for the monotonicity universal.

\begin{figure}[ht!]
    \includegraphics[width=0.95\textwidth]{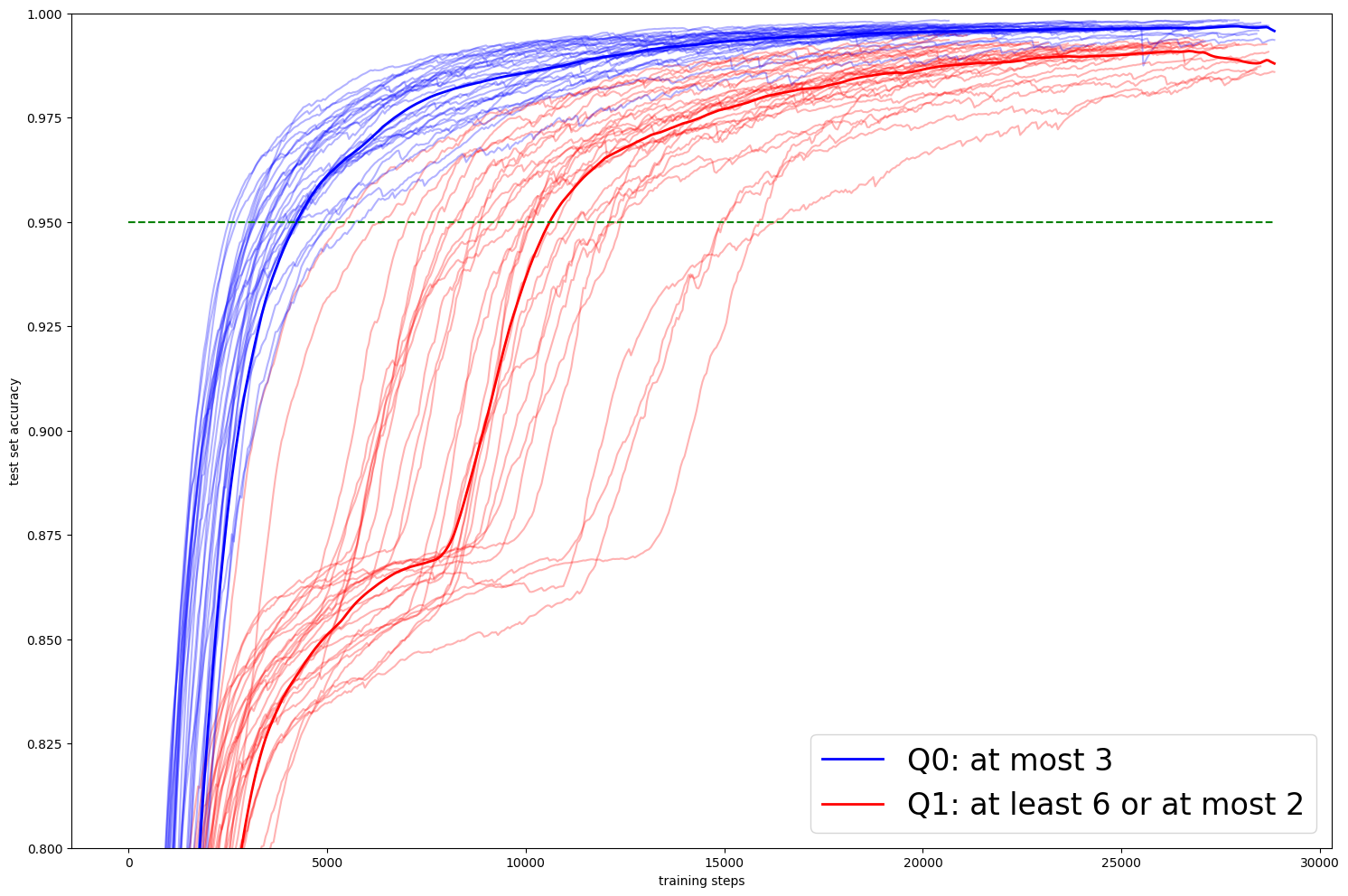}
	\caption{Learning curves for an downward monotonicity experiment: \textsf{at most 3} (blue) compared to \textsf{at least 6 or at most 2} (red).  See Figure~\ref{fig:learning-monotonicity-upward} for information on the axes.}
    \addsource{\cite{steinert-threlkeldLearnabilitySemanticUniversals2019}}
    \label{fig:learning-monotonicity-downward}
\end{figure}

In a second minimal pair learning experiment, \citeauthor{steinert-threlkeldLearnabilitySemanticUniversals2019} replaced \textsf{at least 4} with \textsf{at most 3}, which is a paradigmatically \emph{downward} monotone quantifier.  The results of this experiment are shown in Figure~\ref{fig:learning-monotonicity-downward}.  As in the first experiment, the model learned the monotone quantifier more quickly than the non-monotone quantifier.  This suggests that the ease of learning bias towards monotonicity is not specific to upward monotonicity, but rather a general property of both upward and downward monotonicity.  In terms of the heatmap metaphor: this suggests that the monotone quantifiers do occupy a warm region in the sapce of possible quantifiers.

\subsubsection{Learning and the degree of monotonicity}
\label{sec:degree-monotonicity-learning}

While these initial experiments provide compelling initial support for \textsc{The Ease of Learning Hypothesis}, the minimal pair methodology suffers from a limitation: although the quantifiers were chosen to be as `similar' as possible to each other, except for monotonicity, we do not know if they are fully representative of the space of monotone quantifiers.  To truly test ease of learning as an explanation for monotonicity, one must scale these experiments to a larger set of possible quantifiers.

\begin{figure}[ht]
    \includegraphics[width=\textwidth]{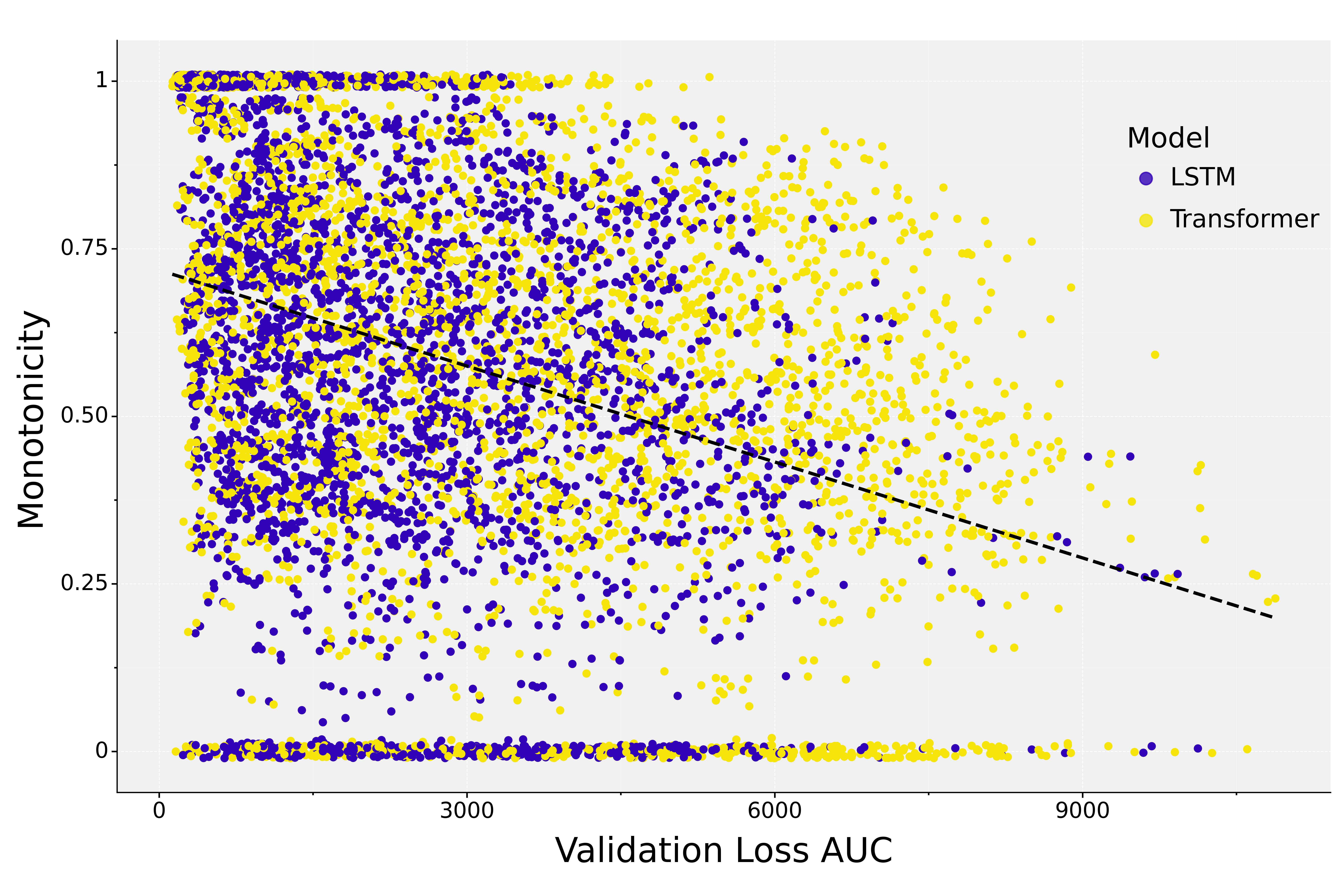}
    \caption{Results from \cite{haberlandQuantifiersThatAre2025}.  Each dot corresponds to one quantifier. The $x$-axis is the area under the learning curves; these are computed with respect to a loss, so lower values mean easier to learn.  The $y$-axis corresponds to the degree of monotonicity of each quantifier.  The line of best fit, in red, shows that there is a correlation between ease of learning and degree of monotonicity.}
    \label{fig:monotonicity-all}
\end{figure}

\citet{haberlandQuantifiersThatAre2025} conducted such an experiment, adopting the same learning protocol as above, but applying it to a much wider range of quantifiers.  In particular, they generated a very wide range of quantifiers using a grammar for generating mathematical expressions for quantifier meanings.  (We will say more about this general approach to meaning representation and its use as a measure of complexity in \S\ref{sec:complexity-cost} and beyond.)  For each quantifier, they measured ease of learning by the area under loss curves for a fixed amount of learning (so lower is better), and compared these to the quantifiers' degrees of monotonicity.
Figure~\ref{fig:monotonicity-all} shows a clear negative relationship between the two: as quantifiers become less monotone, they also become harder to learn. This suggests that the ease of learning of quantifiers is not just a binary property (monotone vs.\ non-monotone), but rather a continuous one, with more monotone quantifiers being easier to learn than less monotone ones.  In terms of our heatmap metaphor: this suggests, more generally, that a heatmap based on degree of monotonicity would look somewhat similar to what generated based on ease of learning.

The grammar used to generate these quantifiers, implemented in the Unnatural Language Toolkit \citep{imelUnnaturalLanguageToolKit2025}, takes its set-theoretic and Boolean primitives from \citet{polQuantifiersSatisfyingSemantic2023}. That study raises a serious alternative to the ease of learning explanation. Generating tens of thousands of quantifiers in this kind of language of thought, van de Pol et al. found that quantifiers satisfying each of monotonicity, quantity, and conservativity have, on average, a shorter minimal description length: perhaps universals hold because the relevant meanings are simple to represent, not because they are easy to learn. The results of \cite{haberlandQuantifiersThatAre2025} suggest that the two can come apart: a follow-up regression analysis of their learning data shows that ease of learning is a better predictor of degree of monotonicity than description length and that learning explains additional variance in monotonicity over-and-above what description length can explain \citep{steinert-threlkeldCognitiveThermometersMachine2026}. Nevertheless, the two explanations are closely linked. The difficulty neural networks have in learning Boolean concepts tracks their logical complexity \citep{carcassiNeuralNetworksTrack2022}, and conversely, the primitives of a Boolean language of thought can be recovered from human learning data \citep{CARCASSI2023105541}. We return to representational complexity in Chapter 3, where it supplies the complexity term of efficient communication analyses (§3.1), and to the relation between learnability and description length in §4.4.

\subsubsection{Iterated learning and monotonicity}
\label{sec:iterated-learning}

At this point, one may ask how to connect the individual learning bias towards monotonicity, established in the previous experiments, to the population level. After all, the universal is not about the learning preferences of individuals but about the structure of languages themselves. This issue is sometimes called in the literature the problem of linkage \citep{kirbyFunctionSelectionInnateness1999}, and a standard solution is to show how structural properties of language emerge in cultural evolution from even slight individual learning biases.  \emph{Iterated learning} has emerged as a prominent modeling framework for this purpose \citep{kirbyIteratedLearningEvolution2014,smithHowCultureBiology2020}.  This framework models intergenerational transmission: a generation of agents learns from its `parents' and then goes on to teach its own `offspring'.  Assuming, for example, that each generation cannot learn perfectly from its parents---for instance, due to cognitive constraints or limited exposure---this process can enable individual biases to shape the learned behavior over time, as each generation may introduce small changes.

\citet{carcassiMonotoneQuantifiersEmerge2021b} use iterated learning to show how artificial neural network bias towards monotonicity may lead to the emergence of monotone languages.  The iterated learning model consists of 300 generations, each with ten simple artificial neural network agents. The earlier generation agents---the cultural parents---are associated with one or more agents in the following generation---their cultural children. As in the experiments from the previous subsections, the cultural parents are trained on a novel determiner meaning. They are given the determiner and the encodings of finite situations and must decide whether the determiner is true in that situation. Every generation gets a fixed number of situation to learn from, the so-called bottleneck size. In the next step, the parents' judgments are used as input for the next generation, cultural children. Based on these data, each cultural child tries to approximate its parent's determiner. Once the agents in the second generation have each learned a determiner, a third generation is created, and the process is repeated three hundred times.  

The model resembles the game of telephone. The agents do not learn their cultural parent's determiner perfectly. There may be many reasons for this, but the bottleneck size controls one: the child can see only a fragment of the language data points, which may not be enough to reconstruct the language perfectly. As in the children's game,  the changes introduced by each child accumulate. However, the critical idea of iterated learning is that those changes are not random but reflect the biases of individual cognitive systems.

\begin{figure}[ht!]
	\centering
	\includegraphics[width=\textwidth]{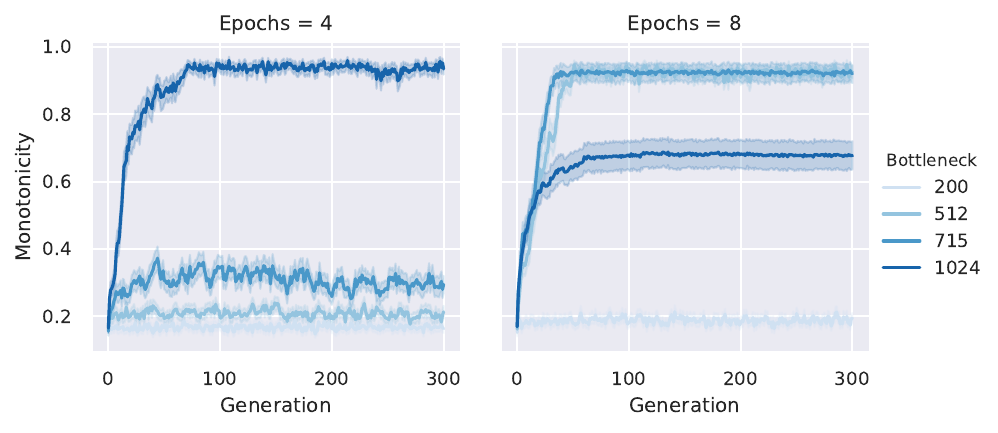}
	\caption{The degree of monotonicity of quantifiers in the iterated learning model. The $x$-axis corresponds to the generation of iterated learning and the $y$-axis to the degree of monotonicity. The different lines correspond to different bottleneck sizes. Across many settings of numer of epochs (how many times the learner was exposed to the examples) and bottleneck sizes, the degree of monotonicity increases to nearly 1 over generations.}
	\label{fig:iterated-learning-monotonicity}
	\addsource{\cite{carcassiMonotoneQuantifiersEmerge2021b}}
\end{figure}

\citet{carcassiMonotoneQuantifiersEmerge2021b} set out to prove that among those biases is the tendency towards monotonicity. They start by generating random quantifiers and showing that such quantifiers are extremely unlikely to have a high degree of monotonicity. In other words, monotonicity is a rare property that should not be expected to appear by chance. Next, the authors initiate the iterated learning chains with randomly generated quantifiers,  varying the bottleneck size and the number of epochs. As Figure~\ref{fig:iterated-learning-monotonicity} illustrates, monotone quantifiers evolve consistently for some values of the simulation parameters, controlling how much of the cultural parent's language is observed by the cultural child. 

This result enriches the computational explanation for the monotonicity universal.  Having previously demonstrated an individual learning bias for monotonicity, this iterated learning experiment demonstrates that the individual learning bias can itself get magnified in the process of cultural evolution in such a way that monotone quantifiers get produced from random starting points.  In other words: cultural evolution can produce monotone quantifiers.

\subsection{Convexity}
\label{sec:convexity-learning}

Having just seen computational explanations for the monotonicity universal for quantifiers, we here introduce a generalization to a related but slightly weaker property which, however, has the advantage of applying to many categories of expression: convexity.  In general, a set $X$ is convex just in case the following holds: for any two points $a, b \in X$, every point $c$ that lies \emph{in between} $a$ and $b$ also belongs to $X$ \citep{gardenforsConceptualSpacesGeometry2000,chemlaConnectingContentLogical2019}.  Particular cases correspond to specifying the relevant betweenness relation.

\subsubsection{Applied to Quantifiers}
\label{sec:convexity-quantifiers}

In the case of quantifiers, the relevant in-betweenness relation can be defined in terms of the submodel relation from \S~\ref{sec:degrees-universals}: $M''$ is in between $M$ and $M'$ just in case $M \preceq_B M'' \preceq_B M'$.  With this definition in hand, convex quantifiers---often also called connected \citep{chemlaConnectingContentLogical2019}---correspond to conjunctions of an upward and downward monotone quantifier. As ane example, \textsf{between m and n} is the conjunction of \textsf{at least m} and \textsf{at most n}.  As noted in \S\ref{sec:monotonicity-definition}, this is why \citet{barwiseGeneralizedQuantifiersNatural1981} formulated the Monotonicity Universal to include conjunctions of monotone quantifiers, i.e. convex ones, so that numerals with an exact semantics (`five' meaning \textsf{exactly five}) are covered. It is also possible that some such conjunctions are lexicalized: Polish \emph{kilka}, a morphologically simple indefinite numeral, is usually glossed as `between three and nine' \citep{wagiel2020several, szymanik2022polish},\footnote{Dictionaries differ on the exact bounds.} although whether its upper bound is semantic or pragmatic has not, to our knowledge, been tested.

The ease of learning hypothesis integrates well with this conception of convex quantifiers: the degree of monotonicity (\S\ref{sec:degrees-universals}) has the property that convex quantifiers are less than fully monotone, but tend to have high degree, and the experiments reported in \S\ref{sec:degree-monotonicity-learning} show that convex quantifiers are harder to learn than monotone ones, but easier than other quantifiers with a lower degree of monotonicity.  Similarly, in simple rule learning experiments with five objects, \citet{chemlaConnectingContentLogical2019} found that monotone properties were easier to learn (for adults) than convex ones, which were easier to learn than non-convex ones. A follow-up study with baboons found that monotone and convex rules were easier to learn than non-convex ones, but equally easy to learn as each other \citep{chemlaConstraintsLexiconsHuman2019}. In a large-scale artificial word learning experiment with adults, \cite{RAMOTOWSKA2026106523} found that the upward monotone \textsf{at least 3} was learned better than both the convex \textsf{between 3 and 6} and the non-convex \textsf{an even number}; an advantage of convex over non-convex emerged only once differences in participants’ learning strategies were controlled for.   Future work can try to explain and refine these results.  Similarly, \citet{gierasimczukConvexityMonotonicityLanguage2023} study the emergence of monotone and convex quantity expressions through computational simulations of a communication game, finding that both do emerge but that convexity tends to dominate the populations of their agents.

\subsubsection{Color Terms and Beyond}

The real power of the convexity property, however, is that it can apply to any domain where a betweenness relation can be defined.  When this can be done, it can be argued that lexical meanings in that domain should be convex \citep{gardenforsGeometryMeaning2014,chemlaConnectingContentLogical2019}.  This corresponds to a kind of `meta' lexical property universal, which has been articulated by \cite{gardenforsConceptualSpacesGeometry2000}:

\begin{itemize}[align=left, leftmargin=*]
	\item[\textsc{Convexity Universal:}]All lexical categories satisfy convexity. 
 \end{itemize}

One further specific lexical domain where this property has been explored in depth comes from color terms (e.g.\ `green' and `blue' in English.).  Here, the betweenness relation is straightforward to define, using the CIELab color space \citep{CIE2004}, a geometric space defined in such a way that Euclidean distance corresponds to human judgments of color similarity and discriminability: if two points $x$ and $y$ belong to the denotation of a color term, then every point on the straight line between $x$ and $y$ must also belong to the denotation of the color term. 
\citet{gardenforsConceptualSpacesGeometry2000} has realized that and put forward a color-specific version of the convexity universal:
\begin{itemize}[align=left, leftmargin=*]
	\item[\textsc{Color Convexity Universal:}]All color terms denote convex regions of color space. 
\end{itemize}
Building on pioneering early work by \citet{berlinBasicColorTerms1969} and using their large-scale World Color Survey dataset, \citet{jagerNaturalColorCategories2010} has analyzed color denotation across languages and provided empirical support for the Color Convexity Universal.

\begin{figure}[ht!]
	\centering
	\includegraphics[width=0.8\textwidth]{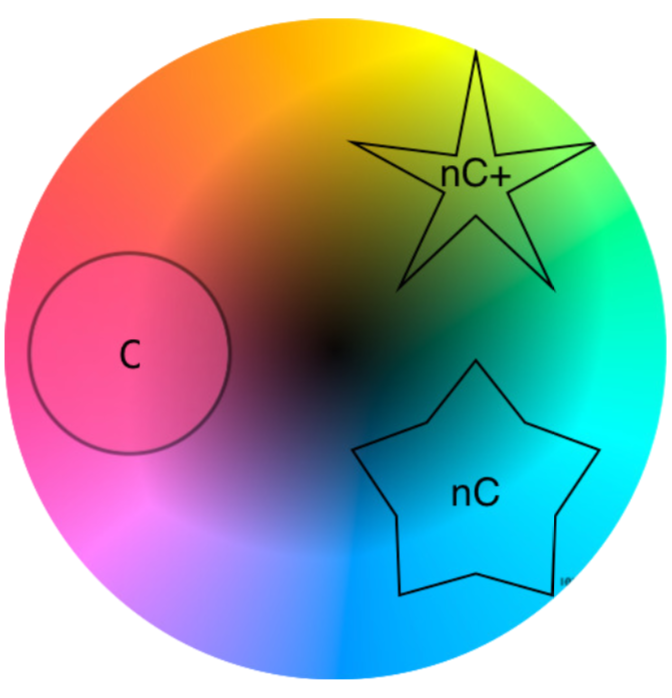}
	\caption{A stylized example of three regions in a color space with intuitively varying degrees of convexity.  The circular region C is fully convex.  Both nC and nC+ are non-convex, but, intuitively, nC is closer to being convex than nC+. The degree of convexity captures this: both regions share the same convex hull (the pentagon connecting the points of the star), but nC covers more of the convex hull than nC+.}
	\label{fig:color-convexity}
	\addsource{\cite{steinert-threlkeldEaseLearningExplains2020}}
 \end{figure}

 \citet{steinert-threlkeldEaseLearningExplains2020} tested \textsc{The Ease of Learning Hypothesis} within the domain of colors.  If true, one should expect convex color terms to be easier to learn than non-convex ones. In short, they generated many artificial color systems, measured their degree of convexity, and then trained artificial neural networks to learn those systems.  The artificial color naming systems were generated by partitioning CIELab space into seven different categories (see the paper for details on the exact algorithm used). Then for each color system, they measured its degree of convexity. The intuition behind this measure is that various non-convex regions may be closer or farther away from their nearest perfectly convex areas; see Figure~\ref{fig:color-convexity} for an example. For every region $X$, they first extend it to the smallest convex region that contains it, its so-called convex hull. Then the degree of $X$ becomes $|X|/ |\mathrm{ConvexHull}(X)|$: how much of the convex hull is covered by $X$, or, conceptually, how much `effort' is needed to extend $X$ to its convex hull. For every color system, they measure the convexity for each color term denotation, and the convexity degree for the entire system is given as a weighted average of the degrees of all colors within the system. 

\begin{figure}[ht!]
	\centering
	\includegraphics[width=\textwidth]{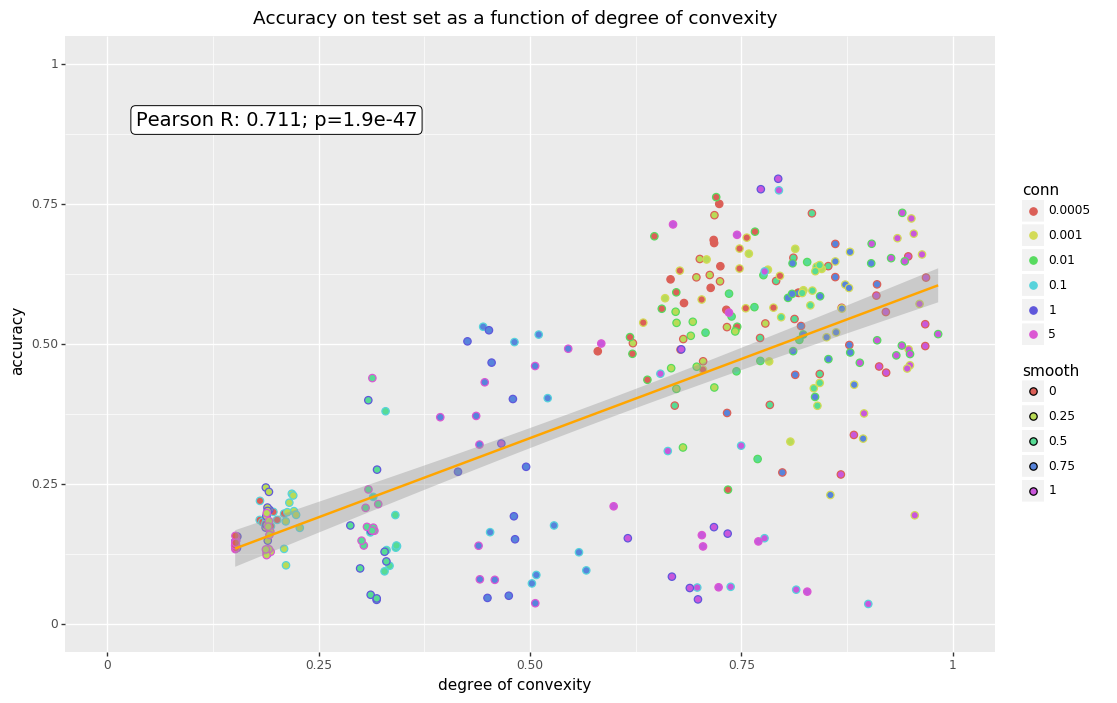}
	\caption{Accuracy on a test set (examples of color points not seen during training) as a function of degree of convexity for artificial color systems. A regression line shows that the more convex the color system, the higher the accuracy.}
	\label{fig:color-convexity-learning}
	\addsource{\cite{steinert-threlkeldEaseLearningExplains2020}}
\end{figure}

To measure the ease of learning of each color system, \citet{steinert-threlkeldEaseLearningExplains2020} trained feed-forward neural networks to categorize single points from the CIELab space into their color category.  They measured the generalization accuracy after a fixed amount of training time, and analyzed this accuracy as a function of the degree of convexity. As shown in Figure~\ref{fig:color-convexity-learning}, the more convex the color system, the higher the learning accuracy. Statistical analysis of the correlation provided in the paper confirmed that convexity is the best explanation for the artificial network's ability to learn color systems (over and above other simple geometrical properties of the color systems). It has also shown that the results hold even when controlling for the linear separability of the color systems. In sum, this research has shown that more convex color systems are easier to learn,\footnote{\citet{douvenLearnabilityNaturalConceptsforthcoming} extends this methodology to compare systems that are all convex already and identify further features of natural systems compared to non-natural ones that map onto learning differences.} extending the ease of learning explanation from a domain of function words (quantifiers) to a domain of content words (colors). We consider this an argument in favor of the robustness of the ease of learning explanation. The next few sections show that the ease of learning explanation has been used in a few other case studies as well.

\subsection{Quantifiers: Quantity}
\label{sec:quantity-learning}
Monotonicity may be one of the better-studied quantifiers universal, but there are other cross-linguistically attested constraints on the space of determiners. One fundamental property exploited already in the definition of generalized quantifiers is the isomorphism closure or, in other words, quantity property. The universal captures the fact that in natural languages (as opposed to some logical formalisms), quantifiers genuinely talk about quantities. In other words, whether or not a sentence of the form Det CN VP is true depends only on the cardinalities of the sets denoted by the common noun and verb phrase (as well as their intersection and differences). Formally: 

\begin{examples}
\item \textsf{Q} is \emph{quantitative} if and only if whenever $\langle M, A, B \rangle \in \textsf{Q}$ and there exists $\langle M', A', B' \rangle$ such that  $|A\cap B|=|A'\cap B'|$, $|A\setminus B|=|A' \setminus B'|$, $|B \setminus A|=|B' \setminus A'|$, and $M\setminus|A\cup B|=M' \setminus |A'\cup B'|$ then $\langle M', A', B' \rangle \in \textsf{Q}$.
\end{examples}

For instance, the determiner ``at least 3'' is quantitative as the sentence truth value of the sentence (\ref{atleast}) does not depend on the specific order in which people are queuing. On the other hand, the sentence (\ref{first3}) will change its truth value with the reordering of the queue. Hence, the quantifier ``the first three'' is not quantitative.   

\begin{examples}
\item(queu) \begin{examples}
	\item(atleast) At least three people in the line are tall. 
	\item(first3) The first three people in the line are tall.
	\end{examples}
\end{examples}
\citet{keenanSemanticCharacterizationNatural1986} proposed the following universal:\footnote{See item (118) on page 311 and the discussion preceding it.}

 \begin{itemize}[align=left, leftmargin=*]
	\item[\textsc{Quantity Universal}] All monomorphemic determiners are quantitative. 
 \end{itemize}

\begin{figure}[ht]
	\centering
	\includegraphics[width=0.49\textwidth]{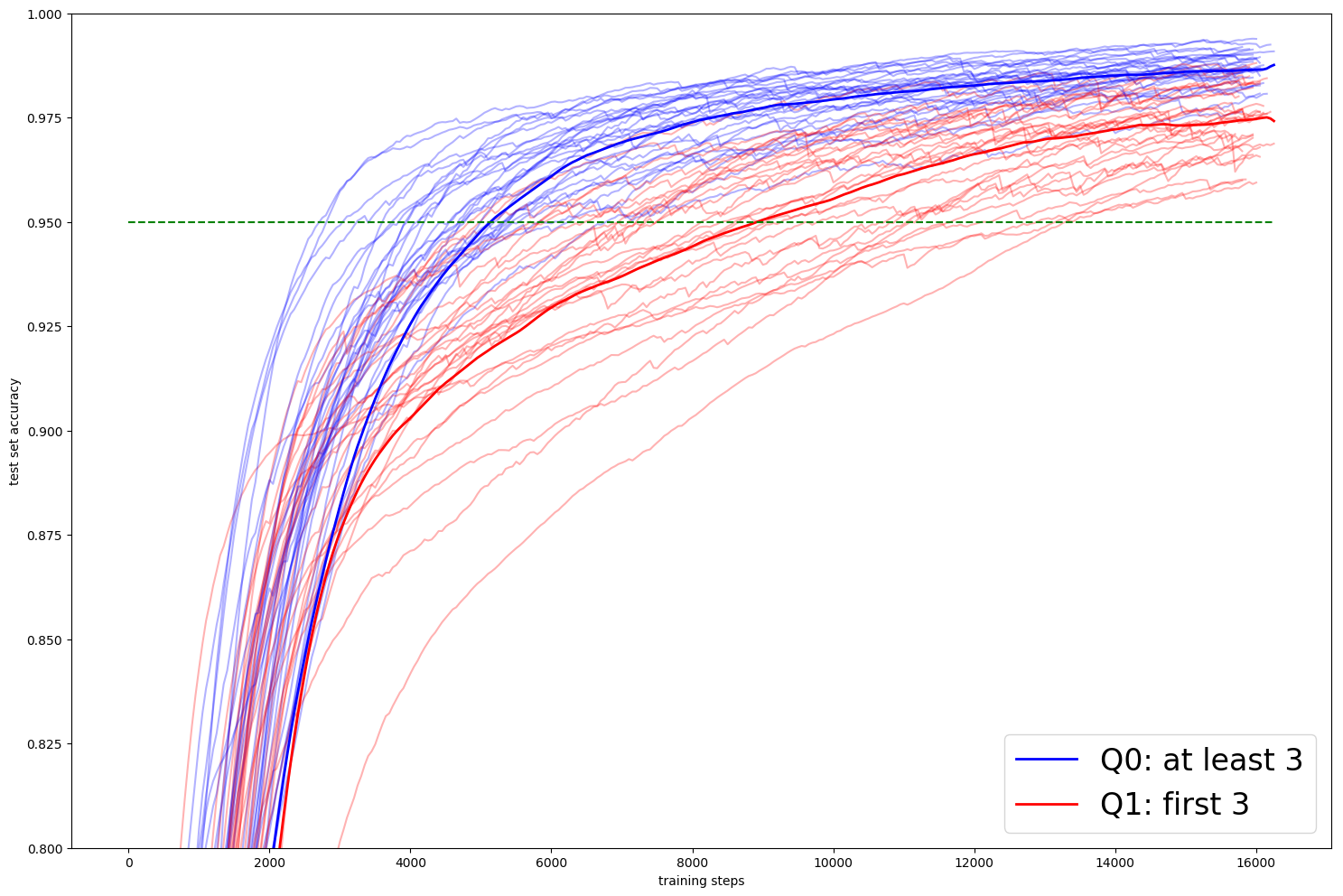}
	\includegraphics[width=0.49\textwidth]{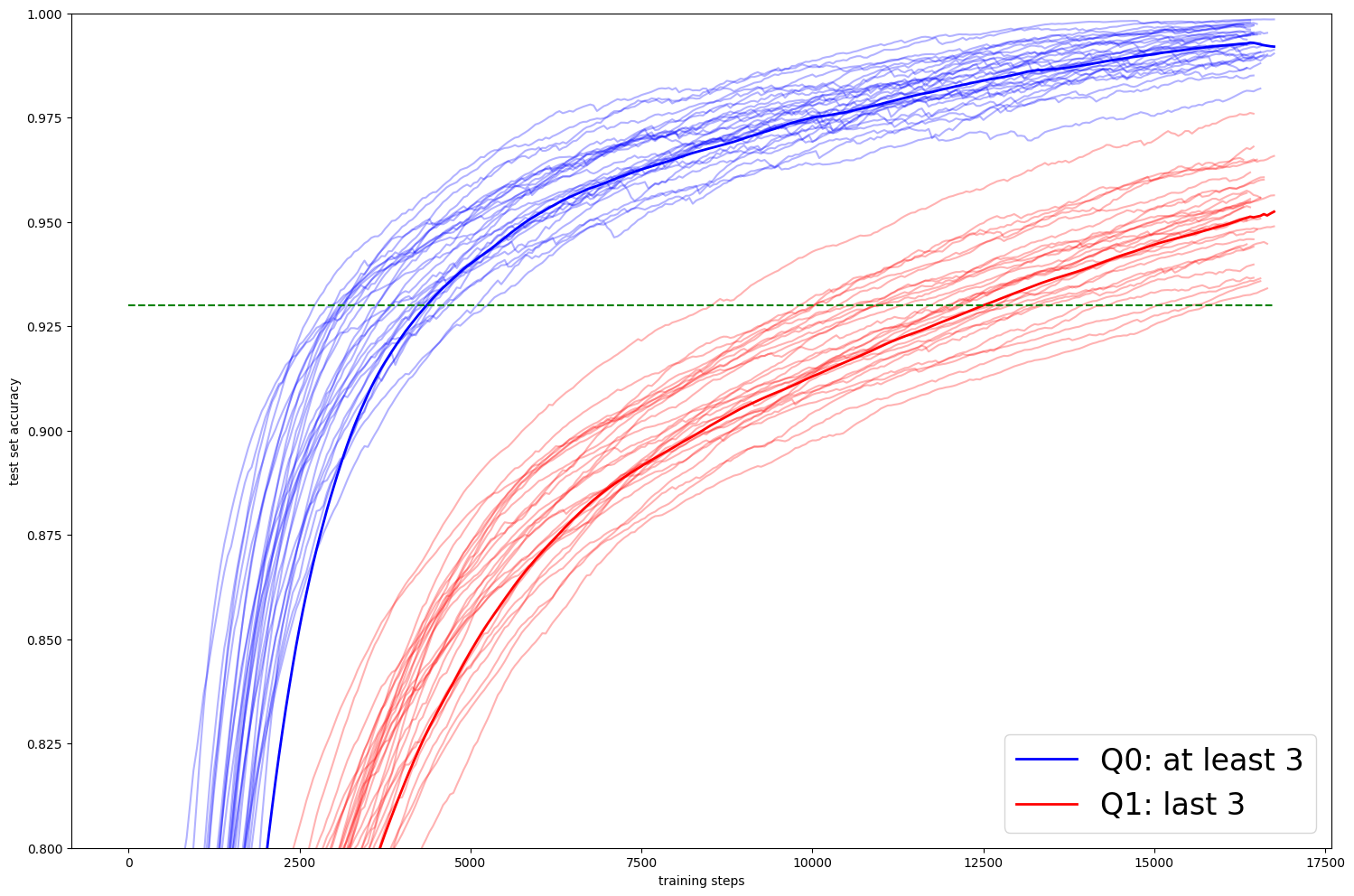}
	\caption{Learning curves for the quantity universal. The left panel shows the learning curves for the quantitative quantifier ``at least 3'' (blue) and the non-quantitative quantifier ``the first 3'' (red). The right panel shows the learning curves for the quantitative quantifier ``at least 3'' (blue) and the non-quantitative quantifier ``the last 3'' (red).}
	\label{fig:quantity-learning}
	\addsource{\cite{steinert-threlkeldLearnabilitySemanticUniversals2019}}
 \end{figure}

 \cite{steinert-threlkeldLearnabilitySemanticUniversals2019} tested the ease of learning explanation for the quantity universal by applying the same methodology as in the case of monotonicity (see \S\ref{sec:monotonicity-learning}). That is, they first chose minimal contrasting pairs of quantifiers; in that case, they pitted against each other the quantitative quantifier ``at least 3'' and two non-quantitative quantifiers ``the last 3'' and ``the first 3''. They ran 29 trials of quantifier learning with LSTMs. As you can see from Figure~\ref{fig:quantity-learning}, the artificial neural network found it easier to discover the meaning of the quantitative quantifier than the non-quantitative ones. Those results support the general argument that universals may hold because expressions satisfying them are easier to learn, and such expressions are more likely to be lexicalized. \cite{RAMOTOWSKA2026106523} provide the first test of this contrast with human learners: adults learned \textsf{at least 3} more accurately than \textsf{the first 3} and \textsf{the last 3}, although this effect was less robust than those for the other universals once learning strategies were controlled for.

\subsection{Veridicality of Responsive Predicates}
\label{sec:responsive-predicates}

\cite{steinert-threlkeldExplanationVeridicalUniformity2020} has shown how ease of learning can explain recently proposed semantic universal in a rather different domain. He studies the domain of responsive predicates, e.g., `know', `believe', or `forget' \citep{lahiriQuestionsAnswersEmbedded2002}. Responsive predicates can embed both declarative and interrogative complements; see sentences (\ref{declar}) and (\ref{inter}), respectively. 
\begin{examples}
\item(climb) \begin{examples}
	\item(declar) Shane knows that Jakub climbs well. 
	\item(inter) Shane knows who is the better climber.
	\end{examples}
\end{examples}
To better understand how such verbs' declarative and interrogative meanings relate, \citet{spectorUniformSemanticsEmbedded2015} suggested studying the veridicality properties of such verbs. A responsive verb V is veridical with respect to declarative complements if and only if for every agent A and declarative sentence P, the following inference is valid: a. A Vs that P, therefore, P. For example, `know' is veridical for declarative complements as it follows from (\ref{declar}) that `Jakub climbs well.' A responsive verb V is veridical regarding interrogative complements if and only if for every agent A and exhaustive question Q, and sentence P expressing a true complete answer to Q, the following inference is valid: A Vs Q and P, therefore, A Vs that P. For example, `know' is veridical for interrogative complements because the true and complete answer to `who is the better climber' is `Jakub,'\footnote{Note that we are merely supposing this for the sake of the example.} %
and therefore, (\ref{inter}) implies that `Shane knows that Jakub is the better climber.'  We say that a responsive verb V is veridically uniform if it is either veridical for both kinds of complement or neither. \citet{theilerUniformSemanticsDeclarative2018} formulated the following universal claim:\footnote{See \cite{roelofsenSearchingUniversalConstraint2020} for a possible refinement of this particular universal on the basis of presuppositions of predicates of relevance.  The learning experiments reported below do not explicitly model presuppositions and so are not powerful enough to distinguish between these two proposals.} 

\begin{itemize}[align=left, leftmargin=*]
	\item[\textsc{The Veridical Uniformity Universal}] All responsive verbs are veridically uniform. 
 \end{itemize}

\citet{steinert-threlkeldExplanationVeridicalUniformity2020} wanted to know whether artificial neural network learnability can also explain the Veridical Uniformity Universal. He devised two non-uniform fake verbs to study this question: `knopinion' and `wondows.' The first verb means `to know' when it embeds an interrogative and `to be opinionated' when it embeds a declarative sentence. It is, therefore, veridical for interrogative complements and non-veridical for declarative complements. The second verb means `to be uncertain' when it embeds an interrogative but `to know' when it embeds a declarative. Hence, it is veridical for declarative complements and non-veridical for interrogative complements. 

\begin{figure}[ht]
	\centering
	\includegraphics[width=\textwidth]{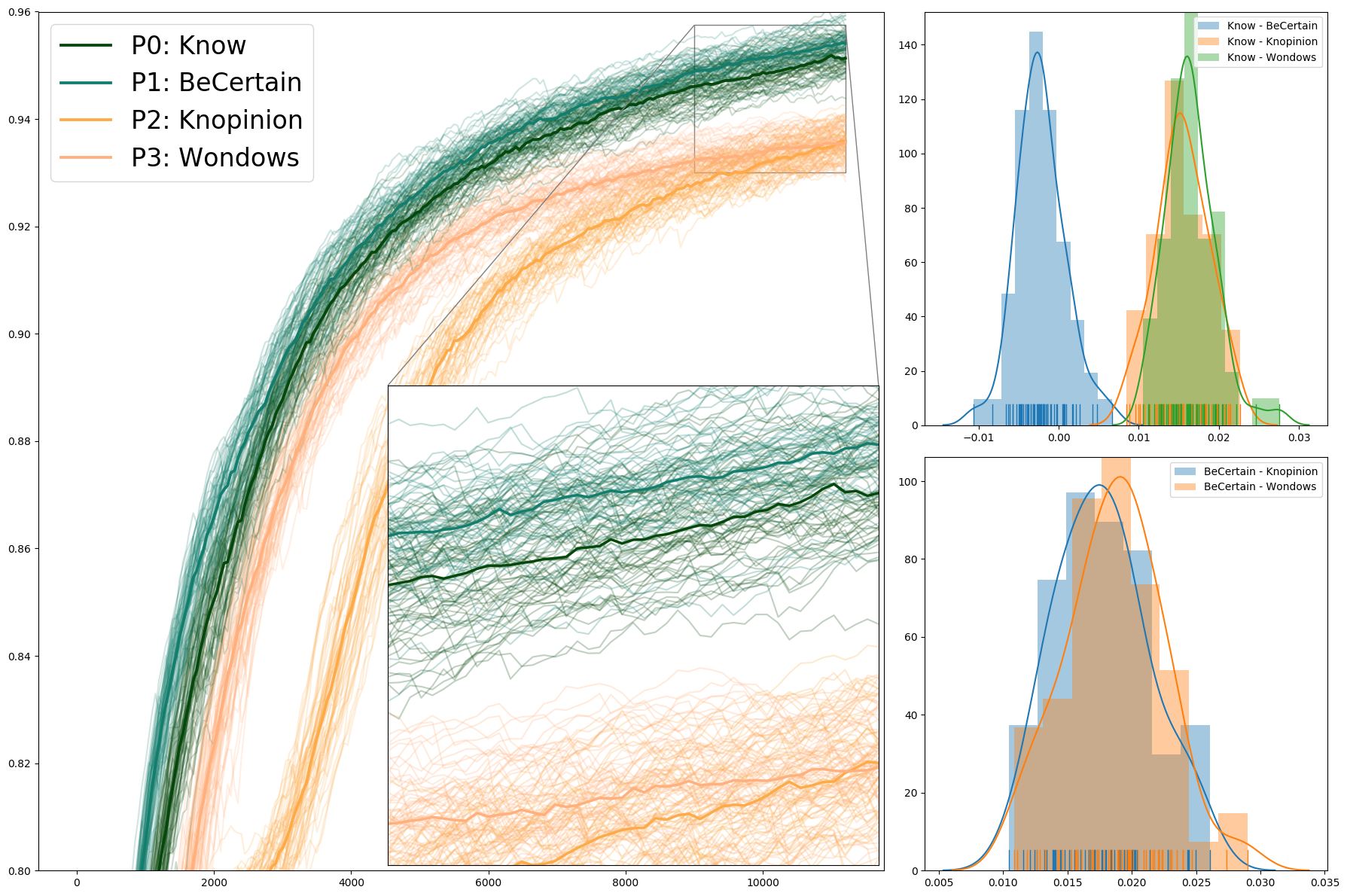}
	\caption{Learning curves for the veridical uniformity universal. The green lines correspond to the attested verbs and the orange lines to the unattested ones.  There is a significant difference in how well the former are learned compared to the latter, and no difference within the two groups.}
	\label{fig:veridicality-learning}
	\addsource{\cite{steinert-threlkeldExplanationVeridicalUniformity2020}}
\end{figure}

Using possible-world semantics to encode the meaning of the verbs, \citet{steinert-threlkeldExplanationVeridicalUniformity2020} presented artificial neural network learning experiments comparing the difficulty of the two fake verbs and two attested verbs: `to know' and `to be certain.' He measured accuracy on held-out examples after a fixed training time for simple feed-forward networks to assess for which verbs it is harder to extract the truth conditions from the training data. As seen in Figure~\ref{fig:veridicality-learning}, the neural networks consistently achieved higher learning accuracy on attested verbs satisfying the veridicality universal than on non-attested verbs violating the universal. This result further strengthens the robustness of the universal learnability explanation and the methodology involving the neural network toolbox to asses the in-principle complexity of learnability for different semantic problems.

In addition to these machine learning results, \cite{maldonadoLearnabilityConstraintsSemantics2022} conducted an artificial language learning experiment with novel responsive predicates.  They compared two such predicates, one having the meaning of `falsely believes $p$' and the other `correctly knows that $p$ is false'.  After learning the meanings in declarative contexts, participants had to extend the meaning to novel interrogative contexts. These predicates are both veridically uniform,\footnote{This holds because, even without specifying a meaning for interrogatives, they will be non-veridical with respect to interogative complements since `A Vs the true answer' will always be false based on their declarative meanings.} but were chosen to study a closely related property called C-distributivity \citep{theilerUniformSemanticsDeclarative2018}.  The results suggested that participants were likely to extend both meanings in a way compatible with C-distributivity.  Although this paradigm does not directly address \textsc{The Ease of Learning Hypothesis}, it demonstrates another method by which learning can be used to test semantic universals.

\subsection{A counter-example? Conservativity of Quantifiers}
\label{sec:conservativity-learning}

One final example of a lexical property universal comes once again from the domain of quantifiers: conservativity.  This property \dash arguably the most widely discussed of the three \dash captures the intuition that the restrictor (the first argument to a quantifier) genuinely restricts what a sentence talks about.  That is, sentences of the form \emph{Det N VP} are in some sense about the \emph{N}s and nothing else.  That this universal holds can be observed by noting the felt equivalence between the following pairs of sentences.
\begin{examples}
\item \begin{examples}
	\item Every student passed.
	\item Every student is a student who passed.
	\end{examples}
\item \begin{examples}
	\item Most Amsterdammers ride a bicycle to work.
	\item Most Amsterdammers are Amsterdammers who ride a bicycle to work.
	\end{examples}
\end{examples}
The formal concept at play here has been called conservativity.
\begin{examples}
\item \textsf{Q} is \emph{conservative} if and only if $\langle M, A, B \rangle \in \textsf{Q}$ if and only if $\langle M, A, A \cap B \rangle \in \textsf{Q}$.
\end{examples}
\cite{barwiseGeneralizedQuantifiersNatural1981} formulated and defended the following universal.\footnote{Because the term \emph{conservative} was not introduced until \citealt{keenanSemanticCharacterizationNatural1986}, the original formulation was in terms of a quantifier \emph{living on} a witness set.  We follow the norm of formulating in terms of conservativity for concision.} 
\begin{itemize}[align=left, leftmargin=*]
	\item[\textsc{Conservativity Universal:}] All simple determiners are conservative.\footnote{In fact, conservativity often is a claim about all determiners, not just the simple ones, a point we return to below.\label{fn:conservativity}}
\end{itemize}
This universal rules out quantifiers that depend on other portions of the model besides $A$, such as $B \setminus A$.  As an example, there is no determiner \emph{equi} in any language such that the following two sentences are equivalent in meaning.
\begin{examples}
\item \begin{examples}
	\item Equi students are at the park.
	\item The number of students is the same as the number of people at the park.
	\end{examples}
\end{examples}

Following the same logic of the experiments described in Sections \ref{sec:monotonicity-learning} and \ref{sec:quantity-learning}, \cite{steinert-threlkeldLearnabilitySemanticUniversals2019} tested the ease of learning explanation for the conservativity universal with two minimal pair experiments, but crucially found \emph{no difference in ease of learning} between conservative and non-conservative quantifiers.\footnote{See Section~4.3 of \citet{steinert-threlkeldLearnabilitySemanticUniversals2019} for the learning curves and quantitative results.} As \cite{steinert-threlkeldLearnabilitySemanticUniversals2019} themselves note, this null result may partly reflect their encoding of models, which treats the two arguments of the quantifier symmetrically; human learners, by contrast, assign the restrictor a special role. Representational complexity does not settle the matter either: in aggregate, conservative quantifiers have a shorter minimal description length than non-conservative ones, but the specific pairs used in learning experiments, such as \textsf{not all} and \textsf{not only}, are equally complex \citep{polQuantifiersSatisfyingSemantic2023}. Although these results may count as evidence against \emph{The Ease of Learning Hypothesis}, they may also point to a deep way in which \textsc{Conservativity} may be different from the other universals we have discussed in this chapter.  In particular: it may not be a lexical property universal at all.  

A growing number of researchers argue that conservativity is not a constraint on which quantifiers are lexicalized as determiners but is rather an artifact of the syntax-semantics interface. One family of views in this vein develop a so-called \emph{structural} explanation of conservativity \citep{foxAntecedentContainedDeletionCopy2002,sporticheDivisionLaborMerge2005,romoliStructuralAccountConservativity2015}.  While the details of the proposals go beyond the present scope, the key idea can be explained: even if determiners in principle could denote non-conservative quantifiers, because of the way that the syntax-semantics interface constructs sentence meanings, any sentence with a non-conservative determiner becomes truth-conditionally equivalent to a sentence with a conservative determiner.\footnote{Strictly speaking, such sentences might also be trivial.  For this reason, \cite{romoliStructuralAccountConservativity2015} assumes that trivial meanings are blocked.} Another family of views argues that conservativity arises because determiners are devices for explicitly building \emph{restricted quantifiers} \citep{knowltonDeterminersAreConservative2020,knowltonPsycholinguisticEvidenceRestricted2023}.  On this view, the two arguments to a determiner are treated asymmetrically: a series of psycholinguistic experiments show that speakers represent the restrictor argument (the ``$A$'' argument) as a set, but only represent the nuclear scope argument (the ``$B$'' argument) as individuals, not as a set.  We should also note that under both of these styles of explanation, there will be no difference between simple and complex determiners.  As mentioned in footnote~\ref{fn:conservativity}, the universal is often formulated in a way that applies to all determiners, not merely the simple ones.

Finally, results of learning experiments with human participants are mixed. While \citet{hunterConservativityLearnabilityDeterminers2013} found a small preference for conservative quantifiers among children, these results failed to replicate in children and adults \citep{spenaderAreConservativeQuantifiers2019}, though both studies were short and had small samples. Using a syntactic frame that forced a determiner interpretation, \citet{knowltonNewEvidenceUnlearnability2022} found that adults had considerably more difficulty learning a non-conservative meaning than a conservative one. In a large-scale study with the same kind of partitive frame (\emph{gleeb of the circles are red}), \cite{RAMOTOWSKA2026106523} likewise found that conservative \textsf{not all} was learned better than non-conservative \textsf{not only}; the latter was learned above chance by only a handful of participants, but it was learnable. Strikingly, an earlier experiment without the partitive frame found no such difference \citep{ramotowska2022quantifying}. One interpretation accords well with the views in the previous paragraph: if determiners are never interpreted non-conservatively, either for structural reasons or because they build restricted quantifiers, then the difficulty lies in the construction that frames the novel word, not in the meaning itself.

Taken together, then, all of the above suggests that conservativity is not a constraint on individual lexical meanings (though see §4.4 for an efficiency-based account at the level of the lexicon and composition together).

\subsection{Discussion}
\label{sec:learning-discussion}

The previous sections have developed evidence supporting \textsc{The Ease of
Learning Hypothesis} through a series of computational experiments in several
empirical domains (quantifiers, color terms, and responsive verbs). In most
cases, we saw that meanings with a given lexical property universal were easier
to learn. This suggests that these properties do lie in a `warm' region in the
relevant space of possible meanings. In some cases (monotonicity and convexity)
we observed a richer correlation between ease of learning and a degree of the
semantic universal. This suggests that a related map of the degree of the
relevant property would globally look like the one for ease of learning.

Where human learners have been tested, the results largely converge with those
from neural networks: adults learn monotone and quantitative quantifiers more
easily than their non-monotone and non-quantitative counterparts
\citep{RAMOTOWSKA2026106523}, and both adults and baboons find convex
rules easier to learn than non-convex ones
\citep{chemlaConnectingContentLogical2019,chemlaConstraintsLexiconsHuman2019}.
Conservativity is the exception that proves instructive. Neural networks show no
learning advantage for conservative quantifiers, while human learners do, but
only when the syntactic frame forces a determiner interpretation. Together with
the structural and restricted-quantifier accounts discussed in \S2.6, this
suggests that conservativity is not a constraint on individual lexical meanings,
but rather a property of the syntax--semantics interface or of restricted
quantification.

Finally, ease of learning is not the only cognitive explanation on offer.
Meanings satisfying the quantifier universals are also simpler to represent, in
the sense of having a shorter minimal description length in a language of
thought \citep{polQuantifiersSatisfyingSemantic2023}.The two explanations
are closely related, since what is easy for a network to learn tracks what is
simple in a language of thought \citep{carcassiNeuralNetworksTrack2022}, but they
can also come apart \citep{haberlandQuantifiersThatAre2025}. In the next
chapter, representational complexity returns as one half of the trade-off that
defines efficient communication, and in \S4.4 we come back to the question of
how learnability, simplicity, and other pressures can be told apart.

\pagebreak

\section{Efficient Communication}
\label{chapter:effcomm}

As discussed in Section~\ref{sec:intro-communication}, the second optimization criterion that we will use for developing explanations of semantic universals concerns communication, as codified in \textsc{The Efficient Communication Hypothesis}: Languages that optimally balance complexity and communicative cost have expressions which satisfy lexical property universals.
This chapter will provide support for this hypothesis by developing several case studies.  After first providing a more detailed account of how to define complexity and communicative cost in general (\S~\ref{sec:complexity-cost}), we will discuss a series of empirical domains where these definitions have been fruitfully applied: quantifiers (\S\ref{sec:quantifiers-effcomm}), indefinite pronouns (\S\ref{sec:indefinites}), and modals (\S\ref{sec:modals}). We will then discuss and make a comparison with the information bottleneck approach to efficient communication (\S~\ref{sec:ib}), before concluding with a general discussion of the status of the efficient communication hypothesis (\S~\ref{sec:communication-discussion}).

\subsection{Measuring Complexity and Communicative Cost}
\label{sec:complexity-cost}

To formalize the intuitive notion of complexity from \S\ref{sec:intro-communication}---a language is complex to the extent that it is difficult to represent in the mind of a user of that language---we will borrow ideas and tools from the literature on the (probabilistic) language of thought \citep{fodorLanguageThought1975,feldmanMinimizationBooleanComplexity2000,goodmanConceptsProbabilisticLanguage2015,quilty-dunnBestGameTown2023}.  On this family of views, concepts are represented in the mind as expressions in a formal language.  Such a formal representation language will often be modeled as a (probabilistic) context-free grammar, which can generate expressions of varying depth.  For example (which we will see in more detail soon), we can represent the quantifier \textsf{all} as the mathematical expression $A \subseteq B$.  We can define the complexity of \textsf{all} as the length of the \emph{shortest} formula in the given language of thought which has the same extension.\footnote{More generally, we can use the probability according to a probabilistic grammar.  These will often be highly correlated and, in the case where all rule probabilities are uniform, perfectly correlated.}  For example: $A \setminus B = \emptyset$ has the same extension as $A \subseteq B$, but is a longer expression, and so is more complex.  We can then define the complexity of a language as the sum of the complexities of all of its expressions. This is the same notion of representational complexity that we met in Chapter 2: \citet{polQuantifiersSatisfyingSemantic2023} used minimal description length in such a grammar to show that quantifiers satisfying the universals are simpler, and \citet{haberlandQuantifiersThatAre2025} sampled the quantifiers for their learning experiments from one (\S\ref{sec:degree-monotonicity-learning}).  

To apply this definition of complexity to a particular empirical domain, we have to specify what the relevant language of thought is; this usually consists in specifying conceptual primitives and rules for building more complex representations out of these primitives.  In the case studies that follow, we will provide these details for each domain, and discuss methods for grounding these choices as well; one such method is to infer the primitives from human learning data \citep{piantadosiLogicalPrimitivesThought2016, CARCASSI2023105541}.

Recall from \S\ref{sec:intro-communication} that we want our measure of communicative cost of a language to be inversely related to how \emph{useful} it is. To be useful means to enable speakers of the language to reliably convey their thoughts to one another. We thus model usefulness (i.e.\ informativity) as an \emph{expected utility}, where the expectation is taken over a hypothetical infinite number of communicative interactions of the following form: a speaker has a thought that they wish to convey to a listener (i.e. a point in meaning space, as discussed in \S\ref{sec:parameters}), using expressions in the language.  We can capture this conception of informativity formally with the following definition, which has origins in the signaling game tradition \citep{lewisConvention1969,skyrmsSignalsEvolutionLearning2010}:
\begin{align}
    I(L) &:= \mathbb{E}_P [ u(p, p') ] \nonumber
    \\
    &:= \sum_{p \in M} P(p) \sum_{e \in L} P_S(e | p) \sum_{p' \in M} P_R(p' | e) \cdot u(p, p') \label{eqn:informativity}
\end{align}
In this definition, $L$ corresponds to a language, which consists of a set of expressions $e$.  $M$ represents the meaning space, which is a set of points $p$ that the speaker might wish to convey (i.e.\ the `thoughts' in the informal gloss). $P(p)$ is the prior probability that a speaker will need to convey point $p$.  This is often referred to as a \emph{communicative need} distribution \citep{kempSemanticTypologyEfficient2018, zaslavskyCommunicativeNeedColour2020}.  $P_S(e | p)$ is a model of the \emph{Sender}: given the point $p$ they wish to convey, this sender model produces a distribution over expressions in the language.  $P_R(p' | e)$ is a model of the \emph{Receiver}: given an expression $e$, this receiver model produces a distribution over points in the meaning space.  We note that these models can vary in complexity, from so-called `literal' agents which only reason about literal meanings, to more complex `pragmatic' agents that engage in iterative reasoning about each other \citep{frankPredictingPragmaticReasoning2012a,goodmanPragmaticLanguageInterpretation2016,degenRationalSpeechAct2023}.  Examples below will make use of both of these options.

Finally, $u(p, p')$ is a utility function that measures the utility of the receiver inferring $p'$ when the speaker intended to convey $p$.  A simple choice is a Dirac delta function---$\delta(p, p') = 1$ if $p=p'$, $0$ otherwise---which captures an `all-or-nothing' sense of utility: the agents get credit only if the receiver recovers exactly the intended point. A more graded alternative gives partial credit according to how close the listener’s guess comes to the intended point, for instance $u(p,p') = 1/(1 + d(p,p'))$ for a distance $d$ on the meaning space, so that near misses count for more than distant ones \citep{jagerEvolutionConvexCategories2007, oconnorEvolutionVagueness2014, carcassi2021alternatives}. We will see a utility of this kind in \S\ref{sec:quantifiers-effcomm}. With all of this in place, if we assume that $u(p, p') \in [0, 1]$, then $I(L) \in [0, 1]$ as well.  We can then define the communicative cost of a language as the complement of its informativity:
$$ C(L) := 1 - I(L) $$
Communicative cost, then, has negative valence: a value of $0$ (no cost) means perfect (degree $1$) informativity, and a value of $1$ (maximal cost) means no successful communication at all.

To apply this definition of communicative cost to a particular empirical domain, we have to specify the meaning space $M$, the communicative need distribution $P(p)$, the sender model $P_S(e | p)$, the receiver model $P_R(p' | e)$, and the utility function $u(p, p')$.  In the case studies that follow, we will provide these details for each domain, and discuss choice points along the way. 

\subsection{Quantifiers}
\label{sec:quantifiers-effcomm}

\cite{steinert-threlkeldQuantifiersNaturalLanguage2021} applied the above methodology to the primary lexical property universal for quantifiers that we have discussed previously: monotonicity.\footnote{The paper---as well as \cite{steinert-threlkeldQuantifiersNaturalLanguage2019}---also includes an analysis of `quasi-natural' quantifiers, based on a typological study \citep{keenanHandbookQuantifiersNatural2012,papernoHandbookQuantifiersNatural2017}, as well as conservativity and a degree thereof.  We focus on monotonicity here given the depth of the discussion in the previous chapters of the present book, but refer the reader to the paper for further details on these other cases.}  For measuring complexity, the relevant grammar included the primitives in Table~\ref{tab:quantifiers-grammar}. This is essentially the grammar used by \citet{polQuantifiersSatisfyingSemantic2023} to measure minimal description length (\S\ref{sec:degree-monotonicity-learning}).  For example, having a Boolean operator for conjunction $\land$ means having the rule $\textsc{Bool} \to \textsc{Bool} \land \textsc{Bool}$ in the grammar.  The final primitives in the grammar are symbols $A$ and $B$, denoting the corresponding sets in a model.

\begin{table}[ht!]
	\centering
	\begin{tabular}{ccc}
		\toprule
		Boolean & Set-Theoretic & Numeric \\
		\midrule
		$\land$, $\lor$, $\lnot$ 
		& 
		$\cap, \cup, \subset, | \cdot |$
		&
		$/ , +, -, >, =, \%$
		\\
		\bottomrule
	\end{tabular}
	\caption{The operators in the grammar for measuring the complexity of quantifiers.}
	\label{tab:quantifiers-grammar}
\end{table}

To apply equation \ref{eqn:informativity} for informativity, \citeauthor{steinert-threlkeldQuantifiersNaturalLanguage2021} considered all models up to a fixed size, and then assumed a uniform communicative need distribution as well as uniform distributions for the sender and receiver ((e.g. $P_S(Q | \mathbb{M}) = 1/n$ if $\mathbb{M} \in Q$, $0$ otherwise, where $n = |\{Q \in L : \mathbb{M} \in Q \}|$ is the number of quantifiers in $L$ containing $\mathbb{M}$)).

For the utility, however, a more fine-grained version than the all-or-nothing option was employed.
In this application, the utility $u(\mathbb{M}', \mathbb{M})$ measures how good it is for the listener to guess $\mathbb{M}'$ when the speaker had in mind $\mathbb{M}$.  We base this on a measure of the distance between models, capturing the notion that non-exact matches can still be better or worse \citep{jagerEvolutionConvexCategories2007,oconnorEvolutionVagueness2014, carcassi2021alternatives}.  More precisely:
\begin{align*}
    u(\mathbb{M}' , \mathbb{M}) &= \frac{1}{1 + d(\mathbb{M}', \mathbb{M})} \text{ where }
    \\
    d(\mathbb{M}' , \mathbb{M}) &= \sum_{X \in A \setminus B, A \cap B, B \setminus A, M \setminus (A \cup B)} \max \{ 0, |X| - |X'| \}
\end{align*}
Intuitively, this measure is inversely proportional to how many elements one has to move to transform the listener's guessed model into the sender's model (by summing this value across the four `zones' in a model of the form $\langle M, A, B \rangle$).\footnote{The addition of 1 in the denominator both prevents division by zero and makes distance-0 models have maximal utility of 1.}  For example, suppose $\mathbb{M}$ has 3, 4, 2, and 1 elements in $A \setminus B, A \cap B, B \setminus A, M \setminus (A \cup B)$ respectively, and $\mathbb{M}'$ has 2, 4, 3, and 1 elements in the same zones.  We then have $d(\mathbb{M}, \mathbb{M}') = 1$, since moving one element from $A \setminus B$ to $B \setminus A$ will make the four zones have the same size in the two models.

\begin{figure}[ht!]
    \centering
    \includegraphics[width=\textwidth]{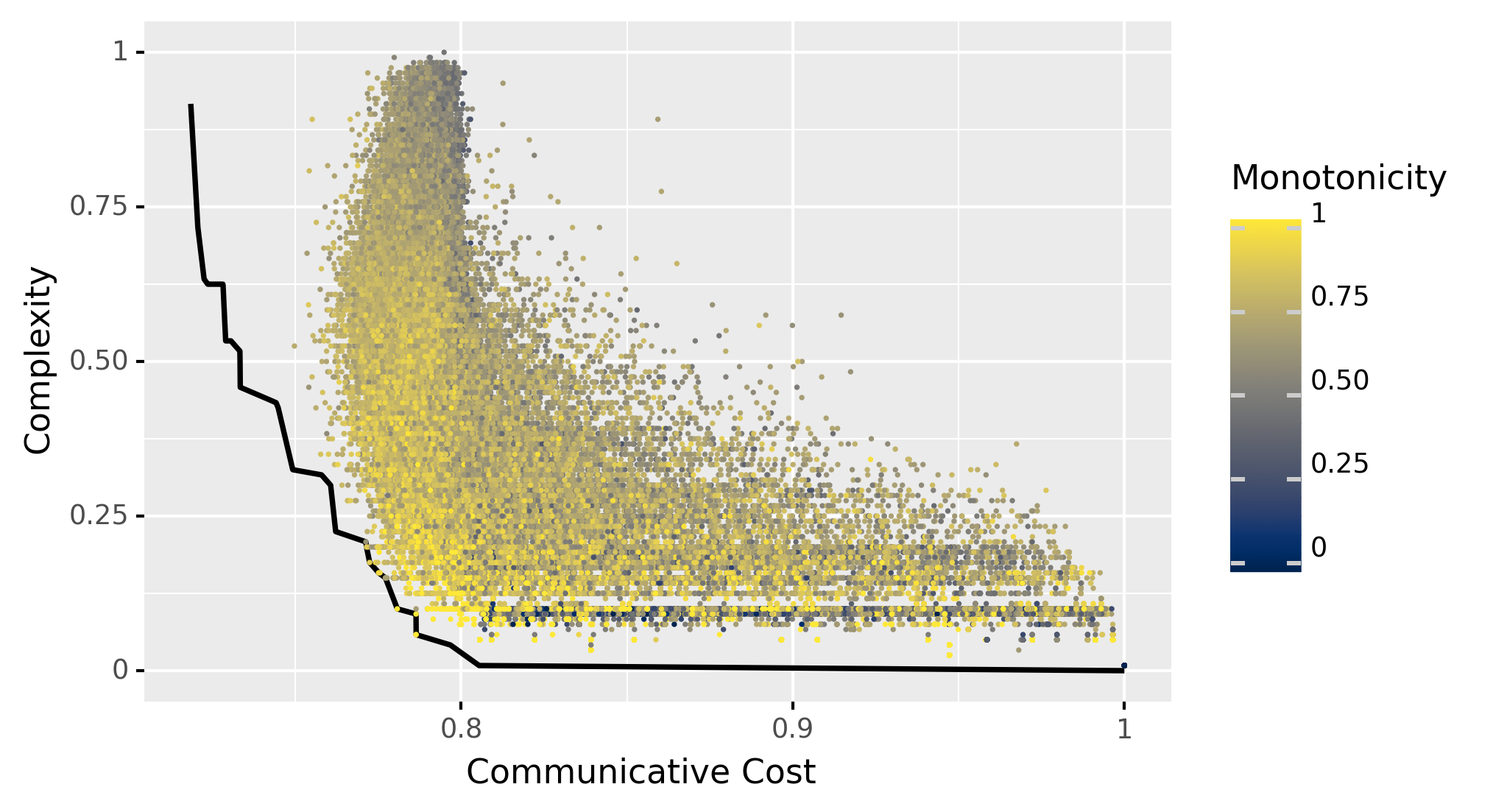}
    \caption{The communicative cost and complexity of quantifiers, colored by the degree of monotonicity.  The increased `yellow'-ness as one nears the Pareto frontier (in black) shows that the degree of monotonicity is correlated with how optimally languages trade off these two pressures.}
    \label{fig:quantifiers-effcomm}
    \addsource{Adapted from \cite{steinert-threlkeldQuantifiersNaturalLanguage2021}}
\end{figure}

The results of this analysis are shown in Figure~\ref{fig:quantifiers-effcomm}.  The $x$-axis shows the communicative cost of the languages, and the $y$-axis the complexity.  The languages are colored by the degree of monotonicity.  The Pareto frontier of optimal languages is depicted in black (the referenced paper contains further information on how this frontier was estimated).  The increased `yellow'-ness as one nears the Pareto frontier shows that the degree of monotonicity is correlated with how optimally languages trade off these two pressures (a statistical analysis confirms this, with degree of optimality measures as Euclidean distance to the closest point on the Pareto frontier).  This constitutes evidence for \textsc{The Efficient Communication Hypothesis}: languages which do better at balancing between the competing pressures to be simple and to be useful also produce more monotone quantifiers.

\subsection{Indefinite Pronouns}
\label{sec:indefinites}

\begin{figure}[ht!]
	\centering
	\includegraphics[width=\textwidth]{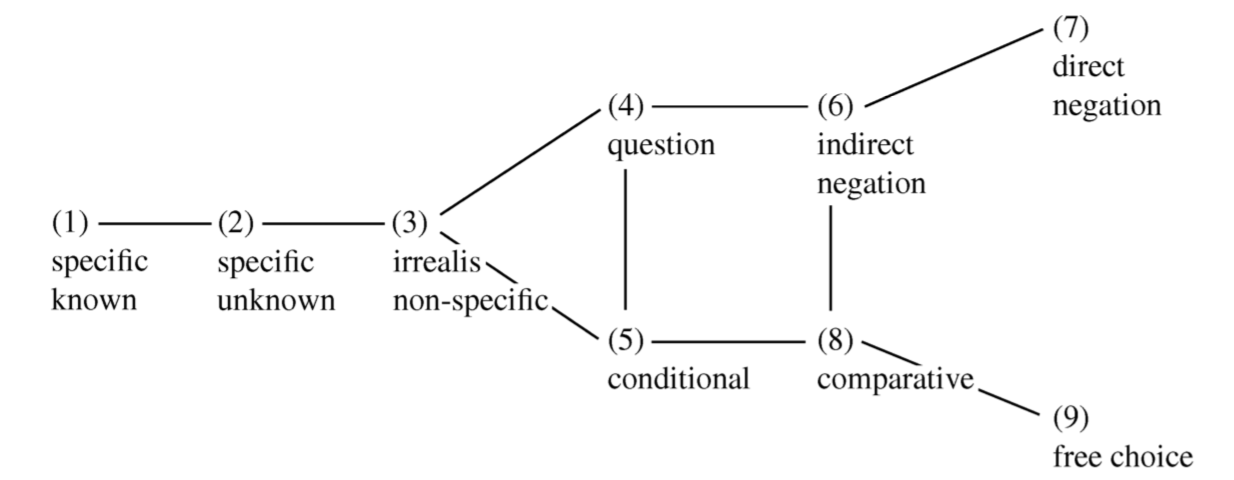}
	\caption{The proposed semantic map of indefinite pronouns.  As discussed in the body of the text, \cite{denicIndefinitePronounsOptimize2022} reduced these functions to a smaller set of six flavors to isolate semantic as opposed to syntactic functions.}
	\label{fig:indefinites-semantic-map}
	\addsource{\cite{haspelmathIndefinitePronouns2001}}
\end{figure}

\cite{denicComplexityInformativenessTradeoff2020,denicIndefinitePronounsOptimize2022} extend the efficient communication approach to another empirical domain: indefinite pronouns.  These are function words like English \emph{somebody, anyone, noone}, which have a rich history of cross-linguistic study.  In one prominent study, \cite{haspelmathIndefinitePronouns2001} proposed a lexical property universal for this domain by first defining a \emph{semantic map}: a graph whose nodes are possible semantic functions that can be expressed by indefinites, with edges between such functions when those functions have been expressed by a single lexical item in some language.\footnote{See \cite{andersonPerfectUniversalLanguageParticular1982,croftInferringUniversalsGrammatical2008,cysouwIntroductionSpecialIssue2010} for more on this particular methodology.}  Figure~\ref{fig:indefinites-semantic-map} shows the proposed semantic map for indefinites. 

With this map in hand, \cite{haspelmathIndefinitePronouns2001} proposed the following lexical property universal: an indefinite pronoun can only express functions which form a \emph{connected} region in the semantic map. For example, if an indefinite pronoun can take functions 2 and 4, it is also able to take function 3.  These statements, of the form `If an indefinite pronoun can take functions X and Y, it is also able to take function Z', are called \emph{Haspelmath's (implicational) universals}.

Two facts are worth noting about these universals.  (1) The property of connectedness in a graph like this very closely resembles that of convexity, as discussed for larger spaces in \S\ref{sec:convexity-learning}.  We will return to this connection in \S\ref{sec:modals}.  (2)  Similarly, like convexity, typologists have proposed this connectedness constraint as a cross-categorical universal.  \citet{croftInferringUniversalsGrammatical2008} explicitly state \textsc{the Semantic Map Connectivity Hypothesis}: ``any relevant language-specific and/or construction-specific category should map onto a connected region in conceptual space'' (p. 4).  We will return to a discussion of both of these points in \S\ref{sec:taxonomy}.

Unlike the quantifiers case, \cite{haspelmathIndefinitePronouns2001} also collected detailed inventories of the indefinite systems in many languages, which allowed \citet{denicIndefinitePronounsOptimize2022} to conduct two separate analyses: one comparing natural languages to other possible languages directly, and another comparing languages which satisfy Haspelmath's universals (`Haspel-ok') to others which do not (`Haspel-notOk').  The semantic space they used comprised six semantic flavors, which also make up the individual points in the semantic space (e.g. specific known, specific unknown, free choice, ...); these were derived by combining some functions in the semantic map above, because some of them reflect syntactic distribution facts as opposed to pure semantic facts. (We refer the reader to \cite{denicIndefinitePronounsOptimize2022} for more details on the translation procedure.)  An individual indefinite pronoun then denotes a set of these flavors, i.e.\ the ones it can be used to express.

Measuring complexity was done as before, with a grammar that contained primitives for combinations of the semantic flavors (e.g. `K' for whether the referent is known to the speaker or not).  Composition rules included set intersection and union.  Table~\ref{tab:indefinites-eng-russ} puts all of these ingredients together and shows the inventories and complexities for two of the languages in the sample (English and Russian).

\begin{table}[ht!]
	\centering
	\resizebox{\textwidth}{!}{%
	\begin{tabular}{lllll}
		\toprule
		Language & Indefinite pronoun & Flavors & Feature formula & $c(i)$
		\\
		\midrule
		\multirow{3}{*}{English} &
		someone & specific known, specific unknown, non-specific & {\small $SE^{-}$ } & 1
		\\
		& anyone & negative polarity, free choice & {\small  $SE^{+} \cap N^{-}$} & 2
		\\
		& no one & negative indefinite & {\small  $N^{+}$}  & 1
		\\
		\midrule
		\multirow{7}{*}{Russian} &
		kto-to & specific unknown, non-specific & {\small $K^{-} \cap SE^{-}$} & 2
		\\
		& kto-nibud' & non-specific & {\small $S^{-} \cap SE^{-}$} & 2
		\\
		& kto-libo & non-specific, negative polarity & {\small $(S^{-} \cap SE^{-}) \cup ((SE^{+} \cap R^{+}) \cap N^{-})$} & 5
		\\
		& nikto & negative indefinite & {\small $N^{+}$} & 1
		\\
		& koe-kto & specific known & {\small $K^{+}$ } & 1
		\\	
		& kto by to ni bylo & negative polarity & {\small  $(SE^{+} \cap R^{+}) \cap N^{-}$ }& 3
		\\	
		& kto ugodno & free choice & {\small  $SE^{+} \cap R^{-}$} & 2
		\\ 
		\bottomrule
	\end{tabular}
	}
	\caption{{Indefinite pronouns in English and Russian, with the flavors they convey, their shortest feature formulae and their resulting complexities (number of atoms).}}
	\label{tab:indefinites-eng-russ}
\end{table}

Two notes about the informativity measure for this case study are worth noting.  First, since the semantic space here has very little structure, the all-or-nothing utility function was used.  Second, the communicative need distribution was not uniform, but rather based on frequency as estimated in a corpus.  In particular, \cite{beekhuizenSemanticTypologyParallel2017} annotate a parallel corpus of subtitles with Haspelmath's functions; \cite{denicIndefinitePronounsOptimize2022} applied their translation to six semantic flavors to then derive an estimate of the frequency of each of those, as seen in Table~\ref{tab:indefinites-prior}.  We will return to the topic of estimating communicative need distributions in Section~\ref{sec:communication-discussion}.

\begin{table}
	\centering
	\begin{tabular}{ll}
		\toprule
		Semantic flavor & Prior probability
		\\
		\midrule
		specific known & 0.08
		\\
		specific unknown & 0.08
		\\ 
		non-specific & 0.26
		\\
		negative polarity & 0.33 
		\\
		free choice & 0.1 
		\\
		negative indefinite & 0.15
		\\
		\bottomrule
	\end{tabular}
	\caption{Prior probability distribution over flavors, as estimated from the corpus in \citealt{beekhuizenSemanticTypologyParallel2017}.}
	\label{tab:indefinites-prior}
\end{table}

\begin{figure}[ht!]
	\centering
	\includegraphics[width=0.49\textwidth]{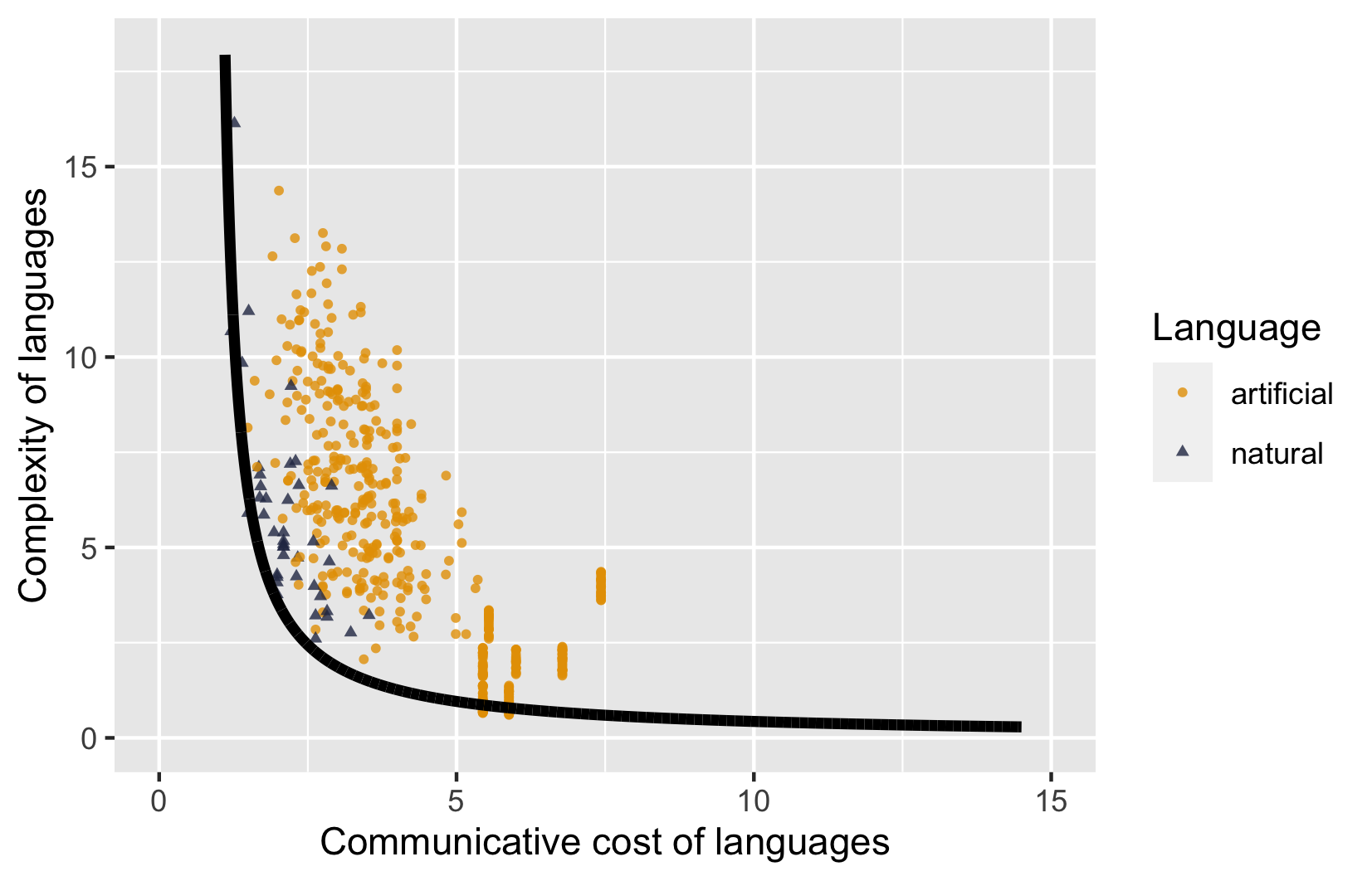}
	\includegraphics[width=0.49\textwidth]{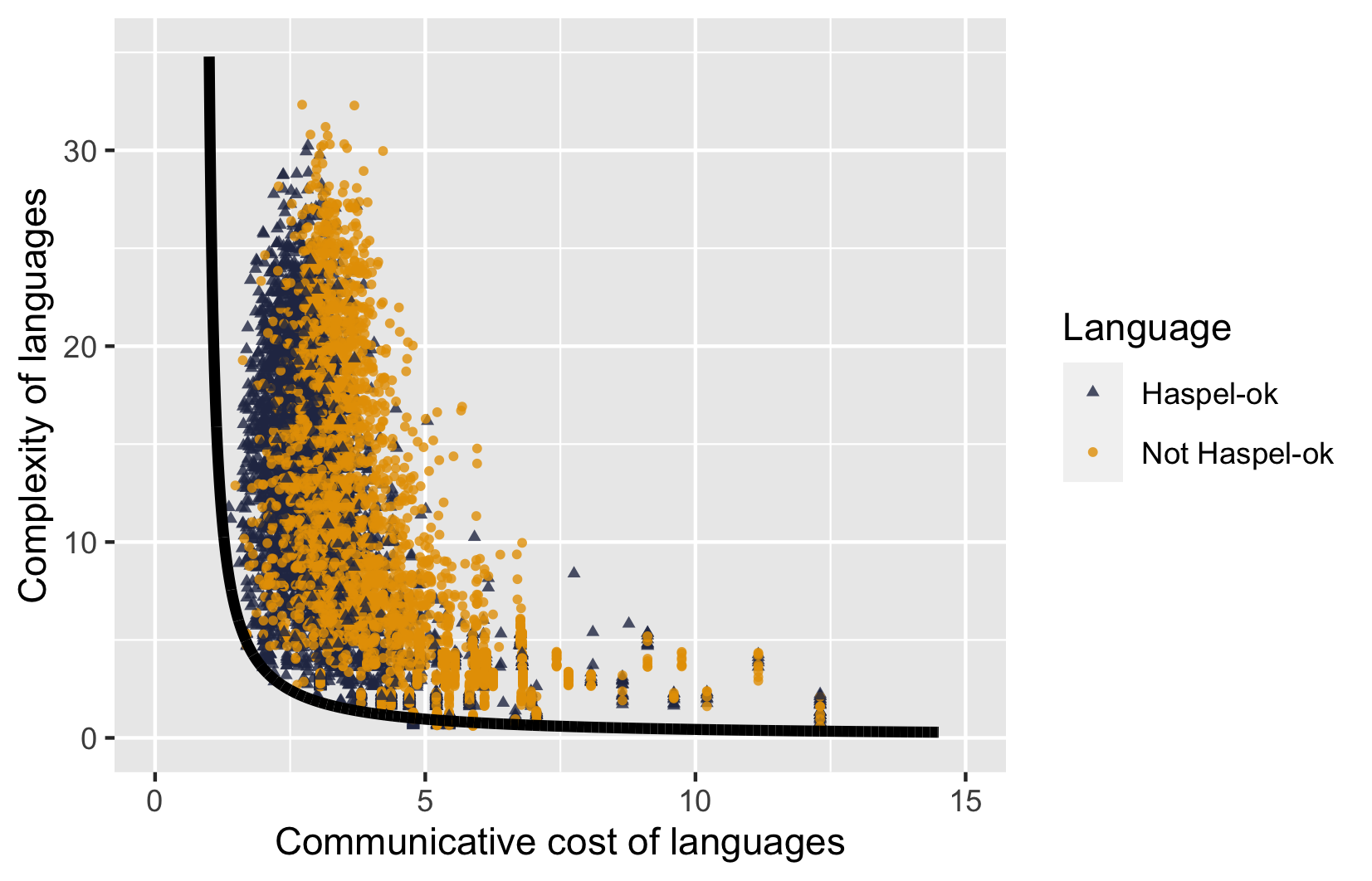}
	\caption{Results for the two experiments.  Left: natural languages (grey triangles) are closer to the Pareto frontier than non-natural languages.  Right: languages which satisfy Haspelmath's universals (grey triangles) are also closer to the Pareto frontier than ones which do not.}
	\label{fig:indefinites-effcomm}
	\addsource{\cite{denicIndefinitePronounsOptimize2022}}
\end{figure}

Putting all of these ingredients together, Figure~\ref{fig:indefinites-effcomm} shows the results of the two experiments.  The left panel shows that natural languages (grey triangles) are closer to the Pareto frontier than non-natural languages.  This result is especially compelling because \citeauthor{denicIndefinitePronounsOptimize2022} controlled the artificial languages to, on average, have the same \emph{coverage} (how much of the semantic space a language can express) and \emph{overlap} (how often multiple terms can be used for the same point in the semantic space) as the natural languages. Please consult the paper for more implementation details. The right panel shows that languages which satisfy Haspelmath's universals (grey triangles) are also closer to the Pareto frontier than ones which do not.

This constitutes further evidence for \textsc{The Efficient Communication Hypothesis}: languages which do better at balancing between the competing pressures to be simple and to be useful have the relevant lexical property universals.  \cite{denicIndefinitePronounsOptimize2022} also compare a language of thought that includes set difference, as well as incorporating \emph{pragmatic} speakers and listeners for $P_S$ and $P_R$, and find that the general results described above continue to hold, suggesting that they are quite robust to particular design choices.

\subsection{Modals}
\label{sec:modals}

The final case study for efficient communication that we will discuss in detail concerns \emph{modals}.  These are expressions like English \emph{must}, \emph{might}, and \emph{can} which are used to talk about situations other than the here-and-now, thus exhibiting one of \citeauthor{hockettOriginSpeech1960}'s design features of language (displacement; \citealp{hockettOriginSpeech1960}).  These expressions have been studied extensively in the formal semantics literature \citep{kratzerNotionalCategoryModality1981,kratzerModality1991,matthewsonModality2019} and have also begun to be explored cross-linguistically \citep{vanderklokModalityUnderdescribedLanguages2022,uegakiCrosslinguisticDatasetForceflavor2024,guoDatabaseModalSemantic2022}.  The exposition in this section closely follows \citet{imelEfficientCommunicationAnalysis2026}, to which we refer the reader for more details.

\subsubsection{Semantic Universal: Independence of Force and Flavor}

To develop a semantic universal in this domain, we must first understand the ways in which languages vary in how they encode modal meanings. Since at least \citet{kratzerNotionalCategoryModality1981}, the semantics of modals have been explicated in terms of two axes of variation: \textbf{force} and \textbf{flavor}.  These axes can be illustrated with the following examples.

\begin{examples}
\item \begin{examples}
			\item \label{must-epi} [Context: a friend walks in and shakes off a wet umbrella.  You say:]
			
					It \emph{must} be raining.

			\item \label{must-deon} [Context: you are reading the specifications of a homework assignment.  It partially reads:]
			
					You \emph{must} upload your homework as a PDF.
		\end{examples}
\item \begin{examples} \label{may}
			\item \label{may-epi} [Context: a friend is leaving and grabs an umbrella on the way out, saying:]
			
					It \emph{may} be raining.
			\item \label{may-deon} [Context: a mother offers a treat to a child for finishing an assignment:]
			
					You \emph{may} have a cookie.
		\end{examples}
\end{examples}

The \emph{must} examples exhibit strong (i.e.\ universal) force, but differ in flavor.  For example, \exref{must-epi} can be glossed as saying: all of the worlds compatible with my evidence are worlds in which it is raining. The universal quantification represents the force, and the domain of worlds (those compatible with my evidence) the flavor, in this case epistemic.  \exref{must-deon} exhibits universal force with deontic flavor, roughly saying that all the worlds in which you follow the rules are ones in which you upload a PDF.  The examples with \emph{may} in \exref{may} exhibit weak (i.e.\ possibility) force: their meaning says that some world satisfies the prejacent.  \exref{may-epi} and \exref{may-deon} again differ in flavor, with the former being epistemic and the latter being deontic.

In addition to epistemic and deontic flavors, many others have been identified: bouletic (worlds in which desires are fulfilled), teleological (worlds in which goals are satisfied), et cetera.  Similarly, there are arguably more forces than just weak and strong: for instance, there are weak necessity modals (e.g., \emph{should}, \emph{ought}) which intuitively express universal quantification over a smaller domain of worlds \citep{fintelHowSayOught2008}. See \citealt{matthewsonModality2019} and references therein for further discussion of these two axes.

The examples above show that English modals lexically specify modal force (each modal has a fixed quantificational force) but exhibit variability across flavors (the modals can express more than one flavor).  We note that such variability does not require that all modals in English can express all flavors: for instance, \emph{might} arguably can only be used epistemically.  Kratzerian semantics for modals capture this by hard-coding quantificational force into the meaning of a modal but relying on context to determine the flavor.\footnote{Typical implementations determine the flavor as the product of two further parameters: a modal base and an ordering source.  We set aside this distinction for present purposes.}

Not all languages are like English: some exhibit so-called \emph{variable force modals}, which specify flavor but do not lexically encode force.  This has been found at least in St'\'at'imcets \citep{Rullmann2008}, Nez Perce \citep{Deal2011}, Old English \citep{Yanovich2016}, and Pintupi-Luritja \citep{Gray2021}.\footnote{We will discuss modals that specify neither force nor flavor in the next section.} We illustrate the phenomenon with elicited examples of St'\'at'imcets \emph{k'a} from \citealt{Rullmann2008}:\footnote{These are examples (5c) and and (5e) from \citealt[p. 321]{Rullmann2008}.  See their footnote 5 on p. 320 for the abbreviations.}

\begin{examples}
	\item \begin{examples}
		\item \label{ka-strong} [Context: You have a headache that won't go away, so you go to the doctor. All the tests show negative. There is nothing wrong, so it must just be tension.]

			\gll nilh \emph{k'a} lh(el)-(t)-en-s-w\'a(7)-(a) ptinus-em-s\'ut \\
			\textsc{foc} \textsc{infer} from-\textsc{det}-\textsc{1sg}.\textsc{poss}-\textsc{nom}-\textsc{impf}-\textsc{det} think-\textsc{mid}-\textsc{ooc}
			\\
			\trans `It \emph{must} be from my worrying.'

		\item \label{ka-weak} [Context: His car isn't there.]

			\gll plan k'a qwats\'ats \\
			already \textsc{infer} leave \\
			\trans `Maybe he's already gone.'
	\end{examples}
\end{examples}

\noindent \exref{ka-strong} shows \emph{k'a} being used with strong force and epistemic flavor. \exref{ka-weak} shows \emph{k'a} being used with weak force and epistemic flavor.  Further analysis in \cite{Rullmann2008} shows that \emph{k'a} can only be used with epistemic flavor, so it is an example with lexically specified flavor but variable force. The discussed semantic variation across English and St'\'at'imcets is summarized by Table \ref{fig:must_vs_ka}.

\begin{table}[h]
  \resizebox{.475\linewidth}{!}{
\centering
\begin{tabular}{lccc}
     \multicolumn{4}{c}{St'\'at'imcets \emph{k'a}}\\
    \toprule
    & epistemic & deontic & $\dots$
    \\
    \midrule
    weak & $\checkmark$ & & 
    \\
    strong & $\checkmark$ & & 
    \\
    \bottomrule
\end{tabular}
}
\hfill
\resizebox{.475\linewidth}{!}{
\centering
    \begin{tabular}{lccc}
        \multicolumn{4}{c}{English \emph{must}}\\
    \toprule
    & epistemic & deontic & $\dots$
    \\
    \midrule
    weak & &  &  
    \\
    strong & $\checkmark$ & $\checkmark$ & 
    \\
    \bottomrule
  \end{tabular}
  }
  \caption{Two kinds of modal semantic underspecification: \emph{variable-force} and \emph{variable-flavor}.}
  \label{fig:must_vs_ka}
\end{table}

Even with this substantial cross-linguistic variation, there are kinds of modal meanings that are unattested. We can describe potential restrictions on modal meanings as semantic universals. In order to state universals for modals in a relatively theory-neutral manner (i.e.\ in a way that does not presuppose a particular formal semantic implementation), we make the following assumptions.  We assume that force and flavor are fundamentally properties of contexts of use.  This reflects current practice in semantic fieldwork as applied to modality \citep{matthewsonMethodologySemanticFieldwork2004,bochnakTechniquesComplexSemantic2020,vanderklokRevisedModalQuestionnaire2021}.\footnote{In addition to the particular studies already mentioned, see \citealt{Matthewson2013,Cable2017} for more examples of the application of these methods.}  For example, the modal questionnaire of \citealt{vanderklokRevisedModalQuestionnaire2021} consists exactly of discourse contexts designed to isolate a single force-flavor pair.  These contexts can subsequently be used for tasks like elicitation, translation, and acceptability.  Finally, we will say that a modal $M$ \emph{can express} a force-flavor pair just in case a bare positive sentence of the form $Mp$ is judged felicitous in a context with that pair.\footnote{We intend `judged felicitous' to also include the case where such sentences are produced naturally in elicitation tasks, as well as when such sentences are found in naturally-occuring contexts which have a clear force-flavor pair.} 

At this level of generality, we will represent the meaning of a modal as being a set of force-flavor pairs.  The semantic universals that we will discuss will be constraints on what kinds of meanings (sets of such pairs) are attested in the languages of the world.  For notation, for a modal $m$, let $\sv{m}$ be the set of force-flavor pairs it can express.  Furthermore, we will write $\force(m) = \{ \force \mid \exists \flavor \text{ s.t. } (\force , \flavor) \in \sv{m} \}$ for the set of forces that a modal $m$ can express and \emph{mutatis mutandis} for $\flavor(m)$ and the set of flavors.

We adopt this level of generality because it avoids commitment to the exact formal semantics of these expressions, which is often still being debated.  For example, we can say that a \emph{variable force modal} is one that can express more than one pair with the same flavor (i.e.\ for which $|\force(m)| > 1$).  This is useful because there are two broad approaches to the semantics of such variable force modals: they actually encode existential quantification but lack an explicit universal counterpart in the language \citep{Deal2011} or they encode universal quantification but rely on some mechanism of domain restriction \citep{Rullmann2008, Bochnak2015b, Mocnik2019}.  On such analyses, the underlying semantics contains one specific quantifier; in the present setting, they will still be considered variable force since bare positive sentences are used in contexts with multiple forces.

With these preliminaries in place, we can now state a semantic universal---indeed, a lexical property universal---for modals first formulated by \citet{steinert-threlkeldSemanticUniversalModality2023}:\footnote{This universal generalizes an earlier proposal of \citet{Nauze2008}.}
\begin{description}
  \item[\textnormal{\textsc{The Independence of Force and Flavor (IFF) Universal}}:] All modals in natural language satisfy the independence of force and flavor property: if a modal can express the pairs $(\force_1, \flavor_1)$ and $(\force_2, \flavor_2)$, then it can also express $(\force_1, \flavor_2)$ and $(\force_2, \flavor_1)$.

  [Alternative formulation: a modal $m$ satisfies the IFF property just in case $\sv{m} = \force(m) \times \flavor(m)$, where $\times$ is the Cartesian product.]
\end{description}

This universal captures the guiding idea from \cite{kratzerNotionalCategoryModality1981} and much subsequent theorizing on the semantics of modals that force and flavor are \emph{independent} axes of meaning, while at the same time still placing substantive constraints on what meanings can be expressed.  For example: meanings with `arbitrary' combinations of force and flavor (e.g.\ a modal that can only express strong epistemic and weak deontic meanings) are ruled out by this universal.  Although the \textsc{IFF} universal is formulated in a way that is specific to modal meanings, \citet{steinert-threlkeldSemanticUniversalModality2023} note that it can be viewed as a form of convexity when applied to discrete meaning spaces.

\subsubsection{Efficient Communication Analysis of Modals}

With this lexical property universal in place, we have a target for explanation in terms of efficient communication.  \citet{imelEfficientCommunicationAnalysis2026} execute such an analysis, using the ingredients that we have been discussing throughout this chapter.  The meaning space consists of all force-flavor combinations from weak and strong force and the epistemic, deontic, and circumstantial flavors.  This particular combination was chosen because it is the largest one for which a large body of cross-linguistic data is available.  To that end, \citeauthor{imelEfficientCommunicationAnalysis2026} combined data from the large cross-linguistic survey of \citet{uegakiCrosslinguisticDatasetForceflavor2024} together with semantic fieldwork data gathered from the published literature, resulting in a total of 27 languages.  We refer the reader to \citet{imelEfficientCommunicationAnalysis2026} for more details on the data.

Communicative cost was measured as described in Section~\ref{sec:complexity-cost}, with a prior distribution over force-flavor pairs extracted from a corpus annotated for modal flavor information \citep{pyatkinPossiblePlausibleDesirable2021}.  Being an English corpus, where force is lexically specified, the authors added force information based on the specific modal annotated.  The full prior is given in Table~\ref{tab:prior-modals}.

\begin{table}[h]
  \centering\begin{tabular}{cccc}
    \toprule
    & epistemic & deontic & circumstantial \\
    \midrule
    weak   & 0.139 & 0.042 & 0.143 \\
    strong & 0.104 & 0.254 & 0.318 \\
    \bottomrule
  \end{tabular}
  \caption{Estimated probability distribution over force-flavor pairs, representing a communicative need distribution over modal meanings.}
  \label{tab:prior-modals}
\end{table} 

The utility function in communicative success gave partial credit: half a point each for getting the force and the flavor right (though the authors also presented comparisons with all-or-nothing utility with qualitatively similar results).

Complexity was measured, once again, via minimum expression length in a logical language of thought.  \citet{imelEfficientCommunicationAnalysis2026} used a basic propositional language with atoms for each force and each flavor, with operations for conjunction, disjunction, and negation.  Table~\ref{tab:modal-complexity} shows the meaning representations, shortest LOT formulae, and resulting complexities for English \emph{may} and two hypothetical modals \emph{mought} and \emph{notcirc}.  The former is a hypothetical modal expressing only epistemic possibility and deontic necessity, a combination that is ruled out by the \textsc{IFF} universal.  The latter is a hypothetical modal expressing all force-flavor combinations except circumstantial ones; this one satisfies \textsc{IFF} but is not maximally underspecified.

\begingroup
\setlength{\tabcolsep}{12pt} %
\begin{table}[ht]
    \scriptsize
    \centering
    \resizebox{\textwidth}{!}{
    \begin{tabular}{cccc}
        \toprule
         \textbf{Modal} & \textbf{Meaning representation} & \textbf{Shortest Formula in LOT} & \textbf{Complexity (\# of atoms)} 
         \\ \midrule 
         \textit{may}   
            & 
            \begin{tabular}{lcccc}
			& $e$ & $d$ & $c$ & $t$
			\\
			$w$ & $\checkmark$ & $\checkmark$ &  &
			\\
			$s$ & & & &
		    \end{tabular} 
		&
		$w \wedge (e \vee d)$ 
		&
		3
		\\ \midrule[0.6\lightrulewidth]
         \textit{mought}   
            & 
            \begin{tabular}{lcccc}
			& $e$ & $d$ & $c$ & $t$
			\\
			$w$ & $\checkmark$ & & &
			\\
			$s$ & & $\checkmark$ & &
		    \end{tabular} 
		&
		$(w \wedge e) \vee (s \wedge d)$ 
		&
		4
		\\ \midrule[0.6\lightrulewidth]
		\textit{notcirc}
            & 
            \begin{tabular}{lcccc}
			& $e$ & $d$ & $c$ & $t$
			\\
			$w$ & $\checkmark$ &$\checkmark$ & & $\checkmark$
			\\
			$s$ & $\checkmark$ & $\checkmark$ & & $\checkmark$
		    \end{tabular}
		&
		$ \neg c $ 
		&
		1
        \\ \bottomrule
    \end{tabular}
	}
    \caption{Measuring complexity for English \emph{may} and two hypothetical modals \emph{mought} and \emph{notcirc}.  First column: meaning representation.  Here: $w$ and $s$ stand for weak and strong force, respectively, and $e$, $d$, $c$, and $t$ for the flavors epistemic, deontic, circumstantial, and teleological. Second column: shortest LOT formula.  Third column: complexity.}
    \label{tab:modal-complexity}
\end{table}
\endgroup

\begin{figure}[ht!]
  \centering
  \includegraphics[width=\textwidth]{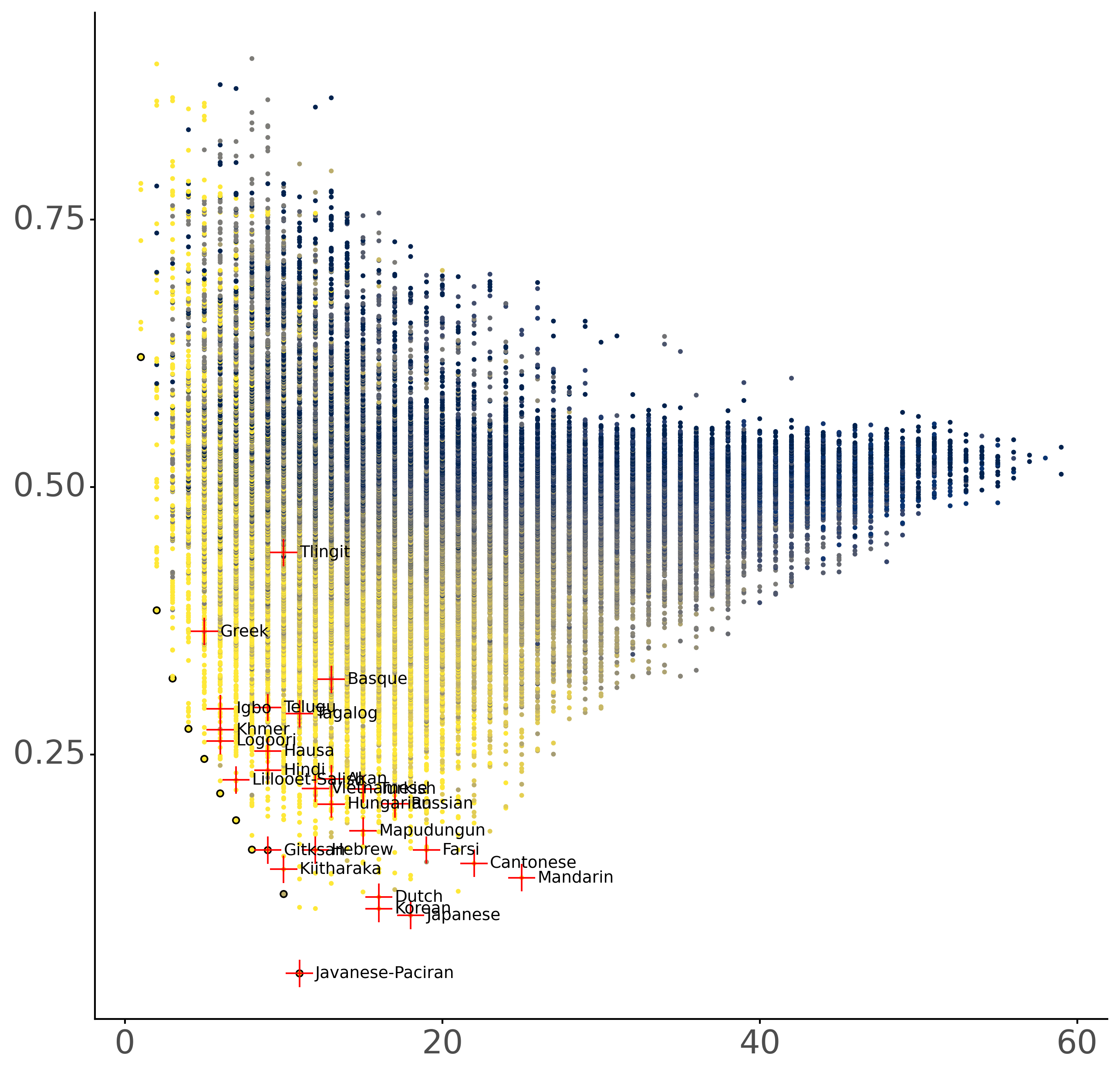}
  \caption{The complexity/communicative cost trade-off for modal languages. The black circles constitute the estimated Pareto frontier of optimal solutions to the trade-off.
  Red `+' are natural languages obtained from \cite{guoDatabaseModalSemantic2022} and are labeled. The color of a language is its degree of naturalness (proportion of modals satisfying the IFF universal).}
  \addsource{\cite{imelEfficientCommunicationAnalysis2026}}
  \label{fig:modals-effcomm}
\end{figure}

With all of these ingredients in place, we can present the primary results from \citet{imelEfficientCommunicationAnalysis2026}, which are shown in Figure~\ref{fig:modals-effcomm}.  The $x$-axis shows communicative cost, and the $y$-axis shows complexity.  Natural languages are shown as red crosses, and the Pareto frontier of optimal languages is indicated by black circles.  Color here corresponds to a `degree of \textsc{IFF}-ness', which is the proportion of modals in a language that satisfy the \textsc{IFF} universal.  %

\citet{imelEfficientCommunicationAnalysis2026} draw three primary conclusions from this figure and resulting analyses.  (Their paper also contains many follow-up analyses on the robustness of these results to variations in the parameters of measuring communicative cost and an alternative complexity measure as well.) One: all Pareto optimal languages have degree-1 \textsc{IFF}, i.e.\ all modals in all optimal languages satisfy the lexical property universal.  Two: natural languages are more optimal than non-natural languages.  Three: the degree of \textsc{IFF}-ness is significantly correlated with optimality, as seen by the `yellow-ness' of languages increasing as one nears the Pareto frontier.  Taken together, these results provide evidence for \textsc{The Efficient Communication Hypothesis}: languages which do better at balancing between the competing pressures to be simple and to be useful also produce more modals that satisfy the IFF universal.

\subsection{Information Bottleneck}
\label{sec:ib}

The previous case studies have all relied on different instantiations of the same general framework for operationalizing communicative cost and complexity: the former is inversely related to communicative success, and the latter is related to minimum expression length in a logical language of thought.  In this section, we briefly outline an information-theoretic framework for operationalizing the \textsc{Efficient Communication Hypothesis} and discuss how it relates to the previous analyses.

In particular, the Information Bottleneck (IB) method---a general framework for analyzing efficient compression of information \citep{tishbyInformationBottleneckMethod1999}---has been applied to semantic typology in domains ranging from color terms \citep{zaslavskyEfficientCompressionColor2018} to person systems \citep{zaslavskyLetsTalkEfficiently2021}, spatial demonstratives \citep{chenInformationtheoreticApproachTypology2023}, and grammatical markers like tense \citep{mollicaFormsMeaningsGrammatical2021}.  In the general method, information about one variable $X$ is compressed into a `codebook'---a set of categories, roughly---in a way that preserves as much information as possible about another variable $Y$.  

In the case of semantic typology, this gets operationalized as follows: the variable $X$ is a set of meanings $M$, which a sender need to communicate; the variable $Y$ is a set $U$ (for universe) of points/referents in a domain that a listener is trying to identify; the codebook is a set of words $W$ that the sender uses to express meanings $M$.  The modeler must specify a joint distribution $p(m, u)$, which is typically factored as $p(m)p(u|m)$.  The first term, $p(m)$, is the communicative need distribution, playing the same role as the prior in the previous case studies that we have described.  The second term, $p(u|m)$, is typically designed to reflect speaker uncertainty about the underlying universe: meanings are probability distributions over points/referents, representing uncertainty of a true underlying state.

In the color case, for example, the universe $U$ a set of 330 chips called the Munsell chips, which were used in a World Color Survey to elicit color term data cross-linguistically \citep{berlinBasicColorTerms1969, Kay2010}.  Meanings are distributions over $U$.  Each color chip $u$ is associated with a meaning $m_u$, itself a normal distribution centered on $u$ with a variance intended to reflect human perceptual discriminability.  There are various methods for estimating the communicative need distribution $p(m)$, which we will discuss in more detail in \S\ref{sec:communication-discussion}.  

Communicative cost and complexity are then operationalized in terms of mutual information.  In particular, complexity corresponds to $I(M; W)$---the information between meanings and the words---and communicative success corresponds to $I(W; U)$---the information between words and points in the universe.  The trade-off between these two is modeled by introducing a parameter $\beta$ that controls the relative importance of the two, and then optimizing the following objective function:
\begin{equation}
	\label{eq:ib-effcomm}
	\mathcal{F}_\beta := I(M; W) - \beta I(W; U)
\end{equation}
In the language of efficient compression mentioned above: a `speaker' aims to compress meanings $M$ into words $W$ in a way that preserves as much information as possible about the universe $U$.  The trade-off parameter $\beta$ controls the relative importance of these two goals: low $\beta$ corresponds to systems with a single word used for all meanings (simple, but not useful), while $\beta \to \infty$ corresponds to systems with a unique word for each meaning (useful, but not simple).

For more concrete details on this method, its implementation, and applications in semantic typology, see \cite{mollicaInformationtheoreticMachineLearning2025} and references therein.

\subsection{Discussion}
\label{sec:communication-discussion}

To summarize this chapter: we have presented several case studies which provide evidence for \textsc{The Efficient Communication Hypothesis}: languages which optimally balance complexity and communicative cost have expressions which satisfy lexical property universals.  We operationalized cost in terms of inverse iterated communicative success and complexity in terms of minimum expression length in a logical language of thought and applied these notions to universals in the domains of quantifiers, indefinite pronouns, and modals.  We then briefly sketched an alternative operationalization of cost and complexity---the Information Bottleneck---which has also been successfully applied in several empirical domains.

Several exciting avenues for present and ongoing work arise from abstracting from and comparing all of these case studies.

On the one hand, systematically comparing and contrasting the methods detailed in this chapter with the IB approach would be fruitful.  To that end, \cite{bruneau-bongardAssessingPressuresShaping2025} have shown that the IB complexity measure has some unusual properties, e.g.\ that it often does not increase when explicit synonyms are added to a language. They have also shown that jointly optimizing more pressures than merely IB cost and complexity can fit typological data better than those two alone.  Relatedly, for a system in which each meaning is always expressed by a single word, IB complexity depends only on how often each word is used, not on the shape of the categories: a convex and a gerrymandered partition with the same usage frequencies are equally complex. Structural properties such as convexity therefore enter IB analyses through the accuracy term and the meaning distributions $p(u|m)$, rather than through complexity. %
\citet{imelEfficientCommunicationAnalysis2026} also analyze the modal data described in Section~\ref{sec:modals} using the IB measure of communicative cost, but with the same complexity measure as above, and found broadly similar results to the original analysis.  These studies invite controlled, systematic comparison of myriad ways of operationalizing cost and complexity to answer questions like whether some combinations fit a wider range of data than others.

Furthermore, a small flurry of recent work has investigated the `meta-universal' of convexity (\S~\ref{sec:convexity-learning}) from the perspective of efficient communication.  \cite{koshevoyConvexityProbablyMakes2025} show that the degree of convexity measure from \cite{steinert-threlkeldEaseLearningExplains2020} (as reported in \S~\ref{sec:convexity-learning}) is highly correlated with distance to the Pareto frontier, using a distinct operationalization of complexity and cost from the ones mentioned in this chapter.  They also show that convexity predicts this degree of optimality better than either individual IB measure. \cite{imelConvexityEfficiencySemantic2026} use the same convexity measure but study it from the IB operationalization directly.  They show that convexity and efficiency are logically distinct, in the sense that neither entails the other; they further find that efficiency is the stronger predictor for discriminating attested color naming systems from hypothetical variants, with convexity adding little on top.  Finally, \cite{ranjanWhenEfficientCommunication2026} generalize the definition of degree of convexity to the probabilistic setting of the IB and study both color naming as well as a range of artificial IB environments in order to identify the circumstances under which convexity correlates with IB efficiency.  They identify the convexity of the communicative need distribution $p(m)$ and the uniqueness of meanings as two key factors.  All of this work aims to enrich our understanding of when and why efficiency can explain interesting semantic universals.

While several of the case studies described above have analyzed robustness with respect to several modeling parameters, one parameter with significant room for further development is the communicative need distribution (i.e.\ $P(p)$ in our notation, $p(m)$ in the IB setting).  This distribution can have a significant effect on efficient communication analyses, since higher-need points are weighted higher in the communicative cost term (and also in complexity in the IB setting).  Aside from using a uniform distribution for need, approaches to empirical estimation often rely on explicit naming data \citep{zaslavskyEfficientCompressionColor2018,zaslavskyCommunicativeNeedColour2020} or corpus frequency \citep{kempKinshipCategoriesLanguages2012, denicComplexityInformativenessTradeoff2020,imelConvexityEfficiencySemantic2026}.  \citet{zaslavskyCommunicativeNeedColour2020} found, in color naming, that priors based on explicit naming data averaged across many languages and a prior based on English corpus frequency performed equally well and better than others based on image statistics.  One major open avenue addresses the language- and culture-specificity of communicative need: in all of the analyses discussed here (even those estimating need from multilingual data), communicative need is assumed to be constant across all languages.  Developing methods for estimating culture-specific need distributions may help analyze which aspects of semantic universals and typological patterns are driven by shared cognitive and communicative principles and which are culturally shaped.

One final avenue for efficient communication analyses focuses on extending the approach beyond the lexicon and the lexical property universals that we have discussed here.  In theory, one would like to analyze the language as a complete system, including word meanings and larger grammatical units, as being subjected to the trade-off between simplicity and usefulness.  Some steps have been made in this direction.  \citet{mollicaFormsMeaningsGrammatical2021} analyze the length of linguistic forms (grammatical markers for tense, number, and evidentials) in an IB-like setting, supporting the idea that shorter forms are attached to more frequent meanings.  \citet{denicRecursiveNumeralSystems2024} analyze \emph{recursive} numeral systems---which are all equally and maximally informative---as trading off between lexicon size and average (across communicative need) morphosyntactic complexity of the numerals, using language-specific grammars for numeral systems.\footnote{\citet{carcassiAssertionDenialEvolution2023a} use a similar notion of usage complexity in analyzing the inventories of logical operators cross-linguistically.} \citet{yang2025re} confirm this result with a grammar framework that generates natural and artificial systems in the same way, which makes the two directly comparable. \citet{prasertsomRecursiveNumeralSystems2026} argue that explicitly incorporating regularity across forms helps explanations in this domain.  Extending these approaches to analyze the richer form-meaning mapping in natural language should enable richer explanations of language structure, including the division of labor between the lexicon and the grammar, multiple different mechanisms for expressing the same meaning, and the interaction between different conceptual domains that a language is used to discuss.

\pagebreak

\section{Discussion}
\label{chapter:discussion}

To summarize what we have shown so far: after introducing lexical property universals and a framework for analyzing computational explanations thereof (Chapter~\ref{chapter:introduction}), we took a deeper look at explanations of several different universals in terms of two different optimization principles: ease of learning (Chapter~\ref{chapter:learning}) and efficient communication (Chapter~\ref{chapter:effcomm}).  While each of those chapters concluded with some discussion local to each family of explanations, we will use this chapter to catalog some outstanding issues and directions for future work.  In the final section, we discuss steps towards a principled taxonomy of explanatory mechanisms for semantic typology.

\subsection{Adjudicating between explanations} 

In addition to seeing explanations of many different universals in this Element in terms of two optimization principles, some universals---e.g.\ monotonicity and convexity---have received treatments in terms of \emph{both} ease of learning and efficient communication.  This points to a hard and deep problem: how can one adjudicate between multiple competing explanations for a given semantic universal?

The difficulty raised by this question reflects the status of the majority of the computational explanations in this Element: they can plausibly be construed as \emph{how-possibly} explanations \citep{drayLawsExplanationHistory1957, drayExplainingHowPossibly1968, bokulichHowTigerBush2014, rohwerHypotheticalPatternIdealization2013}.  
Such explanations aim to isolate possible causal factors that could give rise to an observed phenomenon of interest, without isolating the true causal mechanism.  In the context of semantic universals, such how-possibly explanations serve a crucial epistemic role: they show that it is possible for domain-general cognitive pressures to give rise to the presence of these universals.  It is thus not necessary to posit them as innate features of the human linguistic system.

Despite this positive epistemic role, and despite the fact that it may never be possible to provide complete causal explanations in this domain, there are still general theoretical virtues that can help with the adjudication problem.  For example, the virtue of unification suggests that if one explanatory framework, with as few domain-specific assumptions as possible, can explain many universals, that should be preferred to more bespoke, case-specific explanations.  This consideration alone, however, is not necessarily powerful enough in our scenario since different frameworks tend to differ in where they put case-specific assumptions (e.g.\ the primitives in a language of thought versus the joint distribution in the IB system).  Other virtues like simplicity can be of help here as well.  Similarly, an ambitious future direction is to identify the shared structure of the multiple explanations, which has often proved fruitful in dealing with multiple inconsistent models of a given phenomenon \citep{lisciandraMultipleModelsOne2021}.

\subsection{Diachronic explanations}
\label{sec:diachronic-explanations}

While the explanations offered in this Element have been couched in the language of optimization, they have for the most part been entirely synchronic: they show that languages which satisfy lexical property universals look as if they have been optimized for ease of learning and/or for efficient communication.  A natural next step seeks diachronic explanations, that more systematically incorporate the optimization process itself, in addition to ``the final step'' of the process.

Diachronic explanations can take both empirical and computational shape.  To the former end, \citet{zaslavskyEvolutionColorNaming2022} gather diachronic data from one language---Nafaanra, spoken in Ghana and Côte d'Ivoire---from two points 40 years apart and show that while the language has changed, including with the addition of more and finer-grained color terms, it has remained nearly optimal in the efficient communication sense. 
If language change were not actively subject to a pressure for efficiency, one may expect the language to possibly drift away from the Pareto frontier, contra the observation.  To the latter end, \citet{imelAgentbasedImitationDynamics2026} build an agent-based model incorporating signaling games with an evolutionary dynamic of imprecise imitation and show through simulation that populations of agents under this dynamic converge to efficient numerosity category systems.  This provides another kind of how-possibly explanation, but a diachronic one.

Such explanations also exist in the ease of learning framework, where iterated learning explicitly models the process of learning a language across generations from parents \citep{kirbyIteratedLearningEvolution2014, smithHowCultureBiology2020, smithIteratedLearning2025}.  This process enables the learning biases of the learner to impact the language at each generation.  While much work in this paradigm focuses on the emergence of regular structure in form-meaning mappings \citep{kirbyCumulativeCulturalEvolution2008a}, it has also been recently applied to semantic universals.  As discussed in Section~\ref{sec:iterated-learning}, \cite{carcassiMonotoneQuantifiersEmerge2021b} show that this process can produce monotone quantifiers, suggesting that iterated learning pressures can shape the structure of semantic categories in addition to the form-meaning mapping.

Furthermore, the two approaches integrate very naturally: ease of learning plausibly plays a role in shaping language at the point of inter-generational transmission, while efficient communication plays a role during active communication within a generation.  Indeed, in toy experimental settings, \cite{kirby2015compression} suggest that pressures both for learning and for communication are necessary for the emergence of rich linguistic structure.  Similarly, \cite{smithCommunicativePressuresShape2025} show that use in communication, but not learning on its own, is required to produce communicatively efficient systems for differential case marking.  \citet{carlssonCulturalEvolutionIterated2024} show that iterated learning and communication together produce more human-like systems than either force on its own (although both forces can produce IB-efficient systems on their own, the combination of both provides the most human-like systems). While natural language learning and use in the wild do not exhibit a clean separation into within-generation use and across-generation transmission, this work suggests that empirical and modeling approaches which explicitly incorporate pressures from both efficient communication and ease of learning are likely to produce the most satisfying diachronic explanations.

\subsection{Beyond the lexicon}

Every universal in this Element has the form of the Lexical Property Universal Schema: all 
lexical items of a given kind have meanings with a given property.
The schema looks at items one at a time. But much of what languages do with
meaning is done by complex expressions, and some of the most robust
regularities concern how a system of simple and complex expressions is put
together. Numerals provide one particularly clear example, and they show both what the
methods of the previous chapters can do beyond the lexicon and where they
need to be extended.  In particular, they exhibit system-level properties, but are often indistinguishable by communicative cost, raising issues about additional pressures and the division of labor between the lexicon and composition.  We turn to these points in turn in the remainder of this subsection. 

\paragraph{Numerals as a system}

A recursive numeral system such as the English one contains a handful of
simple words (\emph{one}, \emph{two}, \ldots, \emph{ten}, \emph{hundred})
and a grammar that combines them to name every number in a large range:
\emph{twenty-three} is $2\times10+3$, \emph{three hundred and five} is
$3\times100+5$. The cross-linguistic regularities in this domain are well
documented \citep{Greenberg-1978, hurford2011linguistic}. They concern
which numbers serve as bases, how multiplication and addition are ordered,
and which numbers receive simple words. Decimal (base-10) systems are by far the most
common, followed by vigesimal (base-20) ones \citep{wals-131}, a preference
usually traced to counting on fingers and toes rather than to anything in
arithmetic. Statements
like these are not about the meaning of any single expression. They are about
the division of labor between a small lexicon and the grammar that extends
it.

Numerals are also a domain whose meaning space is unusually well understood.
The meanings are numbers, and the representation of number does not depend
on having number words: infants and many animals discriminate approximate
quantities in a common, well-understood way \citep{Dehaene1997, feigenson2004core}. In the terms of the next section,
number is a domain whose structure is given from outside language. This
makes numerals a good test of how far the explanations of the previous
chapters reach, because the meaning side can be held fixed with some
confidence.
 
\paragraph{When informativeness runs out}
Recall that the \textsc{Efficient Communication Hypothesis} trades off two terms:
complexity and communicative cost (\S1.4). For many numeral systems this
works much as in Chapter~3. Systems with only a few approximate terms, or with
exact terms for small numbers only, lie close to the frontier of the trade-off
between simplicity and informativeness \citep{xu2020numeral}. Recursive
systems are different. They name every number in their range exactly, so all
of them are maximally informative and communicative cost cannot distinguish
one from another. If an efficiency explanation is to work here, the
explanatory weight has to move to a factor beyond informativity.  Plausibly, this requires moving from the meanings to the forms.

\citet{denicRecursiveNumeralSystems2024} took this step. They measured the complexity of a
numeral system in two parts: the size of its lexicon, and the average
morphosyntactic complexity of its numeral expressions, weighted by how often
each number needs to be expressed. A system with very few simple words needs
long expressions for most numbers; a system with a word for every number needs
no composition at all but has an enormous lexicon. Attested systems turned out
to trade off these two quantities nearly as well as the best of a large set of
artificial systems. \citet{yang2025re} confirmed this result with a
grammar framework that generates natural and artificial systems in the same
way, which matters because a comparison class built differently from the
natural systems can make the natural ones look better or worse than they are.
Showing that \emph{only} natural-like systems optimize this trade-off has,
however, proved harder. \citet{prasertsomRecursiveNumeralSystems2026} argue that this is
because the trade-off leaves out a property of the system as a whole: the
regularity of its grammar, that is, how much the same sub-forms are reused
across the number range. Measured by minimum description length, regularity
and processing complexity separate attested systems from possible ones better
than the original trade-off, and the constraints previously needed to rule
out unnatural systems follow from regularity. Towards a diachronic-style explanation,
\citet{silvi2025learningefficientrecursivenumeral} show that pairs of reinforcement-learning agents
communicating through a grammar based on \citet{hurford2011linguistic} can move
their systems toward Pareto-optimal configurations comparable to human ones. 

\paragraph{Several factors in one domain}

This literature has implicitly separated three different
places where complexity can live in a numeral system. One is the complexity of
an individual numeral: how hard \emph{three hundred and five} is to produce or
parse. Another is the regularity of the grammar that generates the whole
range of numerals, which is a property of the system and not of any single expression in it. A
third is the balance between lexicon size and expression length, which
depends on how often each number is needed (i.e.\ communicative need). 
These correspond closely to the
representational, organizational, and communicative factors that we introduce in the next
section.  
A regular, compact numeral grammar plays the same role as a color partition generated by a single global rule, such as assigning each point to its nearest prototype: both are properties of how the system is generated, not of any of its individual members.

Numerals are therefore an unusually good place to ask which factors a
universal actually requires. All three quantities can be measured
independently for every attested and unattested system, and the recent
history of the domain already shows that they come apart: the trade-off
between lexicon size and expression length is real but, on \citeauthor{prasertsomRecursiveNumeralSystems2026}'s
analysis, not enough on its own. The fourth factor that we will discuss in the next section, cultural
transmission, has an obvious role to play as well. As discussed in \S\ref{sec:diachronic-explanations}, iterated learning
experiments show that a pressure to be learnable together with a pressure to
be expressive can produce compositional, regular structure from unstructured
beginnings \citep{kirby2015compression}. 
In the next section (\S\ref{sec:taxonomy}), we will sketch a program that can answer whether the regularity of numeral grammars is an organizational preference already present in individual learners or is the outcome of transmission and communication working together.

\paragraph{Composition and the division of labor}

The lessons of numerals extend to other domains, because every language faces
the same choice: lexicalize a distinction with a dedicated word, or express it
compositionally. English has no basic color term for light blue and so builds
a complex expression, \emph{light blue}; Russian has a basic term, \emph{goluboy}, alongside
\emph{siniy} for darker blues. Gradable
adjectives are routinely modified (\emph{very tall}, \emph{almost empty})
instead of being multiplied into new words. Quantifiers too are mostly
complex, and the universals of \S\ref{sec:monotonicity-definition} and \S\ref{sec:quantity-learning} were deliberately stated for the
monomorphemic ones. In each case, a language reaches high expressive power by
composition, at some cost in the length of its expressions, while keeping its
lexicon small.

The trade-off studied for numerals is a natural model for this choice. It
predicts that distinctions which are needed often should tend to be
lexicalized, and rarer ones composed, in line with the Zipfian finding that shorter
forms attach to more frequent meanings \citep{Zipf1949, piantadosiZipfsWordFrequency2014}.  Testing such predictions requires two
extensions of the methods in this Element. The first is a generalization of
our schema from lexical items to systems of expressions, so that a universal
can state a property of the lexicon together with the grammar that extends
it \citep{carcassiCompositionalityLexiconEvolutionary2026}. 
The second concerns the comparison class: for numerals, the space of
possible languages is a space of grammars, and as the discussion above shows,
how that space is generated can change the result. 
Similarly, languages often mark event participants either with explicit morphemes (case markers) but then tend to have more flexible word order; or they tend to rely on rigid word order to express who did what to whom \citep{sapirLanguageIntroductionStudy1921, mcfaddenMorphologicalCaseWordOrder2003, futrellQuantifyingWordOrder2015a}.  This, too, represents the choice of whether to express a given semantic domain lexically or via syntactic structure.

Both of these cases illustrate the importance of moving computational explanations beyond the lexicon.  In the limit, the ideal target of analysis is the form-meaning mapping as a whole, including both a lexicon and a full grammar, and spanning multiple semantic domains instead of one at a time \citep{steinert-threlkeldExplainingSemanticTypology2022}.

\subsection{Towards a principled taxonomy}
\label{sec:taxonomy}

Looking back at Table~\ref{tab:explanation-outline}, one feature of the explanations in this Element stands out: almost every one of them is a sufficiency result. We saw that
neural networks trained from scratch find monotone quantifiers easier to learn (\S\ref{sec:monotonicity-learning}), that iterated learning can turn random quantifiers into
monotone ones (\S\ref{sec:iterated-learning}), and that languages closer to the Pareto frontier have more monotone quantifiers (\S\ref{sec:quantifiers-effcomm}). Each result shows that a particular
mechanism is enough to favor the universal. None of them shows that the mechanism is \emph{necessary}. This is what one should expect from how-possibly explanations, but it is exactly what makes adjudication difficult, as discussed above. Monotonicity alone now has three sufficient explanations, and nothing in the individual studies tells us which of them, if any, is doing the work in actual languages.

The difficulty does not come from a flaw in any particular study. It comes from how the explanations have been validated: each on its own terms, by
modeling its own mechanism and leaving the competitors out. In this section we sketch what we think the next step should look like. The idea is to put
the competing mechanisms on common ground and to ask, for each universal, which of them are necessary and which are sufficient. We will also suggest
that the answer need not be the same for every universal, and that where it differs, the difference may itself be predictable.

\paragraph{From sufficiency to necessity.}

Recall from \S\ref{sec:parameters} that a universal is substantial only relative to a space of possible meanings, and that our explanations work by comparing attested systems with non-actual ones. This gives a natural way to make necessity precise. Fix a universal, a comparison class of logically possible but unattested systems, and a model in which several candidate factors 
operate. A factor is \emph{sufficient} if it separates the attested systems from the comparison class on its own, and \emph{necessary} if removing it from the model, with the others still in place, makes the model mis-predict that boundary.

Two things make this harder than it sounds. First, the factors are correlated. Efficient systems tend also to be simple, and simple systems tend also to be easy to learn and to survive transmission. A language may lie near the Pareto frontier because communicative pressure put it there, or because learners independently prefer systems that happen to be efficient. Showing that attested systems are near-optimal is therefore a claim about fit, and
it is silent on which pressure is actually required. Second, what counts as a misprediction, and against which unattested systems, has to be fixed before any mechanism is run. Otherwise the comparison class can quietly be tuned to the answer. This is one more reason to take seriously the worry from \S\ref{sec:parameters}: the space of possible languages is not a background assumption of our program but part of its design.

The experimental logic this suggests is a dissociation design: remove each factor in turn while controlling for the others. Some of this can be done with people. A participant can learn a novel lexical system with no (other) pressure applied, then use it with a partner in a reference game in which communicative need is manipulated, and finally pass it on to a new learner in
an iterated learning transmission chain. Each stage adds one mechanism to the previous one, so its contribution can be read off directly. Both ends of such a paradigm already exist: adults learning a novel quantifier track the monotonicity universal \citep{RAMOTOWSKA2026106523}, and monotone quantifiers emerge through iterated learning \citep{carcassiMonotoneQuantifiersEmerge2021b}. The worry from \S\ref{sec:supervised-learning} applies here too, however. Adult participants already carry biases toward the very universals being explained. For that reason, computational agents remain indispensable, since they can be built without those biases, and human experiments serve as a check that an effect found in the models also shows up in real cognition. Where the factor in question is the structure of the space itself, as with the perceptual space of color, experiments can vary the categories people learn over that space but not the space; that part of the evidence has to come from models, which can be run on altered spaces, and from typology \citep{ranjanWhenEfficientCommunication2026}.

Software infrastructure for building these experiments remains essential.  \citet{imelUnnaturalLanguageToolKit2025} introduce the Unnatural Langauge Toolkit (available as well at \url{https://clmbr.shane.st/ultk}) to help systematize much of this work.  In particular, it provides standard formats for reading/writing data about semantic spaces, and then implements many of the efficient communication and learning-style analyses in this paper.  Although development is still ongoing, this library (and others related) can help drive this program by providing a standard setup to compare analyses with different mechanisms and across different meaning spaces.

\paragraph{Four constraining factors}

What should the candidate factors be? The two optimization principles---ease of learning and efficient communication---of this Element are the natural starting point. Looked at closely, though, each of these bundles more than one possible source of pressure.
The complexity term of the \textsc{Efficient Communication Hypothesis} is a measure of how hard a meaning is to represent, computed in a language of thought (\S\ref{sec:complexity-cost}). That is close to what Chapter~\ref{chapter:learning}
measured with learning. Ease of learning, in turn, works both within an individual and, through iterated learning, across generations. Pulling these
strands apart, we arrive at four factors. 

\begin{enumerate}

 \item \emph{Representation.} Some meanings recur because they are simple to represent and easy to learn. This is the core of Chapter~\ref{chapter:learning} and of the complexity term in Chapter~\ref{chapter:effcomm}. It comes with several candidate measures: description length in a language of thought, logical complexity, and learnability by neural networks \citep{steinert-threlkeldEaseLearningExplains2020, polQuantifiersSatisfyingSemantic2023, haberlandQuantifiersThatAre2025}. Whether these measures are one construct or several remains open, and the choice of primitives for the language of thought is one place where the answer could be decided \citep{piantadosiLogicalPrimitivesThought2016, CARCASSI2023105541}.
 
\item \emph{Organization.} Some constraints concern how a whole system divides up a meaning space, not how hard any single meaning is. Convexity is usually stated for a single term. Still, it can also arise at the level of the system: if every point in a space is assigned to its nearest prototype, all categories come out convex at once, as a by-product of one global rule \citep{gardenforsConceptualSpacesGeometry2000, jagerEvolutionConvexCategories2007}. Numerals give a different example. The regularity of the grammar that generates the whole numeral range is a property of the system, not of any individual numeral \citep{prasertsomRecursiveNumeralSystems2026}.

\item \emph{Communication.} Languages are used, and use favors some systems over others \citep{smithCommunicativePressuresShape2025}. This is the subject of Chapter~\ref{chapter:effcomm}, including costs on the side of form, such as the trade-off between lexicon size and the length of numeral expressions \citep{denicRecursiveNumeralSystems2024}. 

\item \emph{Transmission.} Transmission links individual biases to population-level structure (\S\ref{sec:iterated-learning}). It is an open question whether it adds structure of its own or only reveals the learners' bias. Bayesian learners who sample from their posterior converge to their prior \citep{giffithsKalish}, while learners who choose the most probable hypothesis amplify it \citep{kirby2007innateness}. Which regime holds decides whether transmission is a separate mechanism or a population-level expression of the representational one. For neural learners, there is no analogous result that says what the chain converges to \citep{renCompositionalLanguagesEmerge2019, carcassiMonotoneQuantifiersEmerge2021b, renBiasAmplificationLanguage2024, jarvisCompositionalitySystematicityEmerge2026}, which is part of why the question is open. 
\end{enumerate}

Before moving on, we make two remarks on this list. First, whether representational complexity and communicative pressure are separate axes or a single objective optimized
jointly, as in the Information Bottleneck (\S\ref{sec:ib}), is itself something to test. Second, some apparent universals may fall outside all four. On the
view discussed in \S\ref{sec:conservativity-learning}, conservativity is a property of the syntax-semantics interface rather than of the lexicon, that is, an architectural universal in
the sense of \S\ref{sec:semantic-universals}. If so, none of the four factors may be necessary, and the failure of the ease of learning test (and others) is what one would expect. The same example shows that universals within a single domain can have different origins: monotonicity and conservativity both concern determiners, yet they may call for entirely different explanations.

Finally, modeling choices of the kind we have flagged throughout can also flip a verdict. For gradable adjectives, simplicity and accuracy alone predict largely non-monotone meanings; monotone ones prevail only once the agents reason pragmatically \citep{schouwstra2018evolution}. Whether agents are
literal or pragmatic, like the choice of primitives in a language of thought, should therefore be treated as a variable to manipulate in the communication setting, not a setting to fix \citep{denicIndefinitePronounsOptimize2022, imelEfficientCommunicationAnalysis2026}.

\paragraph{Predicting which factors are necessary}

So far we have described how one might establish, for a given universal, which factors it requires.  That is: which factors are \emph{necessary} for accurately separating the attested from the unattested semantic systems. A further, more ambitious question is whether those requirements can be predicted in advance, from properties of the semantic domain alone. We illustrate with the property that recurs most often across Table~\ref{tab:explanation-outline}: convexity. Color terms denote convex regions of color space (\S\ref{sec:convexity-learning}). Convex quantifiers are conjunctions of monotone ones (\S\ref{sec:monotonicity-learning}). Haspelmath's universal for
indefinites requires connected regions of a semantic map (\S\ref{sec:indefinites}), which \citet{croftRadicalConstructionGrammar2001} generalizes as the Semantic Map Connectivity Hypothesis: any language-specific category should map onto a connected region in conceptual space.
The IFF universal for modals can be seen as convexity
on a discrete space (\S\ref{sec:modals}). It is tempting to conclude that one meta-universal calls for one explanation. However, the opposite may be true. 

Convexity presupposes a relation of betweenness, and the domains differ in where that relation comes from. In
color, it is supplied from outside language: perceptual similarity is fixed by the visual system, and the CIELab space is built from human similarity judgments. For indefinites and modals it is not. There the space is constructed by linguists, and largely from the behavior of the expressions themselves: edges in Haspelmath's map are drawn between functions that some language expresses with a single item, and the force-flavor space has no pre-linguistic counterpart that speakers could perceive.

This suggests a prediction. Where betweenness is given from outside
language, the organization factor may be necessary: no account that
does not refer to the substrate should draw the boundary between
attested and unattested systems. Where betweenness is defined inside
language, convexity may instead reduce to representational simplicity under communicative pressure, with no further organizational factor required. The modals case fits this picture. With complexity measured in a propositional language of thought over force and flavor atoms, every Pareto-optimal language satisfies IFF \citep{imelEfficientCommunicationAnalysis2026} (and natural languages are more optimal than artificial ones), even though the model
contains no organizational term at all.
For color, the question is open. \citet{imelConvexityEfficiencySemantic2026} find efficiency to be the better single predictor of attested color systems, with convexity adding little on top, but the two are closely correlated.
Both compare fit, but neither asks whether convexity would still be predicted if the geometry of the space were changed while representation and need were held fixed, for instance by running the same models on a warped or rotated color space.  %
The prediction can therefore fail in either direction: if representational complexity alone drew the boundary in color, or if modals turned out to need an organizational mechanism after all.
Towards this direction, \cite{ranjanWhenEfficientCommunication2026}, after showing that IB optimization produces convex color systems, go on to systematically manipulate various parameters of the semantic space to identify \emph{under what conditions} does a pressure for efficiency produce convexity.  

For the prediction to have content, the predictor must be established
without looking at the lexicon it is meant to explain. The relevant
evidence is non-linguistic: infant and animal discrimination,
psychophysics, similarity judgments. Color passes this test, and so does
number, through the approximate number system \citep{feigenson2004core}.
Modals and indefinites do not. Kinship is an interesting open case: the
genealogical substrate is real, but whether speakers represent kin space
independently of their kin terms is an empirical question
\citep{kempKinshipCategoriesLanguages2012}.

The Semantic Map Connectivity Hypothesis makes the point vivid. A semantic map is built from cross-linguistic co-expression: two functions are linked when some language expresses both with a single item. Indeed, when semantic maps are inferred from data, connectedness of the attested categories is built into the inference problem itself, so the hypothesis can be tested only on held-out languages \cite{regier2013inferring}. This is not a flaw of the method, but it shows that when a space is built from the expressions themselves, the space and the universal stated over it are not independent. Meaning spaces thus come in at least three kinds: given by perception (color), defined by a logical theory (quantifiers), and constructed from co-expression (indefinites). The prediction above concerns the first and the last; where logically defined spaces fall is an open question..%

\paragraph{A research program}
Putting all of these pieces together yields a two-step program. The first is
to establish, universal by universal, a causal profile: which of the four
factors are necessary and which are sufficient. The second step asks whether these causal profiles are themselves predictable from independently measurable properties of a domain, above all whether speakers represent its structure independently of language. Such a mapping can be tested honestly by building it on some domains and using it to predict the profiles of other held-out domains. This program compares domains, but it still analyzes each one on its own. As we noted at the end of \S\ref{sec:diachronic-explanations}, actual languages are single form–meaning systems that span many domains at once. Systems of quantification and modality interact with each other and with color, for instance. %
A profile computed for one domain at a time may therefore miss pressures that act across domains, and extending the program in that direction is a natural next step.

If the mapping works, we would have a theory of why a universal has the
origin it does, one that could anticipate the profile of a universal nobody has yet analyzed. If it does not work, the result would still be of interest. Much current work implicitly assumes that a single kind of explanation carries over from
one domain to the next. Idiosyncratic profiles would show that it does not. Either way, the question shifts from whether some particular factor can produce a universal (a how-possibly explanation) to why a given universal requires the factors it does. One should ask not only whether a general pressure could produce it, but which pressures it actually requires. 

\subsection{Taking Stock}

In this Element, we have introduced the idea of Lexical Property Universals, properties of word meaning shared across all words of a given type across languages, and developed computational explanations of many such universals in terms of two optimization principles: to be easy to learn, and to be efficient in communication.  We consider these explanations to be exciting starting points for explaining why natural language meaning looks the way it does and how languages may reflect more general aspects of representation, communication, organization, and transmission.  Much more work remains to be done, including identifying which factors are necessary for particular universals we have studied, and using those results to build a general predictive theory of what factors will be active in explaining universals based on properties of the semantic domain.  We hope that readers will join us in taking it further.

\appendix  %

\bibliography{refs}

\section{Degrees of Monotonicity}
\label{chap:monotonicity-appendix}

In Section~\ref{sec:monotonicity-definition}, we gave an intuitive description of a measure of the degree of monotonicity.  This Appendix makes that definition fully precise.  After doing so, we then generalize the definition to show how the quantifier definition is a special case of a more general one and then discuss future directions in theorizing about degrees of monotonicity.

\subsection{For Quantifiers}

We measure upward monotonicity in information-theoretic terms as the proportion of uncertainty in the truth-value of a quantifier that is removed after knowing that there is a \emph{submodel} where the quantifier is true.\footnote{See \cite{coverElementsInformationTheory2006} for a more thorough introduction to the information-theoretic concepts used here.} In particular, we define $\mathbb{M}' \preceq \mathbb{M}$ ($\mathbb{M}'$ is a submodel of $\mathbb{M}$) just in case $M' = M$, $A' = A$, and $B' \subseteq B$.  For a perfectly upward monotone quantifier $\mathsf{Q}$, if a structure $\mathbb{M}$ has a submodel which belongs to the quantifier, then $\mathbb{M} \in \mathsf{Q}$ as well. Therefore, for an upward monotone quantifier all the uncertainty is removed and the measure will have value 1.

To make this more precise, call $\mathcal{M}$ the set of all structures. Let $\{ \mathcal{M}, \mathcal{F}, P \}$ be a probability space with $\mathcal{F}=2^{\mathbb{M}}$ and $P$ a discrete probability function with full support.\footnote{This assumption can be relaxed, but is satisfied by all of the distributions that we will use for explanations in this Element.} Then, define two binary random variables $\mathbbm{1}_\mathsf{Q}$ and $\mathbbm{1}_\mathsf{Q}^\preceq$ as follows, with $\mathbb{M} \in \mathcal{M}$:
\begin{align*}
\mathbbm{1}_\mathsf{Q}(\mathbb{M}) &= 1 \text{ iff } \mathbb{M} \in \mathsf{Q}
\\
\mathbbm{1}_\mathsf{Q}^\preceq(\mathbb{M}) &= 1 \textrm{ iff } \exists \mathbb{M}' \preceq \mathbb{M} \text{ s.t. } \mathbb{M}^\prime \in \mathsf{Q} 
\end{align*}
In words, $\mathbbm{1}_\mathsf{Q}$ is the indicator function of the quantifier $\mathsf{Q}$. $\mathbbm{1}_\mathsf{Q}^\preceq$ is 1 at a model just in case that model has a submodel that belongs to $\mathsf{Q}$.

The entropy of $\mathbbm{1}_\mathsf{Q}$, $H(\mathbbm{1}_\mathsf{Q})$, quantifies the uncertainty about what truth value $\mathsf{Q}$ will assign to a model. In other words: how much uncertainty does one have about the raw truth-value of the quantifier.  This can be seen equivalently as a measure of how many binary questions one would need to ask to determine the truth-value, or as the average length of an optimal code for expressing the truth value. 

The conditional entropy $H( \mathbbm{1}_\mathsf{Q} \mid \mathbbm{1}_\mathsf{Q}^\preceq )$ quantifies the uncertainty about what $\mathsf{Q}$ will assign to a model, given that one knows whether or not the model has a submodel belonging to it.  An upward monotone quantifier minimizes (at value 0) $H( \mathbbm{1}_\mathsf{Q} \mid \mathbbm{1}_\mathsf{Q}^\preceq )$: if one knows that a model has a true submodel (i.e.\ $\mathbbm{1}_\mathsf{Q}^\preceq(\mathbb{M}) = 1$), and the quantifier is upward monotone, one knows the truth value at that model ($\mathbbm{1}_\mathsf{Q}(\mathbb{M}) = 1$). The difference between the entropy and the conditional entropy between these variables is known as the mutual information: 
$$I(\mathbbm{1}_\mathsf{Q} ; \mathbbm{1}_\mathsf{Q}^\preceq) := H(\mathbbm{1}_\mathsf{Q}) - H(\mathbbm{1}_\mathsf{Q} | \mathbbm{1}_\mathsf{Q}^\preceq)$$
This measures how much information $\mathbbm{1}_\mathsf{Q}^\preceq$ provides about $\mathbbm{1}_\mathsf{Q}$.  For an upward monotone quantifier, $H(\mathbbm{1}_\mathsf{Q} | \mathbbm{1}_\mathsf{Q}^\preceq) = 0$, and so $I(\mathbbm{1}_\mathsf{Q} ; \mathbbm{1}_\mathsf{Q}^\preceq) = H(\mathbbm{1}_\mathsf{Q})$.  In other words: for an upward monotone quantifier, knowing which structures have a true substructure provides as much information as knowing the entire quantifier.

While this roughly captures what we want from a measure of upward monotonicity, it needs to be normalized to form a degree that applies across quantifiers, since $0 \leq I(\mathbbm{1}_\mathsf{Q} ; \mathbbm{1}_\mathsf{Q}^\preceq) \leq H(\mathbbm{1}_\mathsf{Q})$. We do this by dividing by $H(\mathbbm{1}_\mathsf{Q})$, moving the upper bound to $1$.\footnote{This is sometimes called the \emph{uncertainty coefficient}, defined at least as early as \cite{theilEstimationRelationshipsInvolving1970}.} Putting the pieces together, then, we measure upward monotonicity as
\begin{align*}
    \textsf{mon}(\mathsf{Q}) &:= \frac{I(\mathbbm{1}_\mathsf{Q} ; \mathbbm{1}_\mathsf{Q}^\preceq)}{H(\mathbbm{1}_\mathsf{Q})}
\end{align*}

To see how this measure tracks intuitions, \cite{carcassiMonotoneQuantifiersEmerge2021b} report results for the previously mentioned quantifiers ``some'', ``between 3 and 5'' and ``an even number of'' on all models of a fixed size. ``Some'' receives monotonicity 1.0, because knowing whether a structure has a substructure that verifies ``some'' eliminates all uncertainty about the truth of the structure. The quantifier ``between 3 and 5'' has degree $0.7517$ and one with ``an even number of'' has degree $0.001$, which captures the intuitive order of monotonicity of these quantifiers.
We can show that maximal degree behaves as desired: being monotone simpliciter is equivalent to having the degree of monotonicity 1. 

\begin{examples}
    \item $\textsf{mon}(\textsf{Q}) = 1 \text{ iff } \textsf{Q}$ is upward monotone.\footnote{Here is a proof of this fact:
\begin{proof}
    We assume \textsf{Q} is non-trivial, so that $H(\mathbbm{1}_\mathsf{Q}) > 0$ and $\textsf{mon}(\textsf{Q})$ is defined. First, note that the degree is one just in case $H(\mathbbm{1}_\mathsf{Q} | \mathbbm{1}_\mathsf{Q}^\preceq) = 0$.  It is commonly known that $H(X | Y) = 0$ iff $X = f(Y)$ for some function $f$ from $Y$'s value space to $X$'s value space.  In our case, this means that $\textsf{mon}(Q) = 1$ iff there is an $f:\{0,1\} \to \{0, 1\}$ such that $\mathbbm{1}_\mathsf{Q}(\mathbb{M}) = f(\mathbbm{1}_\mathsf{Q}^\preceq(\mathbb{M}))$. Now, because $\preceq$ is a reflexive relation, $\mathbbm{1}_\mathsf{Q}^\preceq(\mathbb{M}) = 0$ implies that $\mathbbm{1}_\mathsf{Q}(\mathbb{M}) = 0$ in general.  Finally, note that we must have $f(1) = 1$.  Consider a model $\mathbb{M}$ such that $\mathbbm{1}_\mathsf{Q}^\preceq(\mathbb{M}) = 1$.  Then, for some $\mathbb{M}' \preceq \mathbb{M}, \mathbbm{1}_\mathsf{Q}(\mathbb{M}') = 1$.  Because $\preceq$ is reflexive, we also have that $\mathbbm{1}_\mathsf{Q}^\preceq(\mathbb{M}') = 1$, and so $f(1) \neq 0$.  The reader can verify that $f(0) = 0$ and $f(1) = 1$ just in case $\mathsf{Q}$ is upward monotone.
\end{proof}    
}
\end{examples}

The measure defined above is a measure of upward monotonicity. It can be straightforwardly modified to measure downward monotonicity, by replacing the variable $\mathbbm{1}_\mathsf{Q}^\preceq$ with the variable $\mathbbm{1}_\mathsf{Q}^\succeq$, which is true when a model has a \emph{super}model that belongs to $\mathsf{Q}$. These measures can equally well be applied to the first or second argument of a quantifier (the restrictor and the scope), yielding four possible degrees.  Different studies can report any of these four, or choose to aggregate them into one measure by taking the mean or the maximum.

\subsection{Generalized Degree of Monotonicity}

Following \citet{haberlandQuantifiersThatAre2025}, we now generalize the above definition to the general setting where $f:\langle A, \leq_A\rangle \to \langle B, \leq_B\rangle$ is an arbitrary function between partially ordered sets. We define $f^\uparrow$, the minimal monotone extension of $f$, by $f^\uparrow(x) = \sup \{ f(x') : x' \leq_A x \}$. (Note: this definition requires that suprema exist in $B$.) This definition has a few desirableproperties: (i) $f^\uparrow$ is monotone; (ii) $f$ is monotone iff $f = f^\uparrow$; (iii) $f^\uparrow$ is the `least' monotone extension of $f$: if $f \leq g$ (where this means that $f(x) \leq g(x) \forall x \in A$) and $g$ is monotone, then $f^\uparrow \leq g$. 

We then define: $\textsf{mon}(f) := I(f ; f^\uparrow) / H(f)$, where $I$ is the mutual information and $H$ is the entropy. This measures how much uncertainty about the value of $f$ is resolved by knowing the value of its minimal monotone extension $f^\uparrow$.  From property (ii), it follows that $\textsf{mon}(f) = 1$ \emph{iff} $f$ is monotone.  For the application of this definition to quantifiers, $f$ can be taken to be the characteristic function of a generalized quantifier with respect to the $B$ argument. As an example, if $f$ is the characteristic function of ``between 3 and 5'', the resulting minimal monotone extension ($f^\uparrow$) would be the characteristic function of ``at least 3''.  As before, this measure easily generalizes to one for downward monotonicity, and can be aggregated across both arguments of quantifiers by max or mean.

\subsection{Future Directions}

Although this conception of the degree of monotonicity has many desirable properties, more work remains to be done to validate it and compare it to alternatives.  To the former end, we note that on many definitions of distance in function spaces, the minimal monotone extension will not be the closest monotone function to $f$.  It is worth comparing the existing measure to one which more directly finds the closest monotone function to a given one.  Relatedly, tools from the Language of Thought approach to thinking of semantic complexity discussed throughout this Element can be applied to this situation: a natural approach to defining degrees is to use a LoT where the primitives are all the monotone functions, and they can be combined by boolean operations of $\land, \lor, \lnot$, assuming those are defined on the relevant domain of the function.  Length of the shortest formula denoting a function will then be inversely correlated with monotonicity.  Having highlighted these various design considerations in defining a degree of monotonicity, an exciting future avenue is to axiomatize these desiderata and find which definitions (if any) satisfy them.  It is also worth comparing alternative degree definitions to existing human learning data to see whether such data is powerful enough to distinguish between them.  We leave all of this for future work.

\section*{Acknowledgements}

This Element grew out of many years of joint work, and we owe much to the people who shaped it. We thank our students at the University of Trento and the University of Washington, whose questions in classes and seminars pushed us to clarify these ideas, and whose projects made their way into these pages. We are especially grateful to our co-authors on the papers discussed here; much of what we report is as much their work as ours. We have also learned a great deal from colleagues across linguistics, cognitive science, philosophy, and computer science: by reading their work, by discussing it with them at workshops and conferences, and from the feedback of anonymous reviewers on our papers, which often improved them considerably.  The people who have shaped our thinking via these interactions are too numerous to list by name, but we express our gratitude to all of you.  We thank the series editors for their patience (and the reviewers of this Element for their helpful comments).

\end{document}